Predictive Scheduling of Collaborative Mobile Robots for Improved Crop-transport Logistics of Manually Harvested Crops

By

CHEN PENG

DISSERTATION

Submitted in partial satisfaction of the requirements for the degree of

DOCTOR OF PHILOSOPHY

in

Mechanical and Aerospace Engineering

in the

OFFICE OF GRADUATE STUDIES

of the

UNIVERSITY OF CALIFORNIA

DAVIS

Approved:

_______________________________________
Stavros G. Vougioukas, Chair

_______________________________________
David C. Slaughter

_______________________________________
Zhaodan Kong

Committee in Charge

2021

# Acknowledgements

In the spring 2017, I decided to quit my original research project with my first PI at UC-Davis and joined the Bio-Automation Lab of Dr. Stavros Vougioukas. My work started that summer with data collection in strawberry fields. In Nov 2020, my Ph.D. journey came to an end with a successful field experiment in a strawberry field.

I really appreciate my mentor, Dr. Stavros, in both my research and life experiences. He gave tremendous support, encouragement, and assistance in times of confusion and problems as I worked towards my Ph.D. Without his advising, I would be much more challenged to finish my Ph.D. goals. I am also very thankful for my colleagues and lab mates. They contributed much discussion time and support when I struggled with my research. I would be very pleased to keep an ongoing relationship with all of them in my future life.

Certainly, I am also very thankful to my wife, Min Li, for her support, accompaniment, and assistance during all days and nights in these five years. Also, I appreciate my son, Shuoheng Peng, who brought much joy and fun to the whole family. Finally, I want to dedicate this dissertation to my parents, Bin Peng and Aihe Chen, and my parents-in-law, Qindi Xia and Wenping Li. Without your support, it would not be possible to finish my Ph.D. program. Also, thanks to all my friends and relatives who ever encouraged, supported, and inspired me during my life.

# Table of contents

# Abstract

Mechanizing the manual harvesting of fresh market fruits constitutes one of the biggest challenges to the sustainability of the fruit industry. During manual harvesting of some fresh-market crops like strawberries and table grapes, pickers spend significant amounts of time walking to carry full trays to a collection station at the edge of the field. A step toward increasing harvest automation for such crops is to deploy harvest-aid robots that transport the empty and full trays, thus increasing harvest efficiency by reducing pickers' non-productive walking times. Given the large sizes of commercial harvesting crews (e.g., strawberry harvesting in California involves crews of twenty to forty people) and the expected cost and complexity of deploying equally large numbers of robots, this dissertation explored an operational scenario in which a crew of pickers is served by a smaller team of robots. Thus, the robots are a shared resource with each robot serving multiple pickers.

If the robots are not properly scheduled, then robot sharing among the workers may introduce non-productive waiting delays between the time when a tray becomes full and a robot arrives to collect it. Reactive scheduling (e.g., "start traveling to a picker when the tray becomes full") is not efficient enough, because robots must traverse large distances to reach the pickers in the field, thus introducing long wait times. Predictive scheduling (e.g., "predict when and where a picker's tray will become full and dispatch a robot to start traveling there earlier, at an appropriate time") is better suited to this task, because it can reduce or eliminate pickers' waiting for robot travel. However, uncertainty is always present in any prediction, and can be detrimental for predictive scheduling algorithms that assume perfect information. Therefore, the main goal of this dissertation was to develop a predictive scheduling algorithm for the robotic team that incorporates prediction uncertainty and investigates the efficiency improvements in simulations and field experiments.

In the first part of this dissertation, strawberry harvesting was modeled as a stochastic process and dynamic predictive scheduling was modeled under the assumption that, once a picker starts filling a tray (a stochastic event), the time and location when the tray becomes full - and a tray transport request is generated - are known exactly. The resulting scheduling is dynamic and deterministic, and we refer to it as 'deterministic predictive scheduling' to juxtapose it against stochastic predictive scheduling under uncertainty, which is addressed afterwards. Given perfect 'predictions', near-optimal dynamic scheduling was implemented to provide efficiency upper-bounds for stochastic predictive scheduling algorithms that incorporate uncertainty in the predicted requests. Robot-aided harvesting was simulated using manual-harvest data collected from a commercial picking crew. The simulation results showed that given a robot-picker ratio of 1:3 and robot travel speed of 1.5 m/s, the mean non-productive time was reduced by over 90% and the corresponding efficiency increased by more than 15% compared to all-manual harvesting.

In the second part, the uncertainty in the predictions of tray-transport requests was incorporated into scheduling. This uncertainty is a result of stochastic picker performance, geospatial crop yield variation, and other random effects. Robot predictive scheduling under stochastic tray-transport requests was modeled and solved by an online stochastic scheduling algorithm, using the multiple scenario approach (MSA). The algorithm was evaluated using the calibrated simulator, and the effects of the uncertainty on harvesting efficiency were explored. The results showed that when the robot-to-picker ratio was 1:3 and robot speed was 1.5 m/s, the non-productive time was reduced by approximately 70%, and the corresponding harvesting efficiency improved by more than 8.5% relative to all-manual harvesting.

The last part of the dissertation presents the implementation and integration of the co-robotic harvest-aid system and its deployment during commercial strawberry harvesting. The evaluation experiments demonstrated that the proof-of-concept system was fully functional. The co-robots improved the mean harvesting efficiency by around 10% and reduced the mean non-productive time by 60%, when the robot-to-picker ratio was 1:3. The concepts developed in this dissertation can be applied to robotic harvest-aids for other manually harvested crops that involve a substantial human-powered produce transport, as well as to in-field harvesting logistics for highly mechanized field crops that involve coordination of harvesters and autonomous transport trucks.

# Chapter 1   Introduction

Mechanizing the manual harvesting of fresh market fruits constitutes one of the biggest challenges to the sustainability of the fruit industry. Depending on the commodity, labor for manual harvesting can contribute up to 60% of the yearly operating costs per acre (Bolda et al., 2016). Additionally, recent studies indicate that the farm labor supply cannot meet demand in many parts of the world because of socioeconomic, structural, and political factors (Charlton et al., 2019; Guan et al., 2015). Despite recent progress on shake-catch approaches for mechanical harvesting of apples (He et al., 2017) and cherries (Zhou et al., 2016), fruit quality and collection efficiency are still not adequate to justify the adoption of these technologies for fresh market tree fruits that have delicate skin. Shake-catch harvesting is also not applicable to high-value crops like fresh strawberries, raspberries, blackberries, and table grapes, which are very fragile and must be harvested selectively, based on ripeness criteria, without damage.

Robotic harvester prototypes are being developed and field-tested for high-volume, high-value crops such as apples (Silwal et al., 2017), kiwifruit (Williams et al., 2020), sweet pepper (Arad et al., 2020), and strawberries (Xiong et al., 2020). However, the developed robots have not, to date, successfully replaced the judgment, dexterity, and speed of experienced pickers at a competing cost; the challenges of high fruit picking efficiency and throughput remain largely unsolved (Bac et al., 2014).

As an intermediate step to full automation, mechanical labor aids have been introduced to increase worker productivity by reducing workers' non-productive times. For example, orchard platforms eliminate the need for climbing ladders and walking to unload fruits in bins (Baugher et al., 2009; Fei & Vougioukas, 2021). Autonomous vehicle prototypes have been developed to assist in bin management in orchards (Bayar et al., 2015; Ye et al., 2017), to reduce the need for forklift operators.

In strawberry production, mobile conveyors have been introduced to reduce the time pickers spend walking to get the produce from the plants to the designated loading stations and return to resume picking (Rosenberg, 2003). However, such conveyors are specific to strawberries and cannot be adapted to other crops. Furthermore, their adoption has been very slow, partly because of their questionable profitability, due to high purchase cost and limited efficiency gains. Two reasons for their inadequate efficiency are: 1) row-turning in the field is time-consuming because of their large size, and 2) because conveyors move slowly to accommodate slower pickers, often resulting in underutilization of faster pickers (Doody, E., 2019).

The walking time to carry harvested crops constitutes a significant non-productive part of the harvesting cycle for several fresh-market crops, like strawberries (Figure 1), raspberries, blackberries, and table grapes. For strawberries, walking time has been measured to reach up to 22% of the total harvest time (Khosro Anjom et al., 2018); higher inefficiencies are often reported, anecdotally.

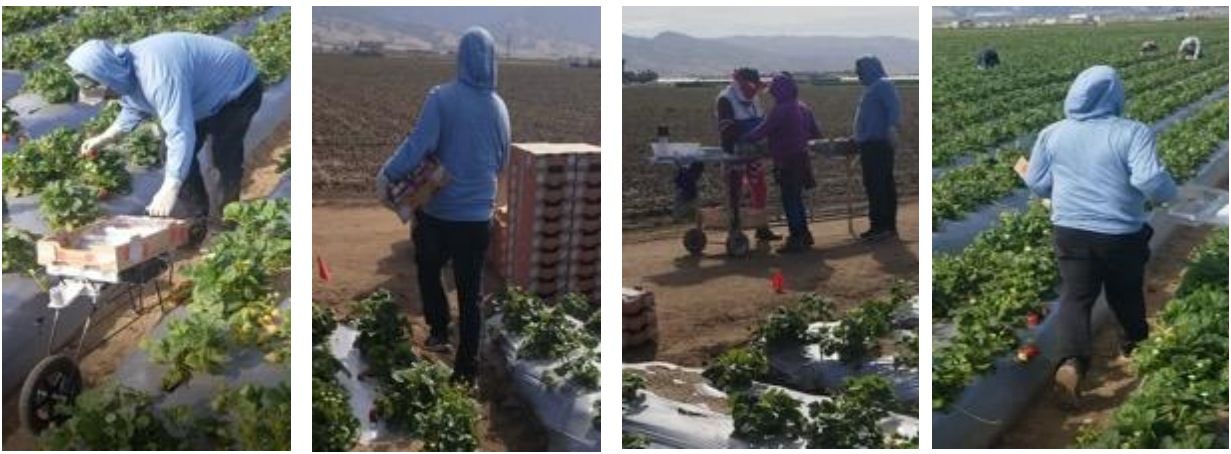

In this dissertation, a collaborative robotic system (aka, co-robotic system) was investigated to assist in such harvesting operations by transporting trays, with strawberries as a case study (USDA REEIS, 2013). During the proposed robot-aided harvesting, each picker walks inside a furrow, harvests ripe fruits, and puts them in a standard-sized tray located on a special instrumented cart (Figure 2.a), in the same way as in all-manual harvesting. These carts are equipped with load cell sensors to measure the weight of the tray and a GNSS (Global Navigation Satellite System) module to record the geodetic locations of the carts (Khosro Anjom et al., 2018). The cart sends data wirelessly in real-time to a computer in the field (we refer to it as the "operation server"). Software running on the server predicts when and where a tray will become full (Khosro Anjom & Vougioukas, 2019). A full tray results in a tray-transport request to the scheduling software running on the server, which dispatches a team of crop-transport robots to serve those requests. The robots travel between the collection station and pickers to bring empty trays (Figure 2.b). The picker walks a small distance to the robot, loads the full tray, gets an empty tray, and pushes a button to command the robot to travel back to the collection station (Figure 2.c).

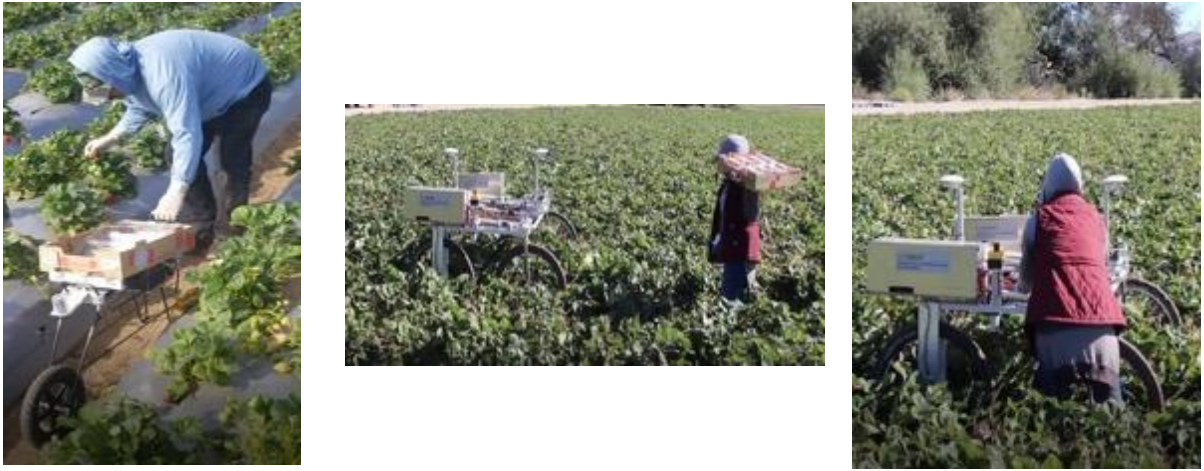

*Figure 2. The working cycle of co-robotic harvesting in an open commercial field: a) the picker is picking strawberries inside the furrow in the same way as the manual harvesting; b) the picker walks a small distance to the serving robot; c) the picker loads the trays on the robot;*

Given the large sizes of commercial harvesting crews (e.g., strawberry harvesting in California involves crews of twenty to forty people) and the expected cost and complexity of deploying equally large numbers of robots, this work explored an operational scenario in which a crew of pickers is served by a smaller team of robots. Thus, the robots are a shared resource with each robot serving multiple pickers. Given that robots travel at relatively low speeds for safety purposes (in our case, 0.5 – 1.5 m/s), and that the distance to a picker can be up to 100 m long, robot sharing among the workers may introduce non-productive waiting delays between the time when a tray becomes full and a robot arrives to collect it, if the robots are not properly scheduled. Hence, *efficient robot dynamic scheduling* is essential to ensure that the overall reduction in walking time is larger than the waiting time introduced by robot operation.

Two main types of scheduling exist: reactive and predictive. In reactive scheduling (Blazewicz et al., 2019), a machine/vehicle/robot at the collection station is allocated to a task only after the scheduler receives the task request. In the context of tray-transport robots for harvesting, reactive scheduling refers to the situation where a robot starts traveling to a picker when the picker's tray becomes full. Seyyedhasani et al. (2020b) showed that when tray-transport robots were scheduled reactively, picker waiting time was reduced when the robot-to-pickers ratio increased. However, above a certain ratio, adding more robots did not reduce further the waiting time. The reason is that a picker's waiting time is at least as much as the time needed for a robot to travel the distance from the collection station to the picker.

Predictive scheduling policies incorporate information about future demand into the scheduling (Ritzinger et al., 2016). In the context of harvesting, 'future demand' refers to knowledge about when and where a worker's currently used tray will become full, giving rise to a tray-transport request. If the time and location are known in advance, a robot can be dispatched – and start moving toward that location - before the tray becomes full; hence, waiting times due to robot travel can be reduced or eliminated. The locations and times of tray-transport requests contain uncertainty because of stochastic picker work-rate and varying – unknown - yield density (Khosro Anjom & Vougioukas, 2019). Uncertainty can be detrimental for predictive scheduling algorithms that assume perfect information (Bertsimas & Ryzin, 2017; Blazewicz et al., 2019). Hence, in this work, dynamic stochastic scheduling algorithms were investigated to account for the prediction uncertainty and improve the performance (Bent & Van Hentenryck, 2004; Blazewicz et al., 2019; Ichoua et al., 2006).

## 1. Dissertation Objectives

The overall goal of this research was to develop a co-robotic harvest-aiding system which performs in-field tray logistics by serving human pickers' fruit-transport requests. At a high level, the research objectives are as follows:

(1) Develop a modeling framework for manual strawberry harvesting supported by a robot team providing crop-transport logistics in a simulation environment.

(2) Mathematically model the predictive scheduling of the robot team in the context of co-robotic harvesting and have the scheduling problem solved in an online fashion.

(3) Develop a prototype of a fully functional co-robotic system that performs tray transportation in real-world harvesting and evaluate its performance.

To achieve these objectives, the following research activities were conducted and presented as follows.

Chapter 2 presents a modeling framework for a simulator of strawberry harvesting activities supported by crop-transport logistics of a robot team. The framework utilizes a hybrid system approach to model the coupled picker and robot activities. Finite state machines model discrete operating states, and difference equations describe motion and mass transfer within each discrete state. In this chapter, the predictive scheduler has access to accurate information of the pickers' next tray-transport requests, referred as "perfect predictions", after their trays start being filled. The effect of prediction timeliness and robot/picker ratio on the scheduling was investigated. In consideration of the operational feasibility and human safety, the simulated robot speed was mainly 1.5 m/s and at most 2 m/s. The optimizing goal of the scheduling problem was to minimize the mean of non-productive waiting time of the pickers. Optimized scheduling is implemented to provide efficiency upper bounds for any predictive scheduling algorithms that incorporate uncertainty in the predictive requests.

Chapter 3 introduces the practical scenarios into the developed simulator including the uncertainty of the prediction and slower robot speed based on the limitations of our mobile robot. Prediction uncertainty was processed by adapting a multiple scenario approach (MSA) in the robot scheduling. Given slower robot speed and small robot-picker ratio, request rejections to some pickers' requests were considered in the scheduling decision. This is to guarantee that the non-productive waiting time for a robot to arrive at a picker will be less than the non-productive time if a picker were to transport the tray himself. The stochastic predictive scheduler is embedded into the simulator, and extensive simulation experiments are performed. The results are analyzed to study the effect of predictive uncertainty on scheduling performance.

In Chapter 4, the physical implementation of the co-robotic system is presented, along with its evaluation during commercial strawberry harvesting. In the system implementation, a request button was designed on the pickers' instrumented carts to allow pickers to control if they want to be served by the robots. The experimental results are processed and analyzed to investigate the system performance.

In Chapter 5, the main conclusions of the dissertation are summarized and potential directions for future work are explored.

## 2. Contributions of the dissertation

The contributions of this dissertation are the following:

(1) Development of a stochastic hybrid systems model to model the activities of workers and robots involved in strawberry harvesting; human picking model calibration using data from commercial strawberry harvesting.

(2) Development of a mathematical model for optimal dynamic deterministic predictive scheduling in the context of co-robotic harvesting, with robots offering automated tray logistics. The model was solved using exact and approximate methods.

(3) Extensive simulation experiments that studied the effects of robot-to-picker ratio and prediction timeliness on scheduling performance, and the explanation of the resulting performance curves.

(4) Adaptation of an online dynamic scheduling solver for crop-transport robots under stochastic transport requests, with application in strawberry harvesting.

(5) Simulation experiments that investigated the effect of prediction uncertainty on the performance of the developed scheduling algorithm.

(6) Software and hardware development of a complex co-robotic harvesting system that includes picking carts, crop-transport robots, a communication system and operations software, and integration of parts into a fully functional system.

(7) Field experiments during commercial strawberry harvesting and assessment of system performance.

# Chapter 2   Deterministic predictive scheduling of robot team

## Notations

| | | | |
|---|---|---|---|
| $\boldsymbol{X_{p,k}}$ | state vector of $p^{th}$ picker at any time step k, in operating state $S_p$ | $v^w$ | the picker walks carrying a cart and tray to relocate (not pick), stochastic variables defined in section 6 |
| $S_p$ | $p^{th}$ picker's operation states | $S_r$ | $r^{th}$ picker's operation states |
| $x_{p,k}, y_{p,k}$ | x, y coordinate picker's location at any time step k | $v_r$ | robot speed: it is assumed to be constant in all operation states in Chapter 2 |
| $x_{r,k}, y_{r,k}$ | x, y coordinate robot's location at any time step k | $\mathcal{S}^{\mathcal{P}}$ | set of pickers in a crew |
| $W_{p,k}$ | gross weight of pickers' tray at any time step k | $\mathcal{S}^{\mathcal{F}}$ | set of robots in a team |
| $T_{p,k}$ | the elapsed time inside the current state of the $p^{th}$ picker | $L^s$ | collection station coordinate |
| $\hat{p}_{s_p}$ | fruit picking rate | $\mathcal{S}^{\mathcal{R}}$ | set of requests available to the scheduler |
| $\theta_{s_p}$ | picker's heading direction in state $s_p$ | $R_i$ | i$^{th}$ request inside $\mathcal{S}^{\mathcal{R}}$ |
| $\theta_{s_r}$ | robot's heading direction in state $s_p$ | $T_{r,k}$ | the elapsed time inside the current state of the $r^{th}$ robot |
| $v^p$ | moving speed of picker in the state of "PICK", stochastic variables defined in section 6 | $\boldsymbol{L}_i$ | predicted coordinates of the cart's (and picker's) locations |
| $\boldsymbol{L}_s$ | Coordinate of the active collection station | $\Delta t_i^f$ | the interval when the tray becomes full |
| $m^c$ | the capacity of the tray, assumed to be constant | $D_{si}$ | corresponding Manhattan distance from $\boldsymbol{L}^s$ to $\boldsymbol{L}_i$ along the path |
| $\Delta t_i^{pick}$ | the time spent in filling the ith tray (in state "PICK"), stochastic variables defined in section 6 | $t_{ki}^d$ | the dispatch instant of the robot |
| $\Delta t_i^u$ | one-way travel time for a robot from collection station to the full tray location | $\Delta t_{ki}^w$ | waiting time of picker at the full tray location |
| $t_{ki}^a$ | the instant when the robot arrive at the full tray location | $\Delta t^{UL}$ | idle time of robots at collection station, assumed to be constant |
| $\Delta t_i^p$ | The total time required by a robot to serve request $R_i$ and be available to serve another request | $\Delta t_k^A$ | the time interval when robot is available to be dispatched again |
| $t_{ki}^C$ | the completion time of service of request $R_i$ by robot $F_k$ | $\Delta t_i^r$ | release constraint of picker's request |
| $\Delta t^L$ | time interval of a picker loading the full tray on the robot, it is assumed to be a constant | $\boldsymbol{Z}$ | assignments of $\mathcal{S}^{\mathcal{R}}$ on $\mathcal{S}^{\mathcal{F}}$ |
| $t_{ki}^C$ | the completion time of service of request $R_i$ by robot $F_k$ | $\Delta t_i^{ef}$ | time interval of $i^{th}$ tray from empty to full (productive time) |
| FR | fill ratio of a tray | $\Delta t_i^{fe}$ | time interval of $i^{th}$ tray from full to empty (non-productive time) |
| $\Delta T^w$ | mean of wait time of the picker crew | $E_{ff}$ | mean of harvesting efficiency |
| $[p^r]$ | relative precision of estimated true mean | $\Delta T^{fe}$ | mean of nonproductive time of the picker crew |

# 1. Introduction

As explained in the previous chapter, dynamic predictive scheduling policies incorporate information about current and future demand into the scheduling. Two issues related to dynamic predictive scheduling are very important: *uncertainty* and *prediction timeliness.* In the context of harvesting, 'future demand' refers to the prediction of when and where a picker's currently used tray will become full, giving rise to a tray-transport request. Uncertainty will be always present in such prediction. It is known that uncertainty can be detrimental for scheduling algorithms that assume perfect information (Bertsimas & Ryzin, 2017; Blazewicz et al., 2019). Dynamic stochastic scheduling algorithms that incorporate prediction uncertainty can improve performance (Ichoua et al., 2006; van Hentenryck et al., 2010). However, to evaluate and compare such algorithms (existing and new ones) in the context of robot-aided harvesting, one needs to compare them against a baseline, i.e., the "best possible" situation. The best possible situation – from an uncertainty point of view - is 'deterministic predictive scheduling', i.e., when the scheduler knows exactly – without uncertainty – where and when the next tray-transport request will take place. Although there is no 'prediction' in deterministic predictive scheduling, the term is used in this dissertation to juxtapose it against 'stochastic predictive scheduling' under uncertainty, which is also addressed.

Prediction timeliness refers to how early a prediction becomes available to the dynamic scheduling algorithm before the actual request takes place. In the envisioned robot-aided harvesting approach, after a picker fills a part of the tray, she/he pushes a button on the cart to indicate that they want a robot to serve them., when they finish harvesting the tray. The scheduler does not have access to a prediction for this next-tray transport request before the button is pushed. The investigation of the effect of prediction timeliness on the performance of predictive scheduling is important. Khosro Anjom and Vougioukas (2019) showed that the earlier a tray-transport prediction is made – while a tray is being filled – the larger the uncertainty will be. Even if predictions are perfect, tray-transport request predictions arriving too late (too close to the actual request) will result in increased picker waiting times, because robots need time to travel to the location of the request.

The first goal of this chapter is to model the stochastic harvesting operation and implement an optimal deterministic predictive scheduling algorithm for tray-transport requests and study its performance for different robot-to-picker ratios. The second goal is to explore the effect of prediction timeliness on the mean waiting time of the pickers, under optimal deterministic predictive scheduling. To achieve these goals, a strawberry harvesting simulator was developed based on a hybrid system approach that models the harvest-related activities and motions of all agents (pickers and robots) involved in harvesting. This chapter is organized as follows: Section 2 describes a model for manual picking activities and robot tray-transport operations in strawberry harvesting. In Section 3, a mathematical model for predictive scheduling is developed, and in Section 4, an exact and an approximate algorithm to solve the model are reviewed and implemented. Section 5 briefly presents the simulator and the embedded tray-transport request and predictive scheduling modules. Section 6 presents the experimental design of our experiments and section 7 presents the results and their discussion. Finally, the main conclusions of this chapter are summarized in section 8.

## 2. Modeling of harvesting activity under deterministic request predictions

Strawberries are planted in rows with furrows between them that accommodate human and machine traffic (Figure 3.a). The field headlands are used for collection/parking/inspection stations and traffic of people, forklifts and trucks involved in the handling and transportation of the harvested crops. A typical harvesting block consists of approximate 80~120 rows, of about 100 meters in length. Before harvesting, collection stations and empty trays are placed at one side of the field, at the headland. The crew (say, $N$ people) start picking as a team in the first $N$ rows on the left or right side of the block. Although there may be several collection stations, the one that is closest to the crew is the active one. To reduce the walking needed to transport full trays to the collection station, the standard harvesting practice is to have workers start picking from the middle of the field block and walk outward, toward one of its edges. Once that section of the block is harvested, the collection stations are moved to the other edge, and the other section (half of the block) is harvested. This method essentially 'splits' a harvesting block into two sections (an example is given in Figure 3.b, where there is an upper and a lower block, above and below the blue dotted line, respectively). The field can be modeled using points at the edges of each furrow; two points to represent the middle line, and points for each collection station.

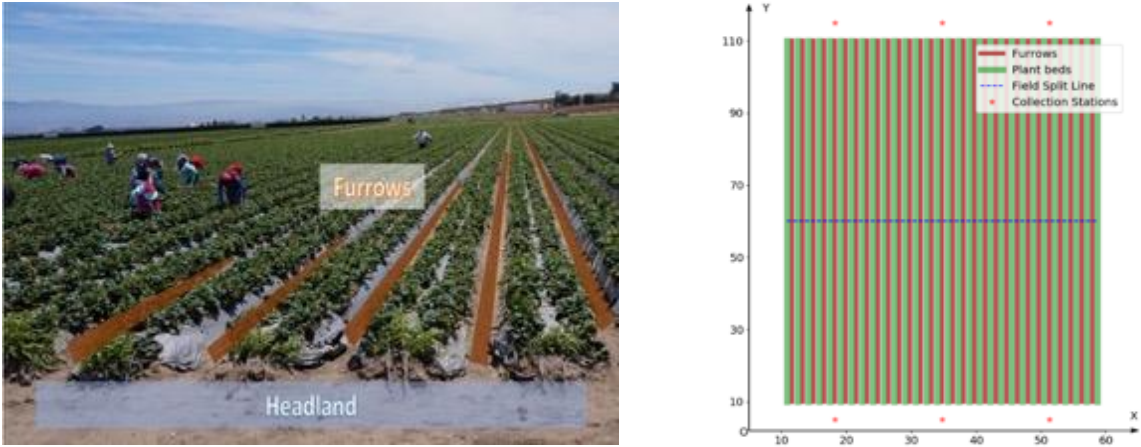

*Figure 3 - a) Layout of a typical raised-bed strawberry field; 3b) schematic figure of the strawberry harvesting field block with two sections (upper and lower); furrows; plant beds; field split line, and collection stations.*

In the envisioned robot-aided harvesting approach, after a picker starts filling a tray, she/he pushes a button on the cart to indicate that they want a robot to serve them. The push-button event is transmitted wirelessly to the server, where a software module predicts the location and time where the tray will become full (next tray transport request). The scheduling software module acknowledges the request (an LED lights up on the cart) and schedules a robot to go to the picker (in Chapter 3, requests may be rejected to guarantee that robot-aided efficiency is always better than all-manual harvesting efficiency). When the tray is full, the robot will either be there, or the picker must wait at that location for the robot to arrive. The picker will take an empty tray from the robot, place the full tray on it, push a button to send the robot back to the collection station, and resume picking. All robots wait at the active collection station and return to it after serving one picker, where they wait for the next dispatch command.

In our previous work (Seyyedhasani et al., 2020b a), a discrete-time hybrid system was developed to model and simulate the activities and motions of all agents involved in all-manual and robot-aided harvesting. A Finite State Machine (FSM) was utilized to model the discrete operating states/modes of the agents and the transitions between the operating modes. In this chapter, for the purposes of simulating predictive scheduling, the FSM was simplified (since all-manual harvesting was not modeled). The activities of a picker during robot-aided harvesting were classified into seven discrete operating states/modes (Table 1), and the operations of a tray-transport robot into eight states (Table 2). The operating states of pickers and robots, and the possible transitions amongst them are shown in Figure 4.

*Table 1- States defined to represent a picker's operating states during robot-aided harvesting*

| Operating state | Action |
|---|---|
| START | A picker leaves the collection station with an empty tray in hand, to start picking. |
| WALK_TO_FURROW_ENTRANCE | A picker walks in the headland, toward an empty (unoccupied) furrow. |
| WALK_TO_FURROW_SPLITLINE | A picker walks inside an empty (unoccupied) furrow until the field's split line is reached. |
| PICK | A picker is picking inside a furrow, with direction from the field split line toward the collection station. |
| WAIT_FOR_ROBOT_ARRIVAL | A picker waits (idle), with a full tray, for a robot to come. |
| EXCHANGE_TRAYS | A picker takes the empty tray brought by the robot and places a full tray on the robot. |
| STOP | A picker stops picking after reaching the end of the last row they harvested in the field block |

*Table 2 - States defined to represent a robot's operating states during robot-aided harvesting*

| Operating state | Action |
|---|---|
| START | Robot is idle at the collection station with no tray on it. |
| AVAILABLE | Robot with one empty tray on it is waiting at the collection station to be dispatched to a tray-transport request. |
| TRAVEL_TO_PICKER | Robot travels from collection station – carrying an empty tray – toward the location of the transport request. |
| WAIT_UNTIL_TRAY_FILLS | Robot arrives at the location of the tray-transport request and waits for the picker to finish harvesting. |
| EXCHANGE_TRAYS | Robot is idle while picker exchanges the empty tray with a full tray. |
| TRANSPORT_FULL-TRAY | Robot travels toward the collection station to deliver a full tray. |
| IDLE_IN_QUEUE | Robot with a full tray waits in a queue at the collection station to have its full tray unloaded, and an empty tray loaded. In this work we assume this tray exchange is fast and very few – if any – robots arrive together. |

| STOP | Robot stops its operation at the collection station after the last tray has been unloaded; end of harvesting of this block. |
| --- | --- |

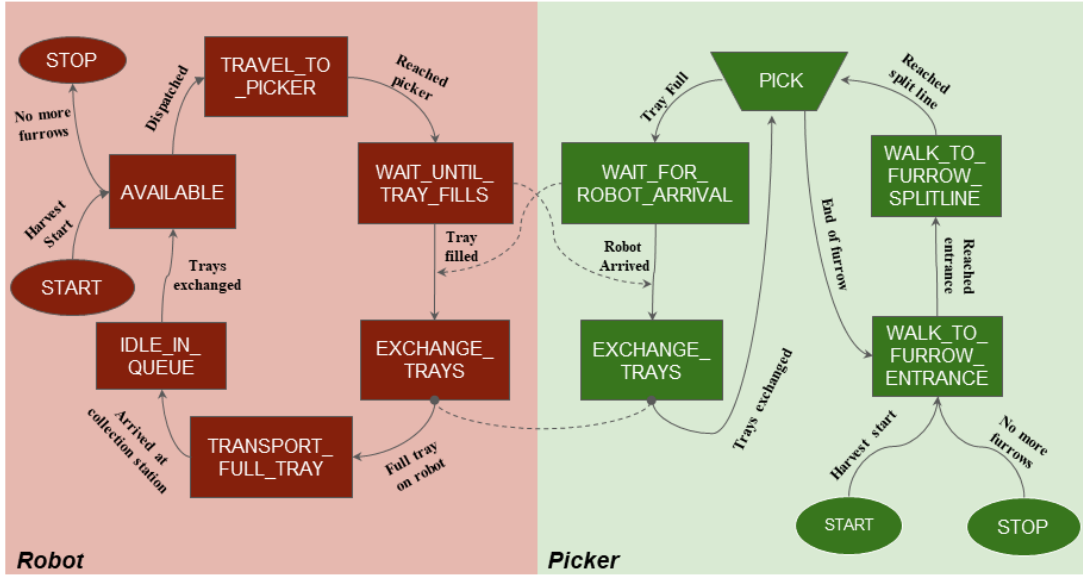

*Figure 4. Two coupled finite state machines model human picker and robot operations during robot-aided harvesting, where robots transport full and empty trays*

Following our previous work (Seyyedhasani et al., 2020b a) the operation of the $p^{th}$ picker at any time step k, in operating state $S_p$ is represented by a state vector, $\boldsymbol{X_{p,k}} = \left(x_{p,k}, y_{p,k}, W_{p,k}, T_{p,k}\right)$ that contains as state variables the picker position coordinates $x_{p,k}, y_{p,k}$; the gross weight $W_{p,k}$ of the pickers' tray, and the elapsed time $T_{p,k}$ inside the current state. The coordinate system is defined in Figure 3.b above. The variables are updated with state-dependent difference equations that model picker kinematics (Eq 1), crop harvesting (Eq 3) and elapsed time (Eq 4):

$$x_{p,k+1} = x_{p,k} + \Delta t \, V_{S_p} cos\theta_{S_p} \qquad \text{(Eq 1)}$$

$$y_{p,k+1} = x_{p,k} + \Delta t \, V_{S_p} \, sin\theta_{S_p} \qquad \text{(Eq 2)}$$

$$W_{p,k+1} = W_{p,k} + \Delta t \, \hat{p}_{S_p} \qquad \text{(Eq 3)}$$

$$T_{p,k+1} = T_{p,k} + \Delta t \qquad \text{(Eq 4)}$$

In state $S_p$, $V_{S_p}$ is the pickers' travel speed, $\theta_{S_p}$ is the picker's heading direction, and $\hat{p}_{S_p}$, is the fruit picking rate. The parameters are assumed constant within $S_p$, i.e., fixed values are used over the duration of the state. Parameters $V_{S_p}$, $\hat{p}_{S_p}$ depend on individual picker performance, yield, and external factors that cannot be modeled (e.g., answering a phone call, stretching), and thus, their values are modeled as stochastic variables. The heading $\theta_{S_p}$ is set by the direction of the furrow or the headland that the picker travels in, depending on the state. The walking speed $V_{S_p}$ is the same inside states "WALK_TO_FURROW_ENTRANCE" and "WALK_TO_FURROW_SPLITLINE", because the picker walks carrying a cart and tray to relocate (not pick), and is equal to a value $v^w$; inside state "PICK", $V_{S_p}$ is equal to a slow moving speed, $v^p$ . $V_{S_p}$ is equal to zero in operating states "WAIT_FOR_ROBOT_ARRIVAL" and "EXCHANGE_TRAYS". If the capacity of the tray is $m^c$, and the time spent in filling the ith tray (in state "PICK") is $\Delta t_i^{pick}$, the picking rate, $\hat{p}_{S_p}$ in the "PICK" state can be estimated as $\hat{p}_{S_p} = m^c/\Delta t_i^{pick}$. Distributions for stochastic parameters $v^w$, $v^p$ and $\Delta t_i^{pick}$ can be estimated experimentally (see Section 7.1).

At any time step k, the operation of the $r^{th}$ robot in operating state $S_r$ is represented by a continuous state vector, $\boldsymbol{X_{r,k}} = (x_{r,k}, y_{r,k}, T_{r,k})$, that contains the robot's position coordinates $x_{r,k}, y_{r,k}$ and the elapsed time $T_{r,k}$ inside the current operating state. Simple robot kinematics are modeled by (Eq 1), (Eq 2), where $\theta_{S_r}$ is the robot's moving direction; time is updated using (Eq

4) ( "r" is used instead of "p" in the subscripts). Robot speed $V_{S_r}$ is equal to a parameter $v_r$ that is assumed constant in all operating states in which the robot is travelling. The robot's heading is the same as the direction of the headland or the furrow. In this work we assume that the time duration of state "IDLE_IN_QUEUE" is constant, i.e., the unloading of the full tray and loading of the empty tray is very fast and very few – if any – robots arrive together. The time duration of the state "EXCHANGE_TRAYS" is also assumed to be constant.

## 3. Deterministic predictive scheduling of harvest-aid robots

Deterministic predictive scheduling of a team of crop-transport robots is a variant of the well-known Capacitated Vehicle Routing Problem (CVRP) (Vougioukas et al., 2012). However, in the current system, each robot can only serve one tray-transport request, with each robot route starting and ending at the collection station. Hence, the scheduling of these crop-transport robots can be modeled as a parallel machine scheduling problem (PMSP) with the objective of minimizing the mean of waiting times of all transport requests (Lawler et al., 1993). Essentially, servicing a tray-transport request with one robot is the equivalent of executing one individual job by one machine in PMSP; robots correspond to independent parallel machines. The optimal criterion of the scheduling system is to minimize the mean waiting time, $\Delta T^w$, of all tray-transport requests.

In the envisioned robot-aided strawberry harvesting operation, each picker from a set $\mathcal{S}^{\mathcal{P}} = \{P_1, \ P_2, \ \dots, P_Q\}$ of Q pickers places harvested fruits in a tray that lies on a picking cart. When the tray fills, a robot $F_k$ from a team of M identical transport robots $\mathcal{S}^{\mathcal{F}} = \{F_1, \ F_2, \ \dots, F_M\}$ brings an empty tray to the picker and carries the full tray to a collection station; the station's coordinates $L^s$ are known at each time instant. The robot scheduling algorithm has access to a set of predicted tray-transport requests $\mathcal{S}^{\mathcal{R}} = \{R_1, \ R_2, \ \dots, R_N\}$, where $0 \leq N \leq Q$ and computes an updated schedule every time a new predicted request enters $\mathcal{S}^{\mathcal{R}}$ and there is a robot available to be dispatched. The dispatching of a robot is non preemptive which means that the scheduler does not change the dispatching decision after the robot starts executing the dispatching command.

Let us assume that at some time $t$, a new predicted request $R_i$ is generated and entered into $\mathcal{S}^{\mathcal{R}}$. $R_i$ contains the following information: (1) a prediction of the remaining time interval $\Delta t_i^f$ (with respect to t) until the tray becomes full with fruit and ready to be transported, and (2) the predicted coordinates $\boldsymbol{L}_i$ of the cart's (and picker's) locations at time $t_i^f$ when the tray becomes full ($t_i^f = t + \Delta t_i^f$) The travel time required for a robot to travel from the collection station location at $\boldsymbol{L}^s$ to $\boldsymbol{L}_i$ – and back – is computed by approximating the path by one straight line segment on the headland and another straight line segment inside the furrow that corresponds to $\boldsymbol{L}_i$ (Figure 5). The one-way traveled distance $\mathrm{D}_{si}$ is the corresponding Manhattan distance from $\boldsymbol{L}^s$ to $\boldsymbol{L}_i$ along the path. The corresponding one-way travel time is $\Delta t_i^u = D_{si}/v^r$, where $v^r$ is the robot's velocity (assumed to be constant and the same for all robots).

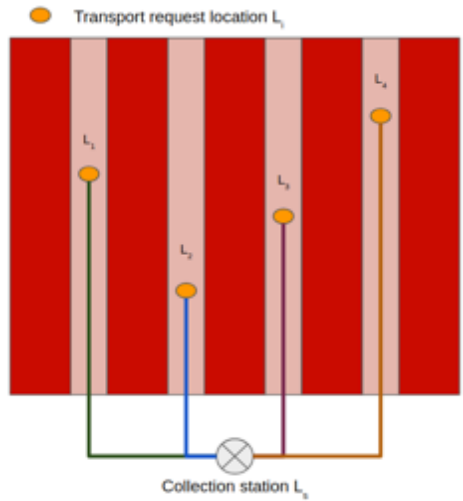

Let us now assume that the scheduling algorithm selects robot $F_k$ to serve some request Ri and dispatches the robot at some time $t_{ki}^d$ (Figure 6). The robot will arrive at L$_i$ at time instant $t_{ki}^a$:

$$t_{ki}^a = t_{ki}^d + \Delta t_i^u \qquad \text{(Eq 5)}$$

Then, the waiting time for request Ri to be served by robot $F_k$ will be $\Delta t_{ki}^w$:

$$\Delta t_{ki}^w = ma\,x(t_{ki}^a - t_i^f, 0) \qquad \text{(Eq 6)}$$

The reason for using the 'maximum' operator is because the scheduling is predictive, and therefore, it is possible that robot $F_k$ arrives at $\boldsymbol{L_i}$ before the predicted tray fill-up time, $t_i^f$; in this case, there is no waiting time, i.e., $\Delta t_{ki}^w$ is zero. Once the tray is full and the robot is at $\boldsymbol{L_i}$ the picker will take the empty tray from the robot and will load the full tray on the robot (and then resume picking). This corresponding time is assumed constant and equal to $\Delta t^L$. Next, the robot will transport the tray back to the collection station; the required time will be $\Delta t_i^u$ When the robot arrives at the collection station it will take $\Delta t^{UL}$ seconds until all previously arrived robots in the queue are served, the full tray it carries is unloaded and an empty tray is loaded on it, so that it is ready to be dispatched again; $\Delta t^{UL}$ is assumed to be a reasonable constant based on idle time of pickers at collection station in the scenario of manual harvesting. The total time, $\Delta t_i^p$, required by a robot to serve request $R_i$ and be available to serve another request is shown in (Eq 7). So, the completion time of service of request Ri by robot $F_k$, is the time instant is (Eq 8).

$$\Delta t_i^p = 2\Delta t_i^u + \Delta t^L + \Delta t^{UL} \qquad \text{(Eq 7)}$$

$$t_{ki}^C = t_{ki}^d + \Delta t_i^p \qquad \text{(Eq 8)}$$

At any time $t$, robot $F_k$ is either serving some request $R_i$ or is idle. In the former case, the robot is available to be dispatched again, after a time interval $\Delta t_k^A$ (Eq 9) that is needed to serve request $R_{i-1}$.

$$\Delta t_k^A = max\big(0, \ t_{k(i-1)}^C - t\big) \qquad \text{(Eq 9)}$$

If it is idle at the collection station, it is available to be dispatched immediately ($\Delta t_k^A = 0$).

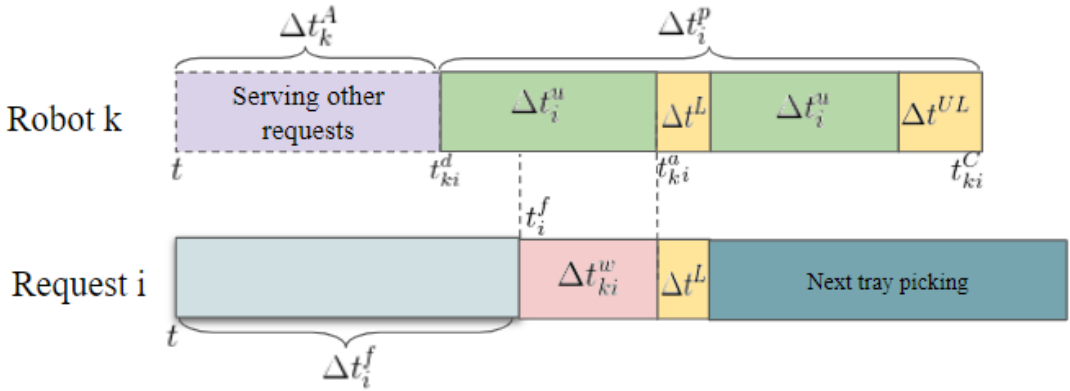

*Figure 6 - Timelines of request $R_i$ served by robot $F_k$*

Let us now focus on the robot dispatch time $t_{ki}^d$. Scheduling becomes more effective as the number of requests in $\mathcal{S}^{\mathcal{R}}$ is larger (Lu et al., 2003). If the robot dispatch time is delayed by some amount $\Delta t_i^r$ (also referred to as 'release delay', i.e., $t_{ki}^d \geq t + \Delta t_i^r$), each new request that may arrive during $\Delta t_i^r$ will cause an updated schedule to be computed. However, too long a release delay may increase the request waiting time, as the robot could depart too late. To select $\Delta t_i^r$, one can note that if the robot arrives early at $L_i$ ($t_{ki}^a < t_i^f$), it will have to wait idle until the tray is full. The greatest value for $\Delta t_i^r$ that eliminates robot idle time at $L_i^f$ is:

$$\Delta t_i^r = max(\Delta t_i^f - \Delta t_i^u, 0) \qquad \text{(Eq 10)}$$

The introduction of the release delay ensures that the robot will not arrive at $\boldsymbol{L}_i$ before $\Delta t_i^f$ and the calculation of picker waiting time (Eq 7) can be simplified as $\Delta t_{ki}^w = t_{ki}^a - t_i^f$. By combining (Eq 7), (Eq 8), (Eq 9), (Eq 10), $\Delta t_{ki}^w$ can be expressed as Eq (11).

$$\Delta t_{ki}^w = t_{ki}^a - t_i^f = t_{ki}^C - \Delta t_i^p + \Delta t_i^u - \left(t + \Delta t_i^f\right) = t_{ki}^C - \Delta t_i^p - \Delta t_i^r - t \qquad \text{(Eq 11)}$$

The timelines integrating all temporal items in a schedule of request $R_i$ served by $F_k$ are shown in Figure 7.

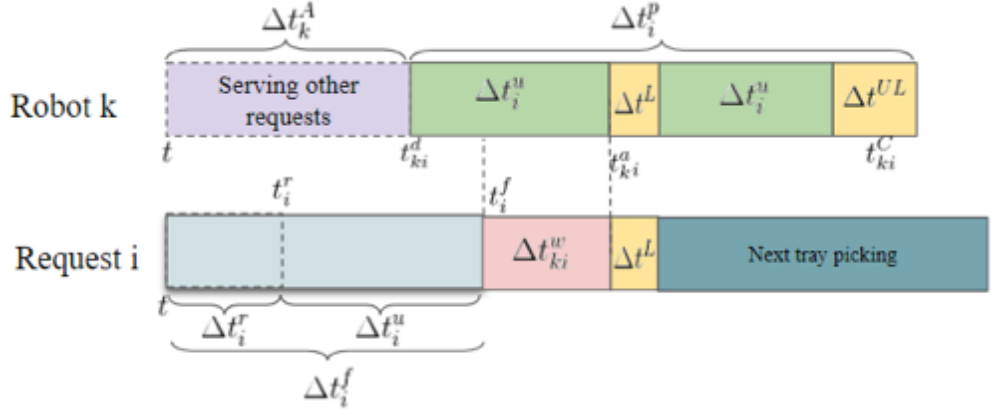

*Figure 7 - Timelines of request $R_i$ served by robot $\boldsymbol{F_k}$*

At any time $t$, $\Delta t_i^p$ and $\Delta t_i^r$ are obtained from $\left(\boldsymbol{L}_i, \Delta t_i^f\right)$, which are independent from the schedule. Hence, minimizing the sum of the waiting times of all generated requests $(\sum_1^M \sum_1^N \Delta t_{ki}^w)$ is equivalent to minimizing the sum of completion times, $\sum_1^M \sum_1^N t_{ki}^C$ of all requests in $\mathcal{S}^{\mathcal{R}}$. A feasible schedule solution includes two parts: the assignments of all generated requests $\mathcal{S}^{\mathcal{R}}$ to robots $\mathcal{S}^{\mathcal{F}}$, and the schedules of each robot $F_k$. An $M \times N$ matrix $\mathbf{Z}$ represents the assignments of $\mathcal{S}^{\mathcal{R}}$ on $\mathcal{S}^{\mathcal{F}}$. Each element $z_{ki}$ of $\mathbf{Z}$ is 1 if request $R_i$ is served by robot $F_k$; otherwise, $z_{ki}$ is 0. The schedule for each robot $F_k$ is composed of a list of dispatching tuples. Each dispatching tuple includes the dispatching location $\boldsymbol{L}_i$ and the dispatching time $t_{ki}^d$.

The mathematical model of predictive scheduling of crop-transport robots is expressed with equations (11-18), which define a minimization problem with associated constraints:

(1) Schedules should consider all the generated requests $\mathcal{S}^{\mathcal{R}}$ (Eq 12)

(2) Each request is assigned only to one robot (Eq 13).

(3) The robot is not dispatched until the release constraint of scheduled request (Eq 16).

(4) The robots start from the active collection station and end with the same one in the planning schedule (Eq 17).

(5) Each robot serves only one picker when dispatched, and no preemption is allowed, i.e., a robot dispatched to a request $R_i$ cannot be re-assigned to another request $R_j$ (Eq 19).

$$min \sum_{k=1}^{M} \sum_{i=1}^{N} t_{ki}^C$$
$$s.t.$$

$$\sum_{k=1}^{M} \sum_{i=1}^{N} z_{ki} = N \qquad \text{(Eq 12)}$$

$$\sum_{k=1}^{M} z_{ki} = 1 \qquad \text{(Eq 13)}$$

$$t_{ki}^C = t_{ki}^d + \Delta t_i^p \qquad \text{(Eq 14)}$$

$$\Delta t_i^r = max(\Delta t_i^f - \Delta t_i^u, 0) \qquad \text{(Eq 15)}$$

$$t_{ki}^d \geq t + \Delta t_i^r + \Delta t_k^A \qquad \text{(Eq 16)}$$

$$\Delta t_i^p = 2\Delta t_i^u + \Delta t^L + \Delta t^{UL} \qquad \text{(Eq 17)}$$

$$\Delta t_i^u = \frac{D_i}{v_r} \qquad \text{(Eq 18)}$$

$$\forall R_i, R_j \in \mathcal{S}_{\mathcal{R}}, \text{ and } i \neq j, [t_{ki}^C - \Delta t_i^p, t_{ki}^C) \cup [t_{kj}^C - \Delta t_j^p, t_{kj}^C) = \emptyset \qquad \text{(Eq 19)}$$

# 4. Exact and approximate solution methods

The above formulated parallel machines scheduling problem with release dates (Nessah et al., 2007, 2008) is a variant of the well-known job shop problem (JSP) (Lawler et al., 1993). Following symbol notations defined by Lawler et al. (1993), the problem can be expressed as $Pm|r_i|\sum C_i$ where $Pm$ represents identical parallel machines, $r_i$ means that the $i^{\text{th}}$ job cannot be processed until its release time, and $\sum C_i$ represents that the objective criterion is to minimize the sum of the completion times of all jobs. Meanwhile, a sub-optimal - but faster - approximation algorithm was also implemented because scheduling must be performed in the field with limited computational resources, and deterministic scheduling is an important component of stochastic predictive algorithms that incorporate uncertainty, such as the scenario-based method (R. W. Bent & Van Hentenryck, 2004). The branch-and-bound (BAB) algorithm converges to the global optimal solution of each modelled problem, while the efficient approximate algorithm, converges to a sub-optimal solution; both were applied and compared. The scheduling algorithms were implemented in C++ and a Python wrapper (Cython) was used to package the algorithm into a Python version that was embedded in the simulator.

## 4.1. Exact branch and bound (BAB)

The BAB algorithm computes a systematic enumeration of all the candidate solutions by creating a tree, using state-space search (Land & Doig, 1960). The root of the tree represents the full set of solutions. BAB explores branches of this tree, which represent subsets of the solution set. It uses estimated upper and lower bounds of the optimal solution along with the value of the best solution found so far to "prune" branches of the tree that will result in suboptimal solutions. Hence, the efficiency of the BAB algorithm depends on the accurate and efficient estimation of the lower bound of the optimal solution.

In this chapter, two lower bounds were applied by relaxing certain constraints (Nessah et al., 2007). One lower bound was obtained by relaxing the release constraints of jobs. When the release time of each job is removed, all jobs can be processed immediately. The optimal solution to this no-release-constraint problem was optimally solvable by the policy of Shortest Processing Time (SPT) (Baker & Trietsch, 2013), which means that the machines serve jobs with the shortest process time first. The second lower bound was achieved by allowing job splitting and preemptive dispatching, which converted the problem to $Pm|pmtn, r_i| \sum C_i$ , where "*pmtn*" means that preemptive processing is allowed in the scheduling). The optimal solution to this problem was achieved by the policy of the Shortest Remaining Processing Time (SRPT): at any time, a released job with the shortest remaining process time is simultaneously processed on all the available machines. The processing is interrupted if another job becomes available with a processing time strictly shorter than the remaining processing time of the job in the process.

### 4.2. SRPT-convert approximation algorithm

Approximation algorithms compute approximate solutions to NP-hard (Non-deterministic polynomial-time hardness) problems with provable guarantees on the distance – given some metric - of the returned solution to the optimal one (Williamson & Shmoys, 2011). The approximation of such algorithms is always guaranteed to be within a multiplicative or additive factor of the optimal solution even in the worst cases. In this work, the efficient CONVERT algorithm was used (Phillips et al., 1998) to solve the original $Pm|r_i| \sum C_i$ problem. CONVERT firstly relaxes the original problem by allowing preemption and job-splitting, thus turning it into a $Pm|pmtn, r_i| \sum C_i$ problem, which can be solved in polynomial time. SRPT-CONVERT was proven to be guarantee 6-approximate algorithm (Phillips et al., 1998). Then, the solution of the relaxed problem is adjusted, to generate a solution to the original problem.

# 5. Implementation of harvest simulator

Simulator software was developed to simulate robot-aided strawberry harvesting based on the hybrid systems model presented in Section 2. This simulator constitutes a significant extension and adaptation of the simulator developed by Seyyedhasani et al., (2020a) to incorporate the predictions of requests and the predictive scheduling of robots. The architecture of the simulator is shown in Figure 8. The simulator is initialized with the geometrical description of the strawberry field (furrow endpoints, split line, collection station locations), the picking crew and robot team parameters, and the initial locations of pickers, robots and active collection station. Modules for "Picker operations" and "Robot operations" implement the coupled hybrid system models of the pickers and robots, respectively. A "Crop, crew & collection station distribution" module updates the harvest status of each furrow (harvested/unharvested/currently harvesting) and the status of the active collection station. It also calculates the sequence of furrows picked by the crew (after harvesting from a furrow, a picker moves to the furrow of the closest unharvested bed). During simulated harvesting, the picker states are input to the "Tray-transport request prediction" module that computes predicted transport requests, and the predicted requests and current robot states are used by the "Predictive scheduling" module to compute robot schedules and output dispatch commands to the robots.

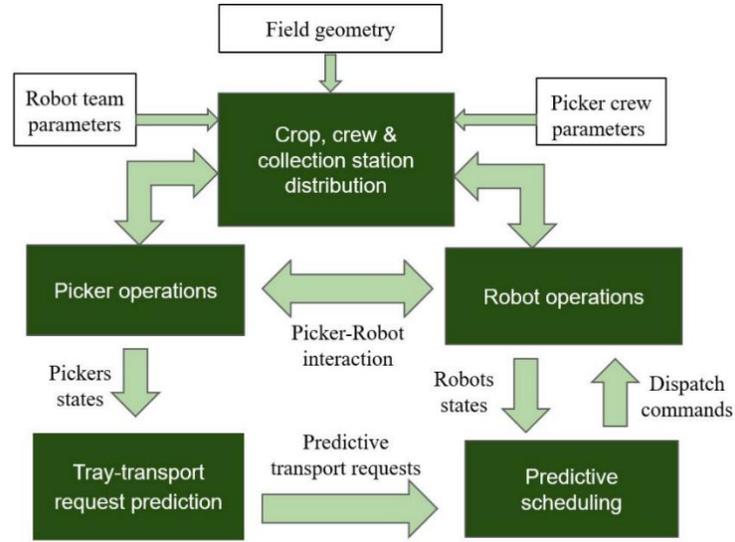

*Figure 8 - Architecture of integrated harvesting simulator and predictive scheduling system*

The simulator uses a global time variable t to represent the current time of the harvesting activity; time starts at t = 0 s and increases by $\Delta t$ (0.5s was used). Before the start of fruit harvest for each tray, the harvesting parameters ($v_p, v_w, \Delta t^{pick}$) are sampled randomly using their respective experimentally derived frequency histograms. The states of pickers and robots are updated at each time step and the simulation terminates when the entire field block is harvested. There are four pre-allocated collection stations on each half of the split block, and at any point in time, only one collection station is active (the one closest to the crew). This procedure is standard practice in commercial harvesting and reflects what was done when the harvesting data were collected.

Figure 9 presents a time-lapse image of the visualization of the robot-aided harvesting process. The currently active collection station is shown as a red star symbol. The locations where trays became full (tray-transport request locations) are marked as green hexagons and are exactly the same as the robot pickup locations (marked as golden crosses), since request predictions are perfect.

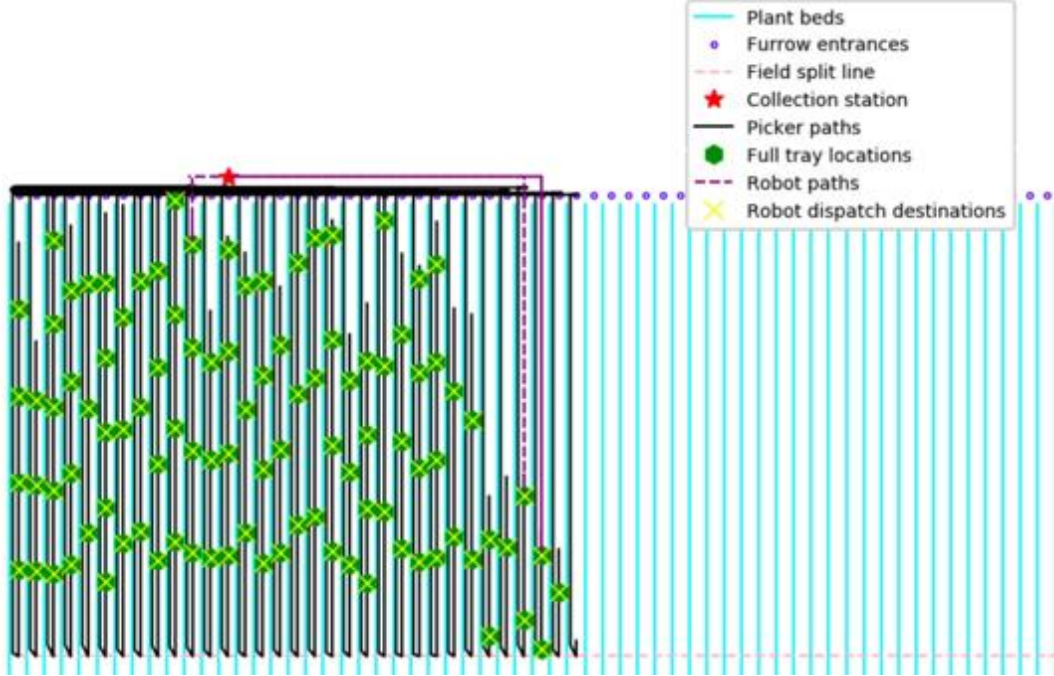

*Figure 9 Visualization of robot-aided harvesting simulator in a commercial strawberry field*

## 5.1. Tray-transport request prediction module

In the simulator, the time $\Delta t_k$ of the tray-transport request (when tray fills up) and moving velocity $v_{s_p}$ of the picker was accessible to the scheduling module after a predicting time is reached. The location of the tray-transport request is predicted as $\Delta t_k \times v_{s_p}$; if the cart will not become full by the end of the current furrow, the prediction is not inserted in the request set $\mathcal{S}^{\mathcal{R}}$ because it is not possible to predict the next furrow the picker will enter to continue picking and filling the tray. The fill-ratio (FR) is defined as the current weight of a tray divided by the tray's maximum weight (capacity); it is zero when the tray is empty and one when the tray becomes full. To study the effect of the timing of the availability of request predictions to the scheduler, the tray fill-ratio (FR) was used as a proxy of the prediction timeliness. The reason that weight was used instead of time is that in real-world operation, the time to fill a tray is unknown until the tray becomes full. However, tray capacity is known in advance (trays are standard), and the weight of each tray is transmitted to the robot scheduling system in real-time; hence, FR can be computed before the tray is full.

In the simulator, the tray-transport request prediction module has access to the pickers' states at each timestamp, so the location and time of the tray-transport request can be predicted perfectly as soon as the pickers start picking (at FR=0). In the real world, perfect "prediction" is only possible when the tray becomes full (FR=1), which is equivalent to reactive scheduling, since in that case the robots are only scheduled when the tray becomes full. To evaluate the effect of prediction timeliness on predictive scheduling, the same harvesting operations were simulated for FR values ranging from zero to one.

### 5.2. Predictive scheduling module

At each time step, the predictive scheduling module receives and checks the updated robot states and tray transport requests. The scheduling algorithm is executed only when there are robots available in the collection station and new predictive transport requests enter the tray-transport request set $\mathcal{S}^{\mathcal{R}}$. The scheduling module will solve the modeled problem and store the computed schedules in a schedule table. Each schedule consists of the available robot index $F_k$, the request location $\boldsymbol{L}_i$, and the dispatch time $t_{ki}^d$ ; the dispatch command is sent to the scheduled robot when actual time equals dispatch time. After a dispatching command is sent, the corresponding request and schedule are removed from $\mathcal{S}^{\mathcal{R}}$ and from the schedule table, respectively.

## 6. Experimental design

The simulation experiments were designed to investigate the effects of the configuration parameters on the predictive scheduling. We defined three evaluation metrics for the scheduling performance in Section 7.1: mean waiting time, mean non-productive time, and mean harvesting efficiency of collected trays. The factorial parameters for the predictive scheduling include: the number of robots, the robot speeds, FRs and scheduling policy applied (BAB search or SRPT). The dimension of the harvested block was fixed, and consisted of 100 furrows spaced 1.65 meters apart and furrows of length equal to 100 m. The size of the picking crew was 25 pickers, and there was one additional worker at the collection station. In the simulator, the distributions of three stochastic parameters ($v^w$, $v^p$, $\Delta t^{pick}$) for manual harvesting were generated from data collected in a commercial strawberry field during high-yield season (Seyyedhasani et al., 2020b).

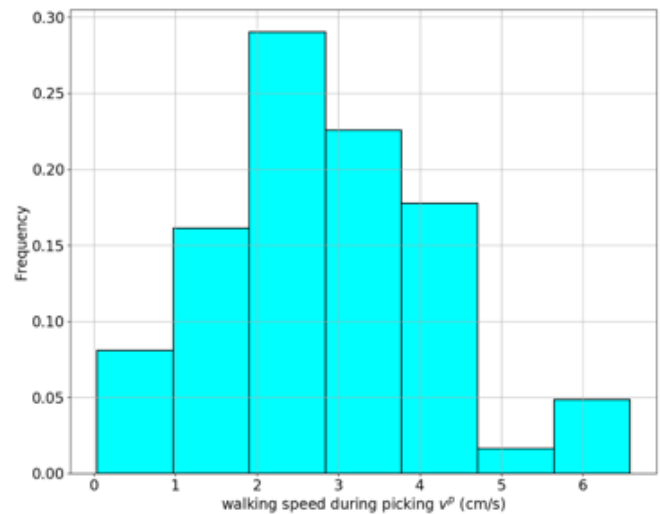

*Figure 10. Frequency histogram of picker walking speed ($v^p$) during picking (inside state "PICK").*

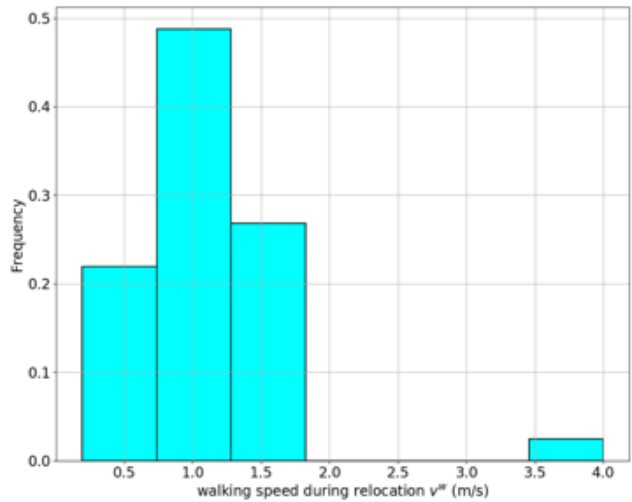

*Figure 11. Frequency histogram of picker walking speed ($v^w$) while picker relocates carrying the tray and cart (in states "WALK_TO_FURROW_ENTRANCE" and "WALK_TO_FURROW_SPLITLINE").*

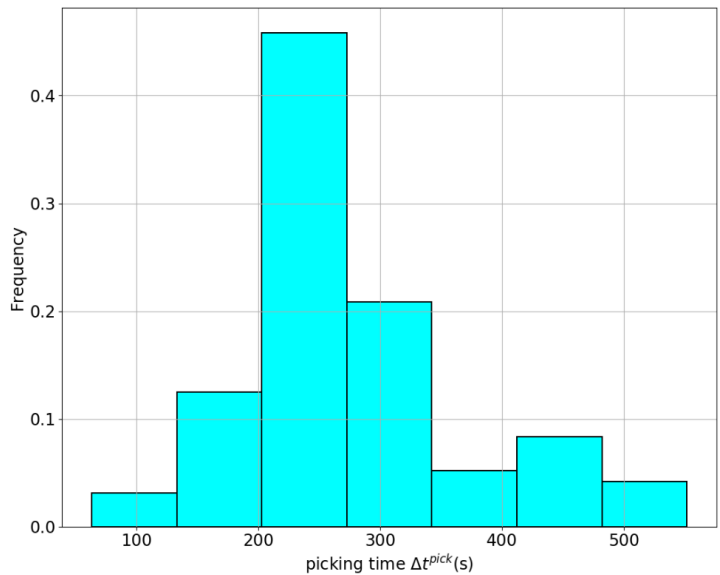

*Figure 12. Frequency histogram of the pickers' picking time ($\Delta t^{pick}$).*

The above distributions were derived experimentally (frequency histograms) and represent the specific combination of season, field, crew and crop conditions. The methodology itself can be applied to any distributions.

A Monte Carlo sampling approach was applied to compute the mean and standard deviation of the waiting time and the non-productive time given the distributions of the worker stochastic variables ($v^w$, $v^p$, $\Delta t^{pick}$). These distributions were sampled at the start of each tray harvesting. Given the sampled values, the states of pickers were updated with the model equations defined in Eq (1,2,3,4). The measurement of each Monte-Carlo experiment is referred to as a harvesting simulation for the whole field block with the size mentioned above. Approximately 360~400 trays were harvested in the modeled block. The number of Monte Carlo experiments was determined based on the metrics of relative precision introduced in section 6.2. The selected configuration parameters to estimate the relative precision were: number of robots equal to 8, robot speed equal to 1.5 m/s and FR equal to 0.5. The hypothesis was that the relative precision did not vary too much for other harvesting configurations given the same number of Monte-Carlo experiments. We applied this number for the evaluation of all simulation experiments.

The performance of the predictive scheduling was evaluated under different configuring parameters. In section 7.2, two scheduling policies were compared in the metrics of mean waiting time and policy computation time under a harvesting configuration of 8 robots, 4 levels of FRs and robot speed at 1.5 m/s. In section 7.3, the robot speed was set as 1.5 m/s and the scheduling performance versus 4 levels of FRs, and 7 different robot numbers was investigated. In section 7.4, the robot speed was set as 1.5 m/s and the scheduling performance versus 8 levels of FRs, and 4 different robot numbers was measured. In section 7.5, the effect of robot speed on the scheduling performance was investigated given 6 robots and different FRs. Grid combinations of 5 different speeds and 13 levels of FR were examined.

## 6.1. Evaluation metrics

In both the all-manual and robot-aided strawberry harvesting trials, the productive time per tray – denoted as $\Delta t_i^{ef}$ – is defined in the same way: it is the time required by a picker to fill the ith tray to its capacity, starting from an empty tray. Productive time includes picking and walking to relocate to a new furrow to resume picking when the tray cannot be finished in the current furrow. Non-productive time per tray – denoted as $\Delta t_i^{fe}$ – is defined as the time interval that is not spent picking or relocating to pick from another furrow. In manual strawberry harvesting, $\Delta t_i^{fe}$ includes the picker's walking time to transport the full tray to the unloading station, the waiting time in a queue to deliver the tray and get an empty one, and the walking time required to return to the previous position to resume picking. In contrast, in robot-aided harvesting, $\Delta t_i^{fe}$ is the sum of the time ($\Delta t_i^{w}$) the picker spends waiting for a robot to arrive in state "WAIT_FOR_ROBOT_ARRIVAL", plus the time ($\Delta t^L$) needed to place the full tray on the robot and take an empty tray from the robot, in state "EXCHANGE_TRAYS". $\Delta t_i^{w}$ is highly dependent on the robot scheduling policy, whereas $\Delta t^L$ is small and is assumed to be constant. $\Delta T^w$ was the average waiting time and $\Delta T^{fe}$ represents the average non-productive time of all the trays measured in a simulation experiment.

The mean harvesting efficiency, $E_{ff}$, when harvesting N trays with or without robots, is defined as the averaged sum of ratios of productive time over total time spent for each tray; it is calculated by (Eq 20):

$$E_{ff} = \frac{1}{N} \sum_{i=1}^{N} \frac{\Delta t_i^{ef}}{\Delta t_i^{ef} + \Delta t_i^{fe}} \qquad \text{(Eq 20)}$$

$\Delta T^{fe}$ and $E_{ff}$ can be used to evaluate the overall performance of all-manual and robot-aided harvesting, and $\Delta T^w$ can be used to evaluate the performance of the scheduling algorithm.

## 6.2. Determination of number of Monte Carlo runs

Since the parameters ($v^w$, $v^p$, $\Delta t^{pick}$) of the picker operating states are stochastic variables, harvesting was simulated using a Monte-Carlo approach. The parameters are sampled - before the start of each tray picking - from the experimentally derived frequency histograms that approximate the respective probability distributions (Figure 12-15). Next, the number M of Monte-Carlo repetitions that are needed to achieve a given value of desired relative precision for the mean waiting time will be derived (Figliola & Beasley, 1995).

Let a picker's waiting time for the $i^{th}$ tray be $\Delta t_i^w$, and let $\overline{\Delta t}^w$ be the sample mean waiting time for all N trays; $\overline{\Delta t}^w$ is predicted by one execution of the harvest simulation. The variation in the sample statistics is characterized by a normal distribution of the sample mean values about the true mean. $\overline{\Delta t}^w$ will be different each time the harvest simulation is executed ($\overline{\Delta t}^w$ is a random variable). If harvesting is simulated M times, a pool of $\sum_{i=1}^{M} N_i$ measurements will be available. The pooled standard deviation of $\langle \overline{\Delta t}^w \rangle$ is expressed as $\left[ S^{\overline{\Delta t_w}} \right]$ in Eq (26). $\left[ S^{\overline{\Delta t_w}} \right]$ represents the absolute precision of the estimation of the true mean $\Delta T^w$:

Finally, the relative precision $[p^r]$ of our estimation of the true mean is given by (Eq 21).

$$[p^r] = \left. \left[ S^{\overline{\Delta t_w}} \right] \middle/ \langle \overline{\Delta t}^w \rangle \right. \qquad \text{(Eq 21)}$$

In this work, a relative precision equal to 1% was deemed adequate. The number M of Monte-Carlo repetitions to achieve 1% relative precision was determined experimentally and is reported in Section 7.1.

# 7. Experimental results and discussion

## 7.1. Determination of sufficient number of Monte Carlo runs

As explained in Section 6.2, when more Monte-Carlo runs are executed for the same harvesting scenario, better precision can be achieved for the estimate of true mean $\langle \overline{\Delta t^w} \rangle$. To quantify the relative precision given different times of measurements, one harvesting scenario was simulated that involved six robots, FR=0.5, robot velocity at 1.5m/s, and the SRPT-CONVERT scheduling algorithm. The relative precision $[p^r]$ was computed as a function of increasing number of Monte-Carlo runs. Figure 13, shows that 100 executions resulted in a relative precision of approximately 1%, which corresponds to 0.081s of absolute precision, given that $\langle \overline{\Delta t^w} \rangle$ was 8.1s. Therefore, the evaluated metrics ($\Delta T^w$, $\Delta T^{fe}$) in all our experiments were estimated based on the pooled means of M=100 Monte-Carlo runs. We hypothesized that the 100 sampled means of each evaluated metric were normally distributed in different group of setting parameters.

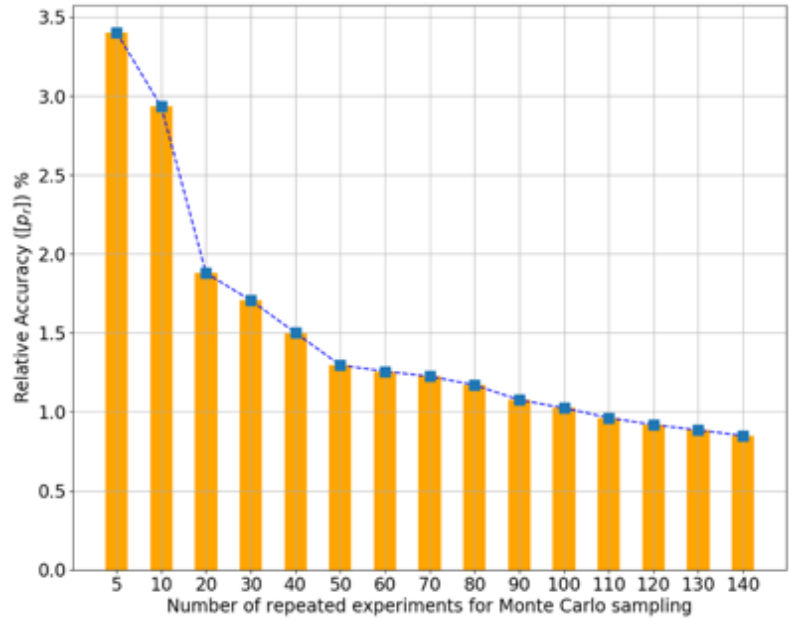

*Figure 13. Relative precision of the estimated mean waiting time $\overline{\Delta t^w}$ as a function of the number of simulation repetitions for Monte-Carlo sampling*

## 7.2. Comparison of exact and approximate Solutions

The approximate (SRPT) and the exact algorithm (BAB search) were used to compute the average scheduling time for FRs equal to 0.3, 0.5, 0.7, 0.8, 0.9. The resulting picker average waiting time was calculated by the total time of executing the scheduling function divided by the number of times running that function. Table 3 shows the relative discrepancy of the approximate solutions compared to the exact ones and the corresponding resulting picker average waiting time for the two algorithms. The largest discrepancy was 6.8%, and in terms of absolute value, it corresponded to a difference of 1.5 seconds for the evaluated average waiting time. However, the BAB algorithm required long computation times and was 110 to 240 times slower than the approximate algorithm.

*Table 3 Relative error between the picker mean waiting time calculated by the exact scheduling algorithm (BAB) and the approximate algorithm (SRPT)*

| FR | Average scheduling time with SRPT | Average scheduling time with BAB | Discrepancy in waiting times between BAB and SRPT (%) |
|---|---|---|---|
| 0.3 | 0.02 min | 12.8 min | 2.0 |
| 0.5 | 0.03 min | 10.6 min | 3.3 |
| 0.7 | 0.01 min | 9.5 min | 6.8 |
| 0.8 | 0.02 min | 8.2 min | 3.1 |
| 0.9 | 0.01 min | 5.9 min | 2.3 |

Since the goals of this work are to investigate changes in performance as the number of robots increases and the FR varies, small errors (in the order of several seconds) in absolute waiting times were not deemed as important. On the other hand, each Monte-Carlo simulation repeats the harvest operation 100 times, and therefore, computational speed was more important than the optimality. Also, the sub-optimal and efficient algorithm is important for the system implementation given the limited computation resources and need for real-time operation. Therefore, the approximate SRPT algorithm was used for all simulation results.

### 7.3. Reactive and predictive schedule performance vs. number of robots

It is expected that as the robot-to-picker-ratio increases and request predictions become available earlier (FR decreases), non-productive time ($\Delta T_{fe}$) and waiting time ($\Delta T^w$) will decrease, causing efficiency to increase. To quantify this behavior, simulations were executed with4, 5, 6, 8, 10, 12, 14 robots and FRs at 0.8, 0.9, 0.95, 1.0. Figure 14 shows the mean non-productive time and its 95% confidence interval (CI) as a function of the robot-to-picker-ratio, with FR as a parameter, and robot speed held constant at 1.5 m/s. The dotted line shows the baseline, i.e., the mean non-productive time for all-manual harvesting, which was 84.5 s, based on data collected in the field (Section 7.1). The curve at FR=1 corresponds to reactive scheduling based on the Shortest Remaining Processing Time criterion (Sections 7.1, 7.2).

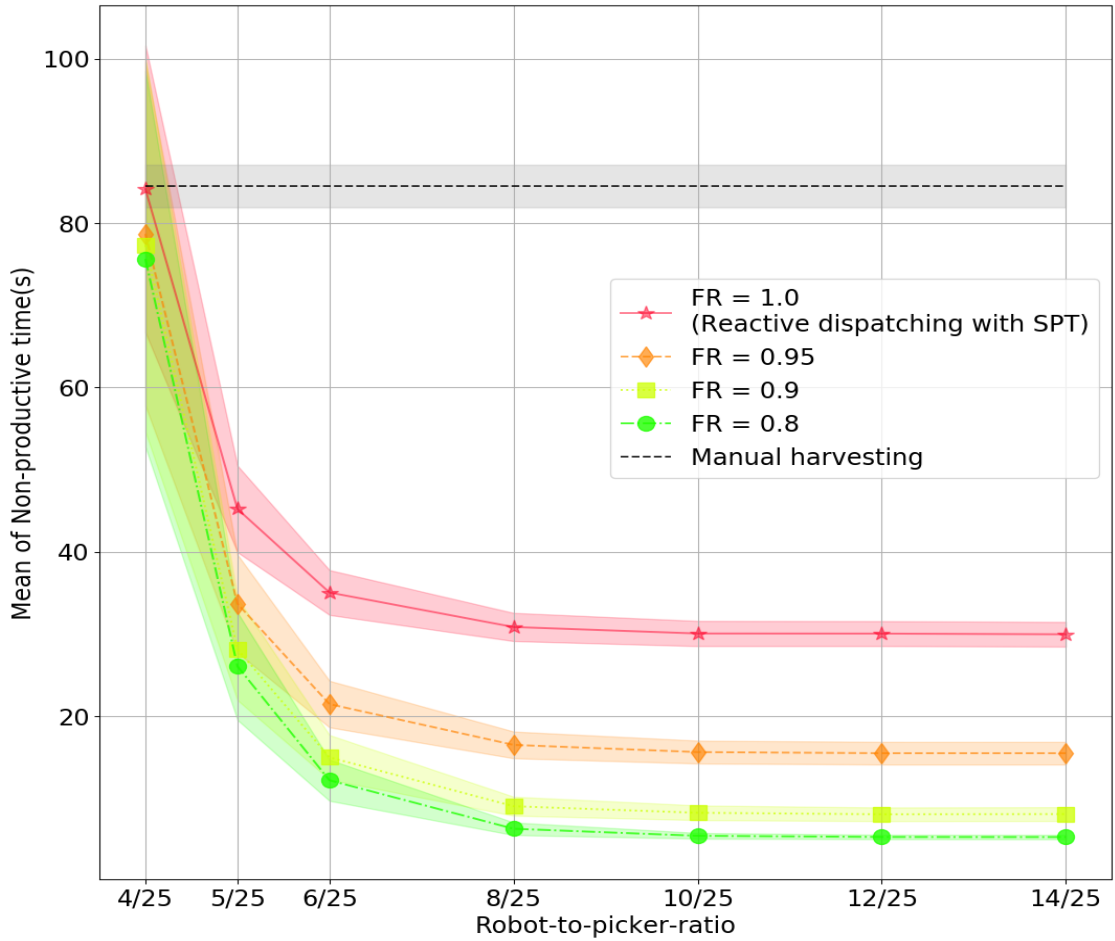

*Figure 14. Mean (points) and its 95% CI (shaded area) of non-productive time as a function of the number of robots, with different Fill-Ratios (FRs); robot speed is 1.5 m/s. The manual harvesting non-productive time (84.5 s) was measured in the field, with a 25-people commercial harvesting crew.*

When only four robots were used, reactive and predictive scheduling did not improve non-productive time dramatically compared to manual harvesting. Deploying fewer than four robots led to worse non-productive time than manual harvesting, and hence these data points were not presented. Introducing five to eight robots decreased non-productive time drastically, and when ten or more robots were used, waiting time was reduced by 64.6% (reactive scheduling) and up to 93.7% (predictive scheduling, with early prediction at FR=0.8).

When reactive scheduling was used, non-productive time plateaued (at 29.9 s) when more robots (in this case ten) were deployed, because picker waiting time cannot be less than the robot travel time from the collection station to the picker. The same was true for predictive scheduling; however, the availability of request predictions reduced non-productive time down to 5.3 s, when predictions were available early enough (FR=0.8). Even very late access to request predictions (FR=0.95) improved the non-productive time (15.5 s). It was noticed that when FR was less than 0.8, robot-aided harvest performance did not improve (hence those curves are not depicted); the reason will be discussed in Section 7.4. The curve at FR=0.8 (and even FR=0) plateaued at 5.3 s. The reason is that the time $\Delta t^{pick}$ required to fill a tray is a random variable that can take small values (e.g., 120 seconds) and in many instances the distance of the picker from the collection station can be large; thus, even when prediction is available immediately when tray-filling starts, the robot will need more time to travel to the picker than $\Delta t^{pick}$.

Figure 14 shows harvesting efficiency (Eff) curves, for all-manual harvesting (black dotted horizontal line), and for robot-aided harvesting with varying number of robots and prediction timeliness (FR) parameters. The efficiency curves exhibit a "reciprocal" behavior to the non-productive time curves ($\Delta t_i^{fe}$ is in the denominator of the efficiency terms, in Eq. 19) and the above discussion about non-productive time vs. robot-to-picker ratio applies to them too.

Next, an analysis is presented to explain the plateaus of the picker waiting (non-productive) time in reactive scheduling, in Figure 14. Each picker needs to wait for the serving robot to traverse from the collection station to them. Given enough robots, if reactive scheduling is used, $\Delta T^w$ can be approximated to be the mean of this distance divided by robot speed, plus the constant tray exchange time. During the simulations, the distances $D_i$ were recorded. These distances depend on the dimension of the field and on yield distribution. Their histogram is shown in Figure 15.

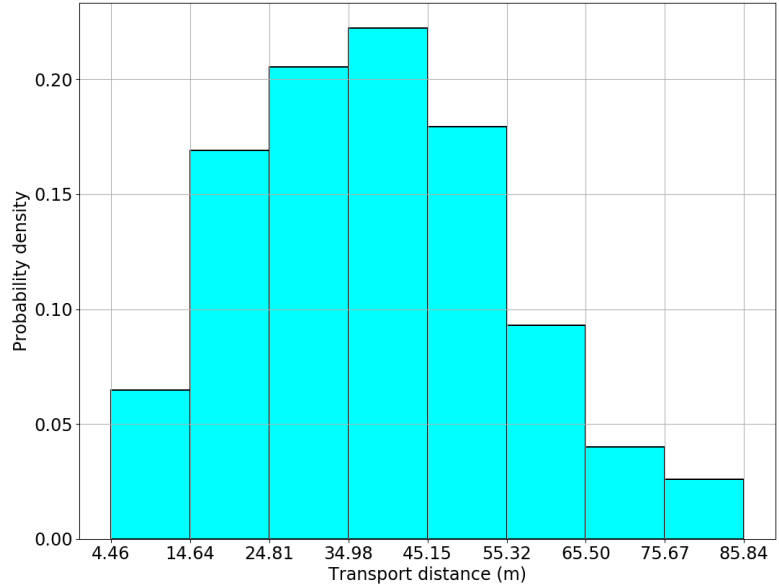

*Figure 15 Frequency histogram of the distance Di from the collection station to the location of the full-tray request; the mean value is 39.74 m*

The mean running distance was $\overline{D} \approx 39.74$ m. Since tray loading time ($\Delta t^L$) was assumed constant –equal to 5 s, in this study - the value of $\Delta T^{fe}$ at the "plateau" - for reactive scheduling - can be estimated from Eq (30).

$$\Delta T_{fe} \approx \frac{\overline{D}}{v^r} + 5 \approx 31.13s \qquad \text{(Eq 22)}$$

The result matches approximately the mean of non-productive time of the curve labeled as "FR = 1.0", in Figure 14. When predictive scheduling is used early enough (FR<0.8) and many robots are available, the robots start traversing $D_i$ early enough to reach pickers before or exactly when the tray becomes full. Hence, the value of the plateau is dominated by the loading time ($\Delta t^L$=5 s).

## 7.4. Predictive schedule performance vs. prediction timeliness of request predictions

Figure 16 shows the relationship between the picker mean waiting time ($\langle \overline{\Delta t^w} \rangle$) and prediction timeliness (FR), when robot speed is 1.5 m/s. When FR was smaller than approximately 0.8, $\langle \overline{\Delta t^w} \rangle$ was almost independent of FR, but it started increasing when FR became larger than this value. The rate of increase became larger as FR got closer to one.

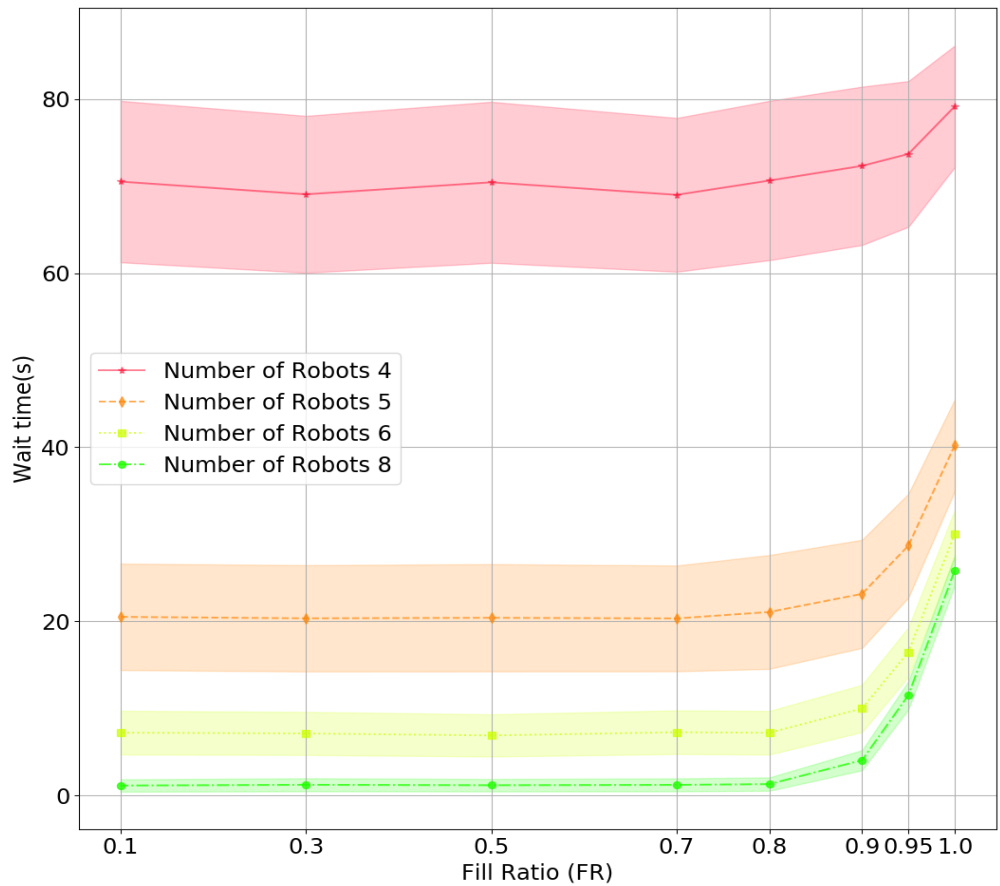

*Figure 16. Mean (points) and its 95% CI (shaded area) of waiting time, $\overline{\Delta t^w}$, as a function of FR, with different the number of robots; robot speed is 1.5 m/s*

To explain this behavior, the concept of the "timeliness" of a predicted request was introduced. The one-way robot travel time from the collection station to the request location $\boldsymbol{L}_t^f$, is $\Delta t_i^u$. A request is characterized as "early" if the robot travel time to the predicted location of the request is shorter than the remaining time, $\Delta t_i^f$, needed to fill the tray (i.e., when $\Delta t_i^u \leq \Delta t_i^f$). When a request is "early" and a robot is available to be dispatched immediately, the corresponding picker will not need to wait for the arrival of a robot. If $\Delta t_i^u > \Delta t_i^f$, then $R_i$ is a "late" request, because even if a robot is available to serve $R_i$ immediately, that picker will still need to wait for the robot, for a time interval equal to $(\Delta t_i^u - \Delta t_i^f)$. The value of $\Delta t_i^u$ depends on robot speed, and the value of $\Delta t_i^f$ is determined by FR (if FR=1, $\Delta t_i^f$ =0; if FR=0, $\Delta t_i^f = \Delta t^{pick}$, else $0 < \Delta t_i^f < \Delta t^{pick}$).

Based on the scheduling model presented in section 3, a robot will not be dispatched to a request until the request's release constraint is satisfied. Therefore, early requests, with an FR value below a certain FR threshold, cannot change the current schedule, and thus will not affect the performance of the system (see the Appendix for the formal proof, and derivation of the threshold equation (Eq. A.8)). This threshold was calculated for our example, as follows. Since the robot velocity is 1.5m/s, $(\Delta t_j^u)_{max}$ can be calculated by the largest possible one-way transport distance, which corresponds to the distance when the collection station is at the center of the headland, and the tray becomes full at the most distance corner of the field from the collection station. Given the dimensions and geometry of the field, and the spacing and dimensions of the beds and furrows, this distance is equal to 71 m. Given that the mean picking time is 275.5s (from the histogram of Fig 13), the FR threshold, according to (Eq A.8) is 0.83. $\Delta T^w$ would be expected to start increasing significantly when FR is above this threshold, a behavior that is seen in Figure 16.

### 7.5. Predictive schedule performance vs. robot speeds

As mentioned above, the robot's one-way travel time $\Delta t_i^u$ for request $R_i$ depends on robot speed $v_r$; higher speeds result in smaller one-way travel times and $(\Delta t_i^u)_{max}$ also increases. Hence, the FR threshold increases. We defined $FR^t$ as the significant changing point for the mean of wait time compared to FR=0.5 on each curve. In Figure 17, one can see that the turning points of $FR^t$ of the curves does shift right (increase) as the robot speed increases. To find the $FR^t$, Tukey's HSD (honestly significant difference) tests were made for a list of candidate FRs that cover $FR^t$ based on the observation of Figure 17. The inspected candidate FRs for different robot speeds are listed in Table 4.

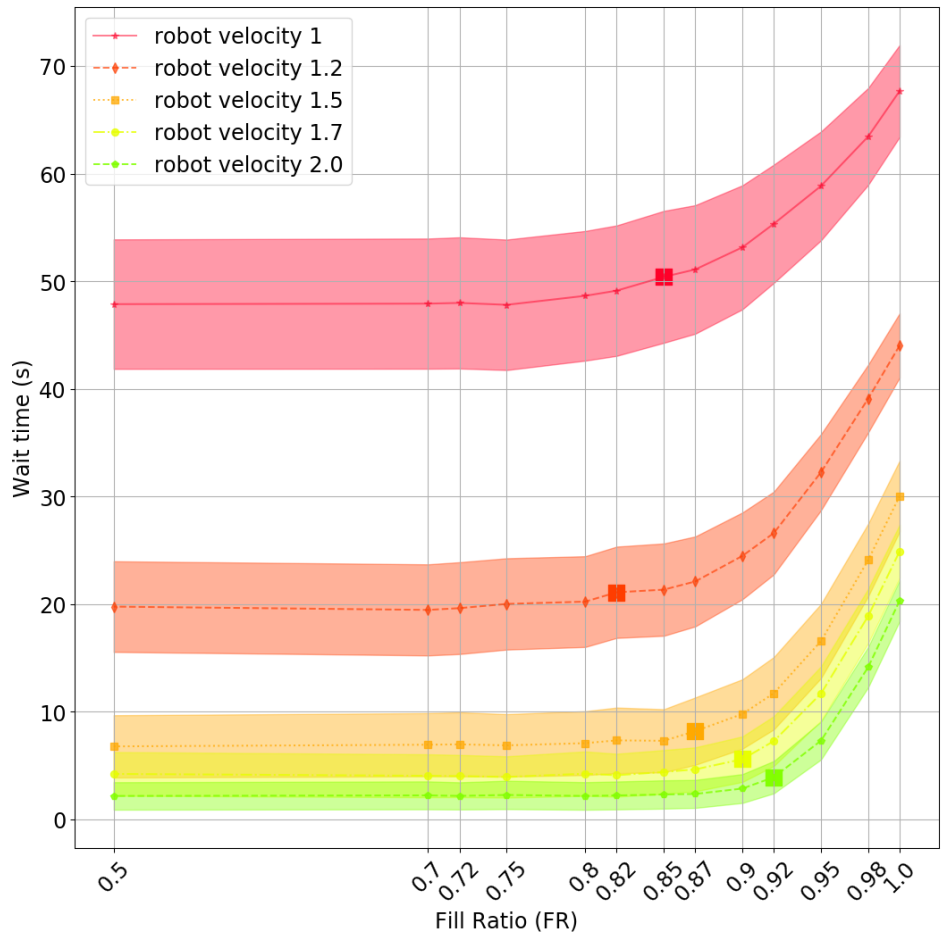

*Figure 17. Mean (points) and its 95% CI(shaded area) of picker waiting time ($\overline{\Delta t}^w$) as a function of FR, at various robot speeds; robot-to-picker ratio is 6/25*

*Table 4. FR candidates to make Tukey's HSD test based on the observation from Figure 17*

| Robot speed (m/s) | FR candidates |
|---|---|
| 1 | {0.50,0.75,0.80,0.82,0.85} |
| 1.2 | {0.50,0.80,0.82,0.85,0.87} |
| 1.5 | {0.50,0.82,0.85,0.87,0.90} |
| 1.7 | {0.50,0.87,0.90,0.92,0.95} |
| 2.0 | {0.5,0.90, 0.92,0.95} |

The null hypothesis of Tukey's HSD test was that all mean wait times for all the FR in the candidate list were the same for each evaluated speed. The alpha value for the tests was set to 1% (Type I error). Taking as an example for the robot speed of 1m/s, the Tukey's HSD table of all comparing combinations of candidate FRs is presented in Table 5. From the column of "Null rejections", one can see that the mean waiting time was significantly different from FR=0.5 when FR was at 0.85. Thus, the start turning point $FR^t$ was chosen at FR=0.85 for the curve of the robot speed of 1 m/s. Similarly, the $FR^t$ and the respective p-values compared to the mean wait times at FR=0.5 for the other robot speeds are listed in column 5 of Table 6 (square points in Figure 17).

*Table 5. Tukey's HSD testing results for all comparing combinations of the candidate FRs for the robot speed at 1m/s*

| FR comparing combinations | | Adjusted p-values | Null rejections |
|---|---|---|---|
| 0.50 | 0.75 | 0.901 | False |
| 0.50 | 0.80 | 0.732 | False |
| 0.50 | 0.82 | 0.301 | False |
| 0.50 | 0.85 | 0.001 | True |
| 0.75 | 0.80 | 0.851 | False |
| 0.75 | 0.82 | 0.612 | False |
| 0.75 | 0.85 | 0.001 | True |
| 0.80 | 0.82 | 0.901 | False |
| 0.80 | 0.85 | 0.001 | True |
| 0.82 | 0.85 | 0.004 | True |

*Table 6. $FR^t$ from the simulation results, $\widehat{FR^t}$ estimated from Eq 29 and their discrepancy percentage*

| Robot velocity (m/s) | $FR^t$ | FR thresholds | Discrepancy (%) | P values |
|---|---|---|---|---|
| 1.0 | 0.85 | 0.74 | 12.9% | 0.0012 |
| 1.2 | 0.82 | 0.79 | 3.67% | 0.0053 |
| 1.5 | 0.87 | 0.83 | 4.59% | 0.0005 |
| 1.7 | 0.90 | 0.85 | 5.56% | 0.0014 |
| 2.0 | 0.92 | 0.87 | 5.43% | 0.0008 |

The estimated FR thresholds from Eq A.8 were shown in the third column in Table 6. FR threshold gave a conservative instructive timeliness for the predictive module to generate the predictive requests. As a result, it can be concluded that FR affects predictive scheduling performance significantly when it is over the estimated threshold. Thus, it is important to utilize the FR below the estimated threshold.

## 8. Summary and conclusions

In this chapter, strawberry harvesting under the assistance of tray-transporting robots was investigated. Dynamic predictive scheduling was modeled and implemented, assuming accurate information of the locations and times of the next tray-transport requests (deterministic predictions). Also, the influence of the earliness of the availability of transport requests on scheduling performance was studied. The study was based on a harvesting simulator that modeled human pickers and transport robots and utilized manual harvesting model parameters estimated from data collected one day during harvesting of a commercial strawberry block in 2018.

Experimental results showed that to achieve a dramatic reduction in picker mean non-productive time and increase harvest efficiency, the robot-to-picker ratio had to be above 1:5 (1 robot for every 5 pickers); deploying fewer than four robots led to worse non-productive time than manual harvesting without robots. When the robot-to-picker ratio was larger than approximately 1:3 (8 robots for 25 pickers) the waiting time and efficiency plateaued, regardless of how early the prediction was available to the scheduler (i.e., how small FR was). The reason is that, if a robot is always available to serve a predicted request, the picker mean waiting time is the sum of mean travel time plus the tray exchange time, which are both constant; when predictions are made very early, waiting time is lower bounded by the tray exchange time. When ten or more robots were used, non-productive time was reduced by 64.6% (reactive scheduling) and up to 93.7% (predictive scheduling) with respect to all-manual non-productive time. The corresponding efficiency increases were 15% and 24%. Reactive dispatching (FR = 1) performed always worse than deterministic predictive scheduling (FR < 1), because robot travel time to the pickers contributed significantly to picker waiting times.

When the robot/picker ratio is up to 1/4 under the predictive scheduling policy SRPT-Convert, the non-productive time is reduced around 40% compared to the reactive scheduling method and over 70% compared to manual harvesting. The influence of request prediction earliness, FR, on the predictive scheduling was also analyzed. FR starts affecting the performance of the predictive scheduling after a FR threshold is reached. The FR threshold can be estimated in advance, given a specific harvesting configuration.

# Chapter 3 Stochastic predictive scheduling of robot team

## 1. Introduction

In this Chapter, the developed simulation system integrated some practical scenarios. During harvesting, the predicted transport requests must rely on real-time sensor data and will always contain uncertainty, as a result of unknown ripe fruit distribution, picker work patterns and sensor noise (Khosro Anjom & Vougioukas, 2019). Also, the robot speed cannot be very fast on the shared headland area where the harvesting facilities and human workers are located. Given the slower robot speed, it may happen that the pickers may need to wait for a too long time if they are constrained to wait for the robot to transport their trays in the low robot/picker ratio. Thus, request rejections are included in the scheduling decisions, which means that the scheduler is to signal some pickers to transport the full tray themselves to minimize average non-productive time of all pickers. In field logistical operations, primary units (equipment, human pickers, etc.) and support vehicles form a closed system. The delays introduced by the support vehicles affect the primary units' temporal distributions of future service requests. During strawberry harvesting, the uncertainty in transport requests is gradually revealed as the harvesting activity progresses, so the scheduling decisions may need to be adjusted in real-time given the dynamically updated input information.

The agricultural vehicle scheduling (or in-field logistics) problem falls under the broad category of online stochastic combinatory optimization (OSCO) (van Hentenryck et al., 2010). Solving the online problem with stochastic programming could guarantee convergence to the optimal solution (Pillac et al., 2013), but may take too long to compute for a single machine. In our case, a fast and near-optimal solution is pursued to exploit the stochastic predictive requests, as the decision time and computation resources for the scheduler are quite limited in the field operation. Bent & Van Hentenryck (2004) proposed a scenario sampling-based algorithm, Multiple Scenario Approach (MSA), on the partially dynamic vehicle routing problem with time windows. In the planning stage, MSA sampled multiple scenarios when the distributions of stochastic request predictions are accessible. Given each sampled scenario, the scheduling was deterministic and solved quickly into a scenario plan. At the decision instant, an optimized plan is formulated by a consensus function (R. Bent & Van Hentenryck, 2004 June) that generates an executing plan most consistent with the optimal solution of the sampled scenarios. This methodology can be adapted and applied to different OSCO problems by building two case-dependent modules: a scenario sampling function to get multiple deterministic scheduling scenarios and a scheduling solver for each sampled deterministic scenario.

## 2. Scheduling of crop-transport robots under stochastic requests

In the previous chapter, the predictive scheduling of crop-transport robot was modelled as a parallel machine scheduling problem (PMSP) with a release time constraint, where the objective was to minimize total waiting time. In this chapter, three practical adaptions were added to the model: (1) predictive tray transport requests are expressed as predicted stochastic distributions; (2) the robot speed is set slower for the consideration of safety; (3) request rejections are included in the decision process of the scheduling algorithm. In the previous chapter, it was assumed that a picker must wait for the robot to come after they fill the tray, even if the waiting time of the picker is longer than the time it would take them to walk and deliver the tray themselves. However, this will not be acceptable by the pickers in real situations, as the pickers are paid based on the number of trays collected and long waiting time leads to less salary. As a result, the rejection of requests was introduced in this chapter (when robot service would result in slower-than-manual transport) as a feature of our scheduling algorithm. The objective is to minimize the expected total non-productive time of all the transport requests.

In robot-aided harvesting, each picker from a set $\mathcal{S}^{\mathcal{P}} = \{P_1,\ P_2,\ \dots,\ P_Q\}$ of $Q$ pickers harvests fruits in a tray that lies on a picking cart. A team of $M$ identical transport robots $\mathcal{S}^{\mathcal{F}} = \{F_1,\ F_2,\ \dots,\ F_M\}$ bring empty trays to the picker and carries the full tray to a collection station; the station's coordinates $\mathrm{L}^s$ are known. The robot scheduling algorithm has access to a set of predicted tray-transport requests $\mathcal{S}^{\mathcal{R}} = \{\mathcal{R}_1,\ \mathcal{R}_2,\ \dots,\ \mathcal{R}_N\}$ where $0 \leq N \leq \mathrm{Q}$.

Let us assume that at an instant $t_0$, $\mathcal{R}_i$ is different from the deterministic request that contains the following (known) information: (1) a prediction distribution of the remaining time interval $\aleph(\Delta t_i^f)$ with respect to $t_0$ until the tray becomes full of harvested fruit, (2) the predicted moving speed along the row $\aleph(v_i^y)$ while picking, and (3) the current location of the picker $\boldsymbol{L_i}$. $\aleph(\Delta t_i^f)$ is calculated from recent measurements from the load cells and $\aleph(v_i^y)$ is computed from recent GPS readings. . The main methodology for building these predictions is explained in this work (Khosro Anjom & Vougioukas, 2019). The distribution of $\aleph(\Delta t_i^f)$ and $\aleph(v_i^y)$ followed Gaussian distributions. $\aleph(\Delta t_i^f)$ was achieved by linear regression model to predict the value of full tray time at the weight of the tray capacity. Mean of $\aleph(v_i^y)$ was obtained by linear regression to estimate the slope parameter and standard deviation of $\aleph(v_i^y)$ was obtained from the standard error of the regression coefficient.

A fast and near-optimal approach, MSA (Pillac et al., 2013), was adopted and adapted to incorporate the dynamic stochastic predictive requests in the computation of the schedule, assuming a limited computational power is available in this agricultural simulation. To implement MSA, two application-dependent functions must be setup: (1) a function GET-SAMPLES ($\mathcal{S}^{\mathcal{R}}, M$) which returns a set of M deterministic scenarios $\mathcal{S}^{\xi} = \{\mathcal{S}^{\xi_1}, \mathcal{S}^{\xi_2}, \ldots, \mathcal{S}^{\xi_M}\}$. Each scenario $\mathcal{S}^{\xi_i}$ contains a set of $N$ sampled requests. Each of the deterministic requests $R_i$ is sampled from predictive transport request distributions $\aleph(\Delta t_i^f), \aleph(v_i^y)$ of $\mathcal{R}_i$ in $\mathcal{S}^{\mathcal{R}}$; (2) a function OPTIMAL-SCHEDULE ($\mathcal{S}^{\xi_i}$) which returns an optimal schedule given a deterministic sampled scenario $\mathcal{S}^{\xi_i}$. The schedule includes the request rejections to some pickers and serving order for the remaining requests (3) a *consensus* function that combines all the individual scenario solutions into a single execution plan. For the function GET-SAMPLES ($\mathcal{S}^{\mathcal{R}}, M$), the Monte Carlo sampling method was used to get the $M$ sampled scenarios from two distributions $\aleph(v_i^y)$ and $\aleph(\Delta t_i^f)$.

In a sampled scenario $\mathcal{S}^{\xi_i}$, each deterministic predictive request $R_i$ is composed of two sampled components, $\Delta t_i^f$ and $v_i^y$. Given them, a deterministic full tray location $\boldsymbol{L}_i^f$. can be calculated. Then, the variables relevant to the modeled scheduling problem can be calculated shown as Table 7 (Peng & Vougioukas, 2020).

*Table 7. Definitions of Symbols used in the modeling of deterministic predictive scheduling*

| | |
|---|---|
| $\boldsymbol{L}^s$: | the collection station location; |
| $\boldsymbol{L}_i^f$: | the full tray location in the field frame. |
| $D_{si}$: | one-way traveling distance, the Manhattan distance from $\boldsymbol{L}^s$ to $\boldsymbol{L}_i^f$ along the path; |
| $\Delta t_i^u$: | the corresponding robot's one-way travel time calculated by $D_{si}$ and robot speed; |
| $\Delta t^L$: | time interval when the picker takes the empty tray from the robot and loads the full tray on the robot (and then resumes picking); |
| $\Delta t^{UL}$: | the time interval for the collection station to unload the carried tray from the picker/robot and return an empty tray to the picker/robot; |
| $\Delta t_i^p$: | The total processing time required by a robot to serve request Ri and be available to serve another request; |
| $\Delta t_i^r$: | release delay of request Ri, the greatest value that eliminates robot idle time at $\boldsymbol{L}_i$. $\Delta t_i^r = max(\Delta t_i^f - \Delta t_i^u), 0)$ |

$\Delta t_k^A$:  The robot is available to be dispatched again, after a time interval $\Delta t_k^A$. $\Delta t_k^A = 0$, if the robot is available at the collection station;

$t_{ki}^d$:  The dispatch time instant of robot $F_k$ to the request $R_i$, which is no earlier than $t_0 + \Delta t_i^T$.

The pickers' requests may be rejected by the scheduler. In this case, they need to transport the full tray themselves and their self-transporting behavior is modeled following our previous work (Seyyedhasani et al., 2020b, 2020a). If the picker transports the tray themselves the total time $\Delta t_i^T$, required to deliver the full tray and take an empty tray back to resume picking is shown as (Eq 23). $\Delta t_i^{u_P}$ is the one-way travel time interval from full tray location $\boldsymbol{L}_i$ to $\boldsymbol{L}^s$ by the picker in $R_i$. $\Delta t_i^{u_P}$ is calculated based on $D_{si}$ and an estimated picker self-transport speed $v_p^i$ from the historic data of pickers. $\Delta t^{UL}$ is assumed to be constant depending on the crew management in the harvesting field.

$$\Delta t_i^T = 2\Delta t_i^{u_P} + \Delta t^{UL} \qquad \text{(Eq 23)}$$

The tray completion time instant, $t_i^{C_P}$ if the full tray is transported by the picker himself, is shown in (Eq 24) with $\Delta t_i^T$ representing the estimated tray-transport time by the picker.

$$t_i^{C_P} = t_i^f + \Delta t_i^T \qquad \text{(Eq 24)}$$

If the request is served by a robot $F_k$, the time instant, $t_i^{C_R}$ to resume picking is expressed as (Eq 25). The picker can start picking the next tray after the robot arrived at the full tray location and the full tray is exchanged with the empty tray from the robot.

$$t_i^{C_R} = t_{ik}^d + \Delta t_i^u + \Delta t^L \qquad \text{(Eq 25)}$$

The nonproductive time, $\Delta t_i^N$ of $R_i$ can be calculated as (Eq 26). The objective of the modeled problem is to minimize the mean of the nonproductive time of all the pickers. In the objective function, both $t_i^{C_P}$ and $t_i^{C_R}$ are represented by $t_i^C$ which is decided by the decision variables.

$$\Delta t_i^N = t_i^C - \Delta t_i^f \qquad \text{(Eq 26)}$$

After building the mathematic equations of these modeled variables, the function OPTIMAL-SCHEDULE ($\mathcal{S}^{\xi_i}$) was built to find the solution for each scenario. First, the exact solution is computed using integer programming to get the best possible solution. Second, a fast and sub-optimal heuristic policy is implemented to get a near-optimal solution in less time, so that the pickers do not need wait for long time caused by the scheduling computation. The results of the exact and heuristic solutions are compared in the proposed performance metrics in Section 7 of this chapter.

## 2.1. Scenario solution with integer programming

The deterministic predictive scheduling problem of each sampled scenario was modelled using an integer linear program. $\mathcal{S}^{\mathcal{T}}$ is used to represent the discretized time set, $\{1, 2, 3..., TB\}$. $TB$ is the upper bound makespan of all requests (from $t_0$ to $t_0 + max_i\{t_i^C\}$). For this problem, the upper bound $TB$ can be expressed as (Eq 27). It is easy to prove that the completion time of any request cannot be larger than the maximum completion time of self-transporting, otherwise that request should be transported by the picker themselves

$$TB \leq t_0 + max_i\{\Delta t_i^f\} + \max_i\{\Delta t_i^T\} \qquad \text{(Eq 27)}$$

The decision variable is defined as $\chi_{ikt}$, where $i$ is the index of request $R_i$. $k$ is the index of the serving robot if $1 \leq k \leq M$; $k = M + 1$ means that the picker transports the tray himself. $t$ is the index of discrete-time instant. $\chi_{ikt}$ is equal to 1 if $R_i$ is served by a robot $F_k$ ($1 \leq k \leq M$) or transported by the picker himself ($k = M + 1$) at the time instant $t$. The problem can be modeled using an integer linear programming (ILP) as follows.

$$min \sum_{i=1}^{N} \Delta t_i^N$$

s.t.

$$\sum_{k=1}^{M} \sum_{t=1}^{\Delta t_i^T} \chi_{ikt} = 0, \ R_i \in \mathcal{S}^{\mathcal{R}} \qquad \text{(Eq 28)}$$

$$\sum_{i=1}^{N} \sum_{t=1}^{\Delta t_k^A} \chi_{ikt} = 0, \ 1 \le k \le M \qquad \text{(Eq 29)}$$

$$\sum_{k=1}^{M} \sum_{t=1}^{t_i^f} \chi_{i(M+1)t} = 0, \ R_i \in \mathcal{S}^{\mathcal{R}} \qquad \text{(Eq 30)}$$

$$\sum_{k=1}^{M+1} \sum_{t=1}^{TB} \chi_{ikt} = 1, \ R_i \in \mathcal{S}^{\mathcal{R}} \qquad \text{(Eq 31)}$$

$$\sum_{k=1}^{M} \sum_{t=max(1, t-\Delta t_i^p)}^{t} \chi_{ikt} \le 0, \ R_i \in \mathcal{S}^{\mathcal{R}} \qquad \text{(Eq 32)}$$

$$t_i^C = \sum_{k=1}^{M} \sum_{t=1}^{TB} (t + t_i^U + \Delta t^L) \chi_{ikt} , \ R_i \in \mathcal{S}^{\mathcal{R}} \qquad \text{(Eq 33)}$$

$$t_i^C = \sum_{t=1}^{TB} (t + t_i^T) \chi_{i(M+1)t} , \ R_i \in \mathcal{S}^{\mathcal{R}} \qquad \text{(Eq 34)}$$

$$\Delta t_i^N = t_i^C - \Delta t_i^f, \ R_i \in \mathcal{S}^{\mathcal{R}} \qquad \text{(Eq 35)}$$

The objective function is the sum of the non-productive time of all requests and the required constraints are explained as follows. In (Eq 28), it represents that any request cannot be served by a robot before their release constraints. (Eq 29) means that the robot's start serving time cannot be earlier than their initial available time. If the request is transported by the picker himself, the start time cannot be earlier than the full tray instant $t_i^f$ as (Eq 30). (Eq 31) represents that all requests must be served either by a robot or by the picker himself. (Eq 32) shows that any requests can be served by only one robot (preemption is not allowed). (Eq 33) expresses the tray completion time of the request served by the robots, while (Eq 34) is the tray completion time served by the pickers themselves. (Eq 35) shows the non-productive time of request $R_i$.

As explained in the previous chapter, predictive scheduling of crop-transport robots is a variant of the Parallel Machine Scheduling Problem (PMSP). Following symbol notations defined by Lawler et al. (1993), the problem is referred to as $Pm|r_i|\sum C_i$, where $Pm$ represents identical parallel machines, $r_i$ means that the ith job cannot be processed until its release time, and $\sum C_i$ represents that the objective criterion is to minimize the sum of the completion times of all jobs. It has been shown that this problem is NP-hard in a strong sense and hence the optimal solution cannot be obtained in polynomial time (Du et al., 1991). In this paper, the modeled ILP was solved by a commercial solver (Gurobi Optimization, LLC., 2020) at the cost of long computation.

## 2.2. Scenario solution with heuristic policy

A heuristic policy, namely, the shortest release time with long process time first (SRLPT), is proposed to achieve a fast but sub-optimal result in each sampled deterministic scenario. The requests reaching the release constraint ($\Delta t_i^r = 0$) will enter a scheduling pool and the request with the longest process time in the pool is ordered to be served by the first available robot. The non-productive time of those requests with large self-transport time can be reduced significantly by the service of available robots. The requests with a shorter transport time are served late, as even if they are rejected, the non-productive time will not be that large. The requests with a full tray location less than 5 meters away from the end of the row are rejected, as those pickers only need to walk a small distance back and forth to resume picking. The performance comparison between the heuristic policy and ILP is shown in Section 6.1.

## 2.3. Consensus function

Given solutions in multiple sampled scenarios, the consensus function is to select the distinguished plan from the current pool of scheduling plans. Bent and Pascal (2004) first applied the consensus function into a modeled partially dynamic vehicle routing problem with a time window. They pointed out that this approach is essentially domain independent. Hence, they applied a similar consensus approach on the classic scheduling problem, the online packet scheduling problem in computer networks. The key idea is to solve each sampled scenario once and to select the packet which is most often in the optimal solution of each scenario. The heuristic idea behind the consensus function is the least-commitment approach, a well-known approach in the artificial intelligence community (R. Bent & Van Hentenryck, 2004 June). By choosing the job that occurs the most often, the consensus algorithm takes a decision that is consistent with the optimal solution of many samples.

This consensus approach was applied to our modeled scheduling problem. After all the deterministic scenarios are solved with the function of OPTIMAL-SCHEDULE, the scheduling plan of each scenario is converted to a serving order based on their scheduled serving times. A score function is defined for each request in one scenario. If the request is rejected, the score value of that request is counted as -1. If the request is served by a robot in the order of $O_i$ among all the serving requests in that scenario, the score of that request is counted as $(N - O_i)$. The score of each request is obtained by adding the scores of the requests among all the sampled scenarios. The consensus serving order is the descending order of the scores of all the requests in $\mathcal{S}^{\mathcal{R}}$. The available robots were dispatched to the first request in the consensus order at the instant when the expected release time of that request is reached. The rejection flags are sent to the pickers if they are not served by the robot at the instant when their trays are full. The scheduler will run to update the scheduling plan only when there are robots available and new transport requests entering the set.

## 3. Modeling harvesting activity under uncertain request prediction

In Chapter 2, a discrete-time hybrid systems model was developed to model and simulate the activities and motions of all agents involved in robot-aided harvesting. A Finite State Machine (FSM) was utilized to model the discrete operating states/modes of the agents and the transitions between the operating modes. In this chapter, the FSM is significantly extended to consider more possible cases for serving stochastic transport requests and integrate the tray-transport request rejection policy in pickers' harvesting activities. The activities of a picker during robot-aided harvesting were classified into 14 discrete operating states/modes (Table 8), and the operations of a tray-transport robot into 9 states (Table 9). The operating states of pickers and robots and the possible transitions amongst them are shown in Figure 18. In FSM of pickers, they need to transport the trays themselves if they receive the request rejections. In FSM of robots, it may happen that the robot is dispatched to a row where the served picker cannot fill their tray and take the half-filled tray to the next unharvested row. In this case, the robot drives back to the collection station to wait for the next dispatching command.

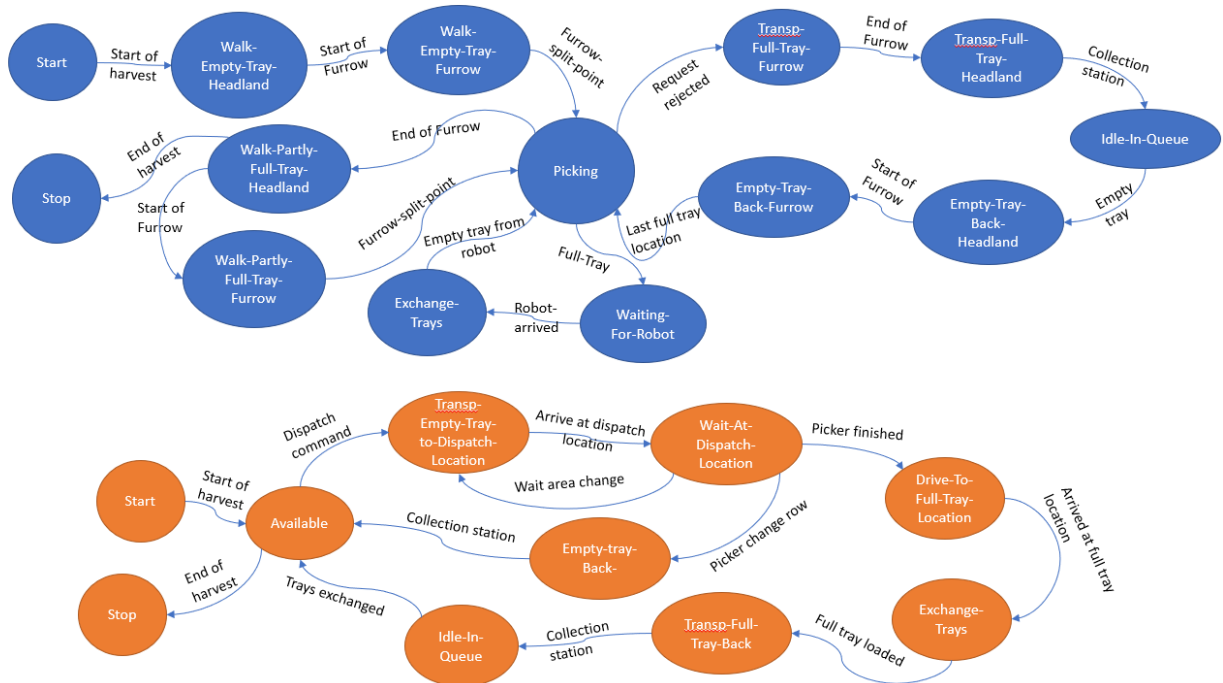

*Figure 18. State diagram of picker states and transport robot states during human-robot collaborative harvesting*

*Table 8. States defined to represent a picker's operating states during robot-aided harvesting*

| Operating state | Action |
|---|---|
| Start | A picker leaves the collection station with an empty tray in hand, to start picking. |
| Walk-Empty-Tray-Headland | A picker walks with an empty tray on the headland, toward an empty (unoccupied) furrow. |
| WALK-Empty-Tray-Furrow | A picker walks inside an empty (unoccupied) furrow with an empty tray until the field's split line is reached. |
| Picking | A picker is picking inside a furrow, with direction from the field split line toward the collection station. |
| Waiting-For-Robot | A picker waits (idle), with a full tray, for a robot to come. |
| Exchange-Trays | A picker takes the empty tray brought by the robot and places a full tray on the robot. |
| Walk-Partly-Full-Tray-Headland | A picker takes partly full tray on the headland, toward an empty (unoccupied) furrow. |
| Walk-Partly-Full-Tray-Furrow | A picker takes a partly full tray inside an empty (unoccupied) furrow until the field's split line is reached. |
| Transport-Full-Tray-Furrow | A picker takes a full tray inside a furrow towards the headland |
| Transport -Full-Tray-Headland | A picker takes a full tray on the headland towards the collection station |
| Idle-In-Queue | A picker waits in a line at the collection station to deliver her/his full tray and receive an empty tray. |
| Empty-Tray-Back-Headland | A picker walks in the headland - toward the last full tray furrow - carrying an empty tray, to continue harvesting. |

| Empty-Tray-Back-Furrow | A picker walks back to the last full tray location with an empty tray, to continue harvesting. |
|---|---|
| STOP | A picker stops picking after the last tray is picked up by a robot. |

| Operating state | Action |
|---|---|
| Start | A robot at the collection station starts operation with no tray on it. |
| Available | A robot with one empty tray on it is waiting at the collection station to be dispatched to a tray-transport request. |
| Transp-Empty-Tray-to-Dispatch-Location | A Robot travels from a collection station – carrying an empty tray – toward the dispatched location. |
| Wait-At-Dispatch-Location | A robot arrives at the location of the tray-transport request and waits for the picker to finish harvesting. |
| Drive-To-Full-Tray-Location | A robot drives to picker's full tray location after served picker fills the full tray in its dispatched row. |
| Empty-Tray-Back | A robot runs back to collection stations as the served picker cannot fill the tray in its dispatched row |
| Exchange-Trays | A robot is idle while the picker exchanges the empty tray with a full tray. |
| Transp-Full-Tray-Back | A robot travels toward the collection station to deliver a full tray. |
| Idle-In-Queue | A robot with a full tray waits in a queue at the collection station to have its tray unloaded, and an empty tray loaded. |
| Stop | A robot stops its operation at the collection station after the last tray has been unloaded. |

Request rejections are integrated into the operation of pickers. When the rejection flags are received, the pickers will transport the full tray by themselves as in manual harvesting. The relevant pickers' states are "Transport-Full-Tray-Furrow", "Transport -Full-Tray-Headland", "Idle-In-Queue", "Empty-Tray-Back-Headland" and "Empty-Tray-Back-Furrow". If the picker is served by a scheduled robot, they will wait at their full tray locations to exchange trays from the coming robot. In this case, the states after picking are "Waiting-For-Robot" and "Exchange-Tray". Time interval spent between the full tray instant, and the starting instant of next tray picking is denoted as "non-productive" time. The other pickers states, like our previous work (Seyyedhasani, Peng, Jang & Vougioukas, 2020a), are represented in the same form and updated with the same state-dependent difference equations. The stochastic parameters are estimated experimentally with the same distribution (Peng & Vougioukas, 2020).

Since the predictions of tray-transport requests contain uncertainty, the exact location when the tray will become full is not known. Therefore, instead of sending the robot to the predicted location, a safety distance to the full tray location was introduced for the robots' goal points. The robot is dispatched 5 meters away from the predicted full tray location to wait for the picker to fill his/her tray. The pickers need to travel this small distance to load the tray onto the robot and take an empty tray back. Also, given the imperfect full tray predictions, it may happen that a picker who is predicted to fill his/her tray inside the current furrow may not fill the tray in that furrow, but the robot has been dispatched at that furrow before the picker travel to the new row. In this case, the robot must return to the active collection station ("Empty-Tray-Back") and become available there. The other states of the crop-transport robots are updated like our previous work (Peng & Vougioukas, 2020).

## 4. Implementation of the harvest simulator

Software was developed to simulate robot-aided strawberry harvesting based on the hybrid systems model presented in Section 3. This simulator constitutes a significant extension and adaptation of the simulator developed by Seyyedhasani et al., (2020a) to incorporate the stochastic predictions of requests and the MSA scheduling of robots. The architecture of the simulator is shown in Figure 19. The simulator is initialized with the geometrical description of the strawberry field (furrow endpoints, split line, collection station locations), the picking crew and robot team parameters, and the initial locations of pickers, robots, and active collection station. The "Picker operations" and "Robot operations" modules implement the coupled hybrid system models of the pickers and robots, respectively. The "Crop, crew & collection station distribution" module updates the status of each furrow (harvested/unharvested/currently harvesting) and the active collection station and calculates the sequence of furrows picked by the crew (after harvesting from a furrow, a picker moves to the furrow of the closest unharvested bed).

During simulated harvesting, the picker states are input to the "Tray-transport request prediction" module that generates stochastic predicted transport requests given the state of pickers. The stochastic predicted transport requests include a Gaussian distribution of full tray time interval expressed as its estimated mean and variance, and Gaussian distribution of picker' moving speed. Those predicted distributions and robot states are used by the "MSA scheduling" module to compute a schedule plan for the robots and pickers. Given the calculated schedule, the MSA scheduling module will output dispatch commands to the available robots and request rejections to the pickers.

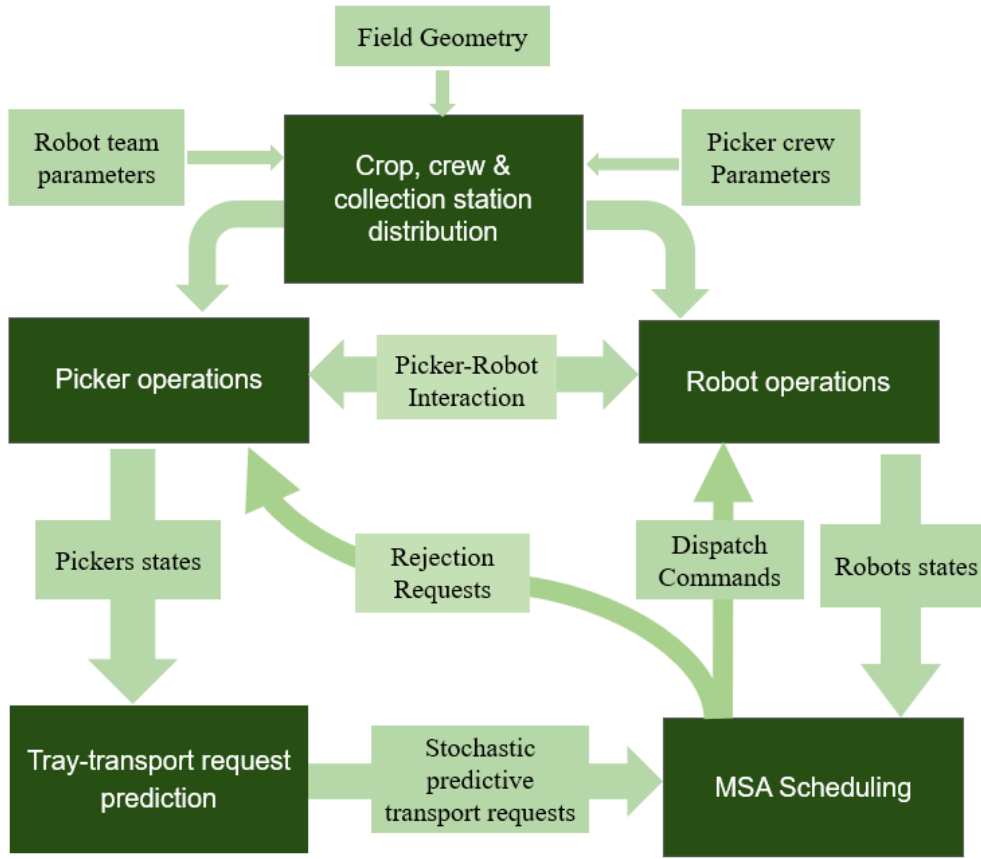

*Figure 19. The architecture of integrated harvesting simulator and predictive scheduling system.*

The simulator uses a global time variable $t$ to represent the current time of the harvesting activity; time starts at $t = 0$ s and increases by $\Delta t$ (0.5s was used). The stochastic harvesting parameters are sampled randomly for each tray - before the tray starts getting filled – from the experimentally derived frequency histograms mentioned in Chapter 2. The states of pickers and robots are updated at each time step and the simulation terminates when the entire field block is harvested.

The commanded robot speed is 0.4 m/s when it is on the headland and 1.2 m/s when inside the furrow. These speeds were set based on field experiments with the mobile robot (Chapter 4). When the picker harvesting parameters and the field dimensions are known, the FR threshold can be estimated using Eq. 36 (Peng & Vougioukas, 2020).

$$FR \leq 1 - \frac{(\Delta t^u)_{max}}{\Delta t^{pick}} \qquad \text{(Eq 36)}$$

In this equation, $(\Delta t^u)_{max}$ is the maximum one-way travel time from the collection station to the location of the request; $\overline{\Delta t^{pick}}$ is the mean time required to fill one tray. Using the parameter values from our experiments, the FR threshold was equal to 0.7.

The uncertainty of the predictive requests is generated based on the results from the work of Khosro Anjom & Vougioukas (2019). They introduced a grey box model to predict the online full tray time of the filling tray. The mean average percentile error (MAPE) $p_e$ was used to evaluate the bias of the predicted mean relative to the ground truth and the standard error $\sigma_e$ to represent the deviation of the prediction relative to the predicted mean. In the real situation, $p_e$ and $\sigma_e$ are dynamically updated in a time series. The performance of MSA scheduling under different possible combinations of $p_e$ and $\sigma_e$ are evaluated. The results of $p_e$ and $\sigma_e$ at FR=0.7 were taken from (Khosro Anjom & Vougioukas, 2019) as an example case when each tray enters the request set $\mathcal{S}^{\mathcal{R}}$. In the simulator, the bias of the mean of each tray was assumed to come from a uniform distribution $\mho\left(-pe \cdot \overline{\Delta t^{pick}}, pe \cdot \overline{\Delta t^{pick}}\right)$. For each transport request, a bias value is sampled from $\mho$. The ground truth of the full tray interval plus the sampled bias is used as the predicted mean of full tray interval. $\sigma_e$ is kept as a constant for each tray. The stochastic full tray interval is expressed as a Gaussian distribution, $N\left(\Delta t^{gt}_{ef} + bias, \sigma_e\right)$.

In the simulation, a uniform noise $\mathcal{U}(-l, +l)$ was added to the y locations of the picker. $l$ was set to 0.5m, to represent the localization accuracy of the Satellite Based Augmented System (SBAS). Linear regression was used to estimate the moving speed $v_y$ of the picker along the row based on successive noisy location measurements.

# 5. Experimental design

Monte-Carlo simulations were performed to investigate the performance of the MSA approach. The field block dimension, crew size, collection station locations, and pickers' parameters were the same as in Chapter 2. The robot speed was 0.4m/s on the headland and 1.2 m/s inside the furrows, and the FR threshold was 0.7. Each harvesting scenario was simulated by running 100 Monte-Carlo runs, as described in Chapter 2. The main underlying assumption in our analyses of the results, is that the 100 sampled means of each evaluated metric were normally distributed.

## 5.1. Evaluation metrics

As introduced in Section 6.1 of chapter 2, the mean harvesting efficiency, $E_{ff}$, and nonproductive time $\Delta T_{fe}$ of $N$ harvested trays were utilized as the metrics for evaluating the performance of the harvesting operations. $E_{ff}$ is estimated with the averaged sum of ratios of productive time over total time spent for each tray; it is calculated by (Eq 20).

The number of harvested trays per hour – denoted as $NumTrays_{hour}$ was used to express the harvesting rate of a crew of pickers. Naturally, if there are more pickers in a crew, a higher number of trays will be harvested per hour. Given a field with a typical dimension, the time interval to harvest the whole field is denoted as $T_{harvest}$ hours and the number of trays harvested from that field denoted as $N_{harvest}$. $NumTrays_{hour}$ can be calculated by Eq 37. The average of $NumTrays_{hour}$ of multiple Monte-Carlo experiments were used as the evaluation metrics for the crew harvesting rate.

$$NumTrays_{hour} = \frac{N_{harvest}}{T_{harvest}} \qquad \text{Eq 37}$$

## 5.2. Experiment introduction

The targets of the experiments in this chapter are as follows: (1) evaluate the effect of request rejections; (2) compare the performance differences between the heuristic and exact algorithms in OPTIMAL-SCHEDULE functions introduced in Section 2; (3) select the number of scenarios the MSA must sample; (4) evaluate the performance of the system using a given prediction uncertainty.

In section 6.1, the effect of request rejections was investigated given the deterministic requests for different robot numbers with the exact algorithms. For the case without request rejections, we applied the BAB search algorithm developed in Chapter 2. For the case with request rejections, we applied the ILP solver introduced in section 2.1. In section 6.2, the scenario scheduling policies in the OPTIMAL-SCHEDULE module of MSA were compared for different robot numbers. In section 6.3, the performance of MSA under different sampling numbers was displayed with the robot number at 8. For MSA, when more scenarios are sampled, better performance MSA can be achieved, but it takes a longer time to calculate a solution. The requests were under the uncertainty measured by Khosro Anjom et al. (2019): the mean of bias for the full tray time prediction is less than 10% of one tray picking time ($\Delta T_{ef}$) and the standard error of the prediction is 30 s. From the results, we chose the critical sampling number that did not improve the scheduling performance dramatically. In section 6.4, the performance of the MSA was evaluated given different numbers of robots. The scenario sampling number was from the result of the last experiment. Also, the crew harvesting rate of 25 pickers under the harvest-aiding system with different numbers of robots was investigated and compared with the manual harvesting with different crew sizes.

# 6. Experimental results and discussion

## 6.1. Comparison of transport scheduling with and without rejections

In this section, the simulation experiment was run to compare the deterministic scheduling performance with request rejections and without rejections. Figure 20 showed that the predictive scheduler with request rejections obviously performs better than without rejections when the robot/picker ratio is smaller than 10/25. The efficiency data of two policies for 10, 11, and 12 robots was evaluated with T-tests to compare their differences, as the 95% confidence intervals of the two mean efficiencies intersected when the number of robots was more than 10. The null hypothesis for the T-tests was that the mean efficiencies of the two policies have no significant differences in the robot ratios of 10/25, 11/25 and 12/25. The alpha value for the tests was set to 1% (Type I error). From the results in Table 10, when the robot number was at 10, the harvesting efficiency with request rejections performs significantly better than the efficiency without requestion rejections given the small p values of the T-test results. However, when the number of robots was 11 or more, there was no significant difference as p value was greater than 0.05.

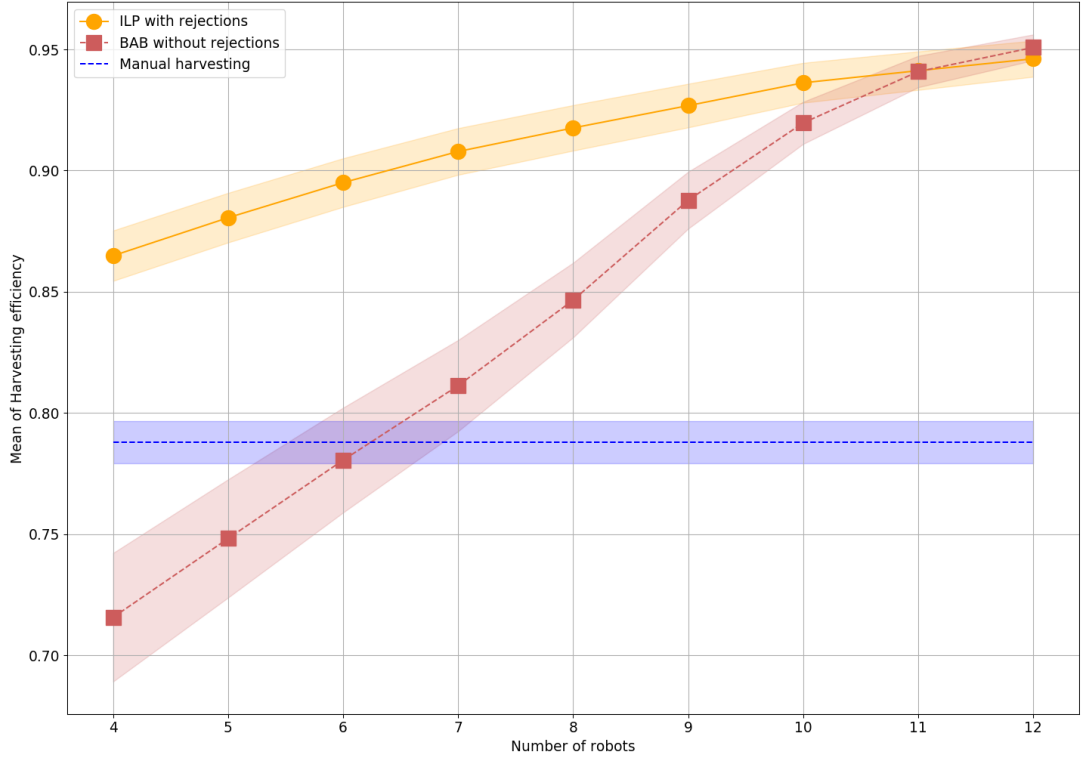

*Figure 20. Mean (points) and its 95% CI (shaded area) of harvesting efficiency as a function of the number of robots for the scheduler with request rejections and without request rejections.*

*Table 10. P values of T-test of scheduling efficiencies with/without request rejections for 10, 11, and 12 robots*

| Number of Robots | 10 | 11 | 12 |
|---|---|---|---|
| P value | 5.24e-11 | 0.823 | 0.141 |

### 6.2. Comparison of serving transport request with ILP and SRLPT

The goal of this experiment was mainly to investigate the performance difference between the ILP and SRLPT methods for the OPTIMAL-SCHEDULE module, given deterministic requests (as in Chapter 2). The comparison of evaluation metrics $E_{ff}$ is shown in Figure 21 for an increasing number of robots.

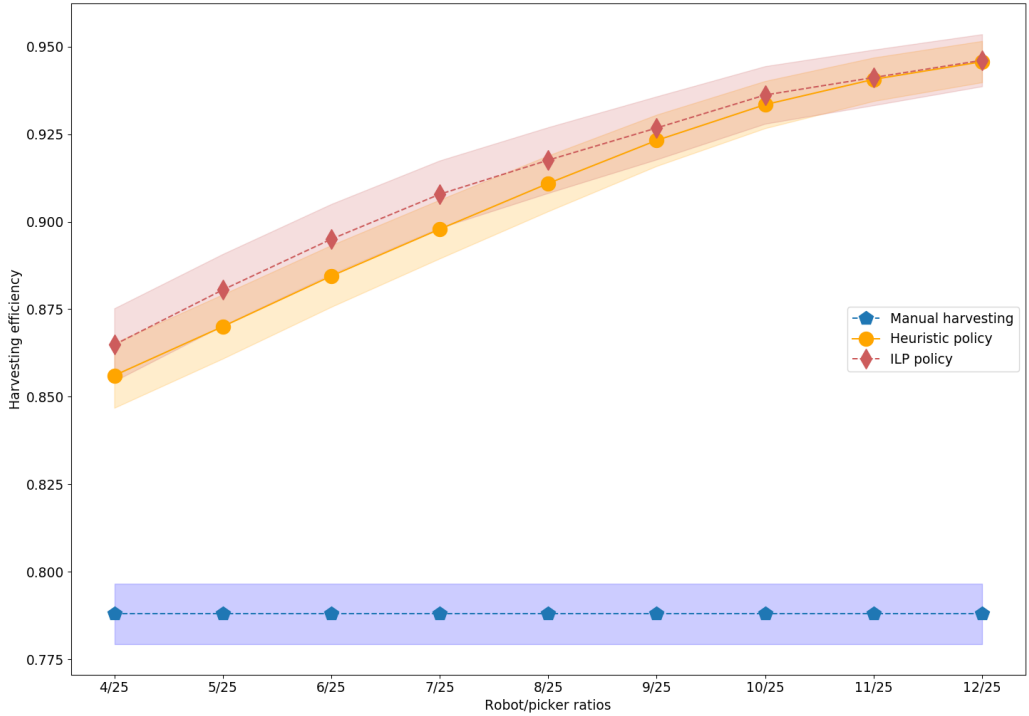

*Figure 21. Mean values of harvesting efficiency (points) and their 95% CI (shaded areas) as a function of the number of robots for the scheduler with ILP and heuristic SRLPT.*

T-tests were applied for the mean efficiencies of the two policies under different robot-picker ratios. The null-hypothesis of each T-test was that the mean efficiencies of the two policies were the same. The alpha value for the tests was set to 5% (Type I error). The results are presented in Table 11, where one can see that the performance of two policies was not significantly different when the robot/picker ratio was over 8/25, as all p values of the T-test results were over 0.05.

*Table 11. P values of T-tests for the efficiencies of two policies in different robot/picker ratios*

| Robot/picker ratio | 4/25 | 5/25 | 6/25 | 7/25 | 8/25 | 9/25 | 10/25 | 11/25 | 12/25 |
|---|---|---|---|---|---|---|---|---|---|
| P values | 0.030 | 0.015 | 0.013 | 0.024 | 0.055 | 0.118 | 0.140 | 0.379 | 0.488 |

### 6.3. Scheduling performance under different sampling scenarios

In this section, we investigated the performance of the MSA scheduler as a function of the number of sampled scenarios, given the experimentally measured distribution of the request prediction uncertainty from the work of Khosro Anjom et al. (2019). The simulation results Mean values of harvesting efficiency (points) and their 95% CI (shaded areas) as a function of the number of sampling scenarios are shown in Figure 22.

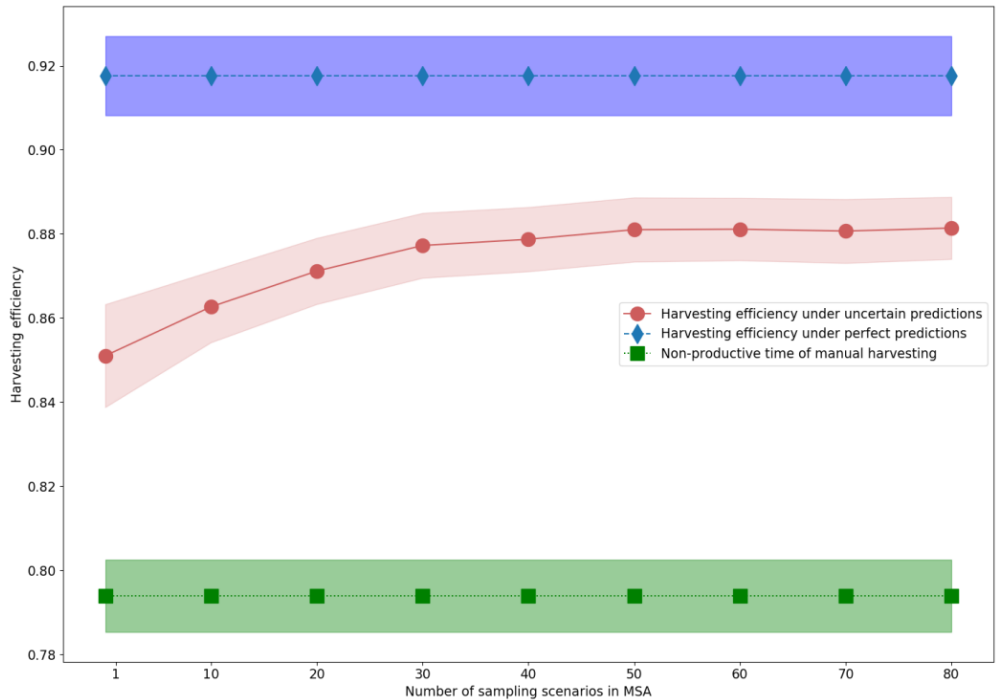

*Figure 22. Mean values of harvesting efficiency (points) and their 95% CI (shaded areas) as a function of the number of sampling scenarios, under uncertain transport request predictions. The harvesting efficiency (blue lines) under perfect transport request predictions and the manual harvesting efficiency (green lines) are also presented.*

From the results, one can see that as the number of scenarios in MSA increases, the mean harvesting efficiency (red curve) increases when the sampling scenarios are smaller than 50. However, when the scenario number is over 50, the efficiency and non-productive time come to a plateau. Tukey's HSD tests were used for different sampling scenarios {10,20,30,40,50} to examine their efficiency differences, which is presented in Table 12. The null hypothesis was that the mean efficiency of MSA in different sampling scenarios were the same. The alpha value for the tests was set to 5% (Type I error). From the combinations of {30, 40}, {30, 50} and {40, 50} in Table 12, one can see that the scheduling performance of MSA did not show significant improvement when the number sampling scenarios was over 30.

*Table 12. Tukey's HSD results of harvesting efficiencies for different sampling scenarios in MSA*

| Scenarios comparing combinations | | Adjusted p-values | Null rejections |
|---|---|---|---|
| 10 | 20 | 0.0012 | True |
| 10 | 30 | 0.0013 | True |
| 10 | 40 | 0.0010 | True |
| 10 | 50 | 0.0005 | True |
| 20 | 30 | 0.0022 | True |
| 20 | 40 | 0.0031 | True |
| 20 | 50 | 0.0020 | True |
| 30 | 40 | 0.7631 | False |
| 30 | 50 | 0.0752 | False |
| 40 | 50 | 0.1221 | False |

However, the mean of computation time for a schedule with 50 scenarios is approximately 5 seconds (on the Intel Core i7-3770@3.40 GHZ laptop used as a server), which is adequate for real-time operation. As a result, we set the number of sampling scenarios for the MSA to be 50.

## 6.4. Scheduling performance under experimentally derived prediction uncertainty

This section was to evaluate the performance of MSA under different robot/picker ratios. The number of sampling scenarios for MSA was set to 50. The harvesting efficiencies of robot-aided harvesting with stochastic scheduling (using the MSA) and deterministic scheduling (perfect predictions) are compared against all-manual harvesting in Figure 23. As expected, the MSA performed better than manual harvesting and worse than deterministic scheduling with perfect predictions. T-tests were applied for the case of 4 robots, as the confidence intervals of the mean efficiencies (banded areas in Figure 23) intersected. The null hypothesis was that there was no significant difference between the mean harvesting efficiency of stochastic and deterministic scheduling. The alpha value for the tests was set to 1% (Type I error). The evaluated p-value was 1.6e-7, so the harvesting efficiency was significantly better under perfect predictions.

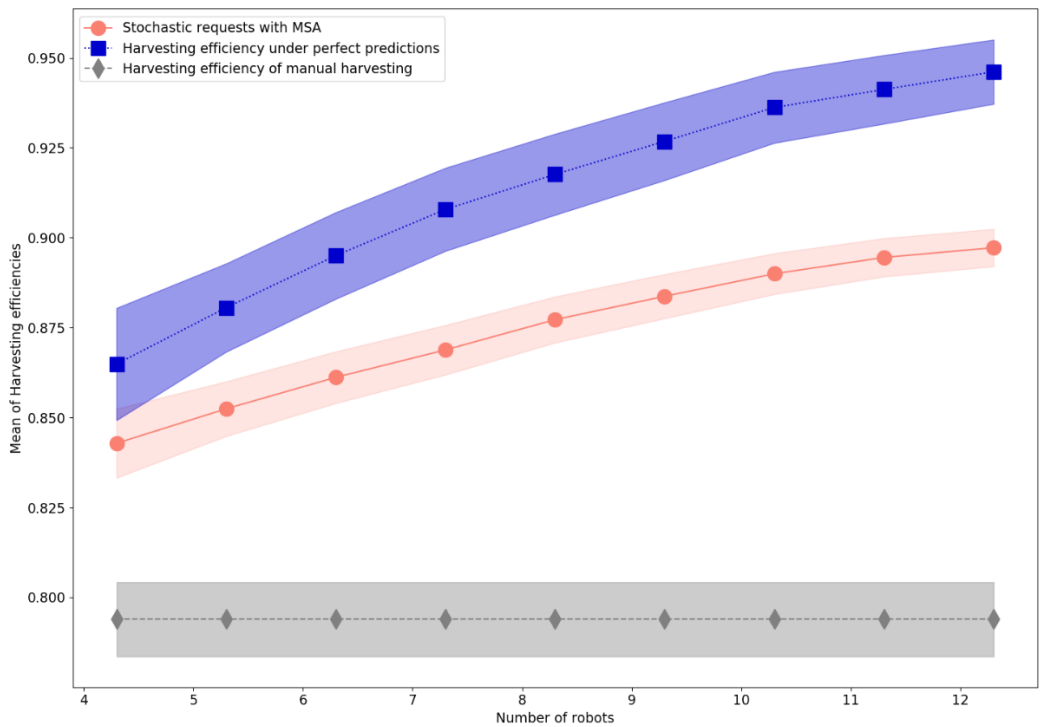

*Figure 23. Mean values of harvesting efficiency (points) and their 95% CI (shaded areas) as a function of the number of robots, for stochastic scheduling (with MSA), deterministic scheduling with perfect predictions and manual harvesting*

The number of trays per hour, $NumTrays_{hour}$, is used to denote the harvesting rate of the entire crew of pickers. Of course, given a fixed field size, the more pickers participating in harvesting, the higher the crew harvesting rate will be. In Figure 24, the red line shows how the $NumTrays_{hour}$ improves as the number of pickers increases (top X-axis is the number of pickers). Similarly, the crew harvesting rate $NumTrays_{hour}$ will increase when more robots (bottom X-axis is the number of robots) are introduced in the harvesting given a fixed crew size of 25 (shown as a navy line in Figure 24). One can observe that, for a picker crew size of 25, the crew's mean harvesting rate improves even when only a few robots are introduced compared to manual harvesting. The mean harvesting rate of 25 pickers with the assistance of crop-transport robots was similar to that of 30 pickers with manual harvesting, but obviously smaller than with 35 pickers. T-tests were conducted to compare the harvesting rates of a 25-picker crew working with 6, 7, and 8 crop-transport robots with the harvesting rate of a 30-picker crew doing manual harvesting. The null hypothesis was that there were no significant differences between the harvesting rate of 25 pickers under the assistance of 6, 7, and 8 robots and that of 30 pickers with purely manual harvesting. The significance level of p-values (alpha) for rejecting the null hypothesis (Type I error) was chosen as 1%. The calculated p values are shown in Table 13.

*Table 13. P values of T-tests that compare the mean harvesting rate of a 30-picker crew picking manually, against the mean harvesting rates of a 25-picker crew aided by 6, 7 or 8 robots*

| Robot numbers | 6 | 7 | 8 |
|---|---|---|---|
| P values | 0.022 | 2.3e-8 | 3.4e-17 |

The results show that when more than 7 robots were introduced, the $NumTrays_{hour}$ of the 25-picker crew using robots was significantly higher than that of the 30-picker crew without robots.

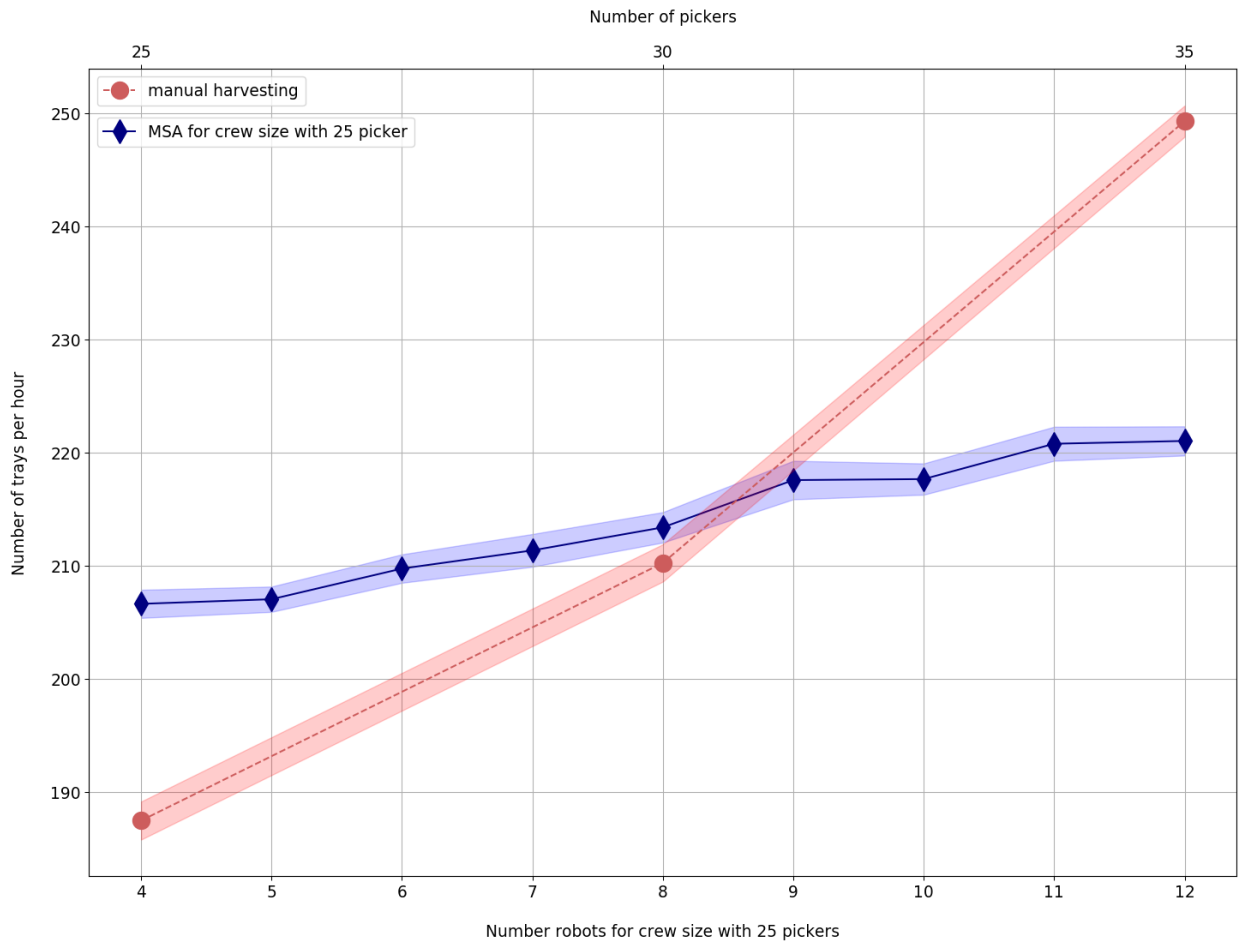

*Figure 24. Mean and its 95% CI of harvesting trays per hour for different crew sizes (red line) and harvesting trays per hour for 25 pickers but different number of crop-transport robots*

## 7. Summary and conclusions

In this part, robot-aided strawberry harvesting was developed further by considering more realistic conditions: uncertainty in the predicted transport requests and slower robot speeds. Predictive scheduling of crop-transport robots under stochastic predictive transport requests was investigated. Given slow/safe travel speeds for the robots, tray-transport request rejection is essential to guarantee that the harvesting efficiency with robots cannot be worse than the efficiency of all-manual harvesting, regardless of the number of robots deployed. We also evaluated the influence of uncertainty on the performance of the stochastic predictive scheduling, as well as the performance in typical uncertainties of prediction from previous work (Khosro Anjom & Vougioukas, 2019)

Experimental results showed that the scheduling performance without request rejections was significantly worse than the scheduling with request rejections when the robot picker ratio was smaller than 11/25. Under typical uncertainty of full tray prediction and a robot picker ratio of 1:3, the adapted MSA scheduling algorithm increased harvesting efficiency by more than 8% and was only 2.5% worse than deterministic scheduling under perfect predictions. When we introduce over 7 robots for a crew of 25 pickers, the harvesting rate of the whole crew under the assistance of crop-transport robots is significantly better than 30 pickers harvesting manually.

# Chapter 4   System design, implementation, and evaluation

## 1. Introduction

In this chapter, the implementation and integration of the co-robotic system and its deployment in a commercial strawberry harvesting operation is presented. In the envisioned crop-transport robotic aided harvesting, each picker enters an unoccupied furrow to start picking strawberries selectively from the plants on the raised beds on each side of that furrow. When a certain fill ratio is reached (Peng & Vougioukas, 2020), the picker will press a request push-button on their instrumented cart. The button allows a picker to decide for themself if they want a robot to carry their tray. For example, if a picker prefers to walk to deliver a tray - to take a break from stooped work - they can do so. The button also helps to establish a simple communication protocol between the workers and the robotic system: a transport request is initiated and is either accepted by the system or rejected. The automated weighing system is still used by the transport-request prediction module, after the button has been pressed. The scheduling system signals the picker (LEDs on their instrumented carts) if their transporting request will be served by a dispatched robot or rejected by the system. If the picker will be served, the dispatched robot starts from the active collection station, drives with an empty tray to the assigned picker's full tray location, waits for the picker to switch empty and full trays, and takes the full tray back to the active collection station, where it waits for the next dispatching. If a tray-transport request is rejected, the picker transports the full tray to the collection station, just like in manual harvesting.

The co-robotic harvest-aiding system comprises three sub-systems: instrumented carts, robots (aka FRAIL-Bots), and an operation server. The carts used by pickers weigh approximately 2.2 kg and our instrumented carts should have the same form factor and not be significantly heavier, to be accepted by pickers (4 kg max). The volume and weight constraints introduced a battery size – and available energy – constraint. Also, small amounts of data from each cart must be transmitted to the field computer/operation server over distances that span typical fields (> 300 m) at rates of approximately 1 Hz; communication must be bidirectional, since the scheduler must inform pickers if their requests are rejected or will be served. The robots must also communicate in real-time with the operation server to send their state and receive dispatching commands and reference paths. Robot collision avoidance on the headlands is performed centrally on the server and thus the robot-server communication system must have high-bandwidth. The complexity of the software running on the carts, robots and operation server requires a distributed software architecture that can provide real-time performance.

The system architecture is shown in Figure 25, and is described next in greater detail.

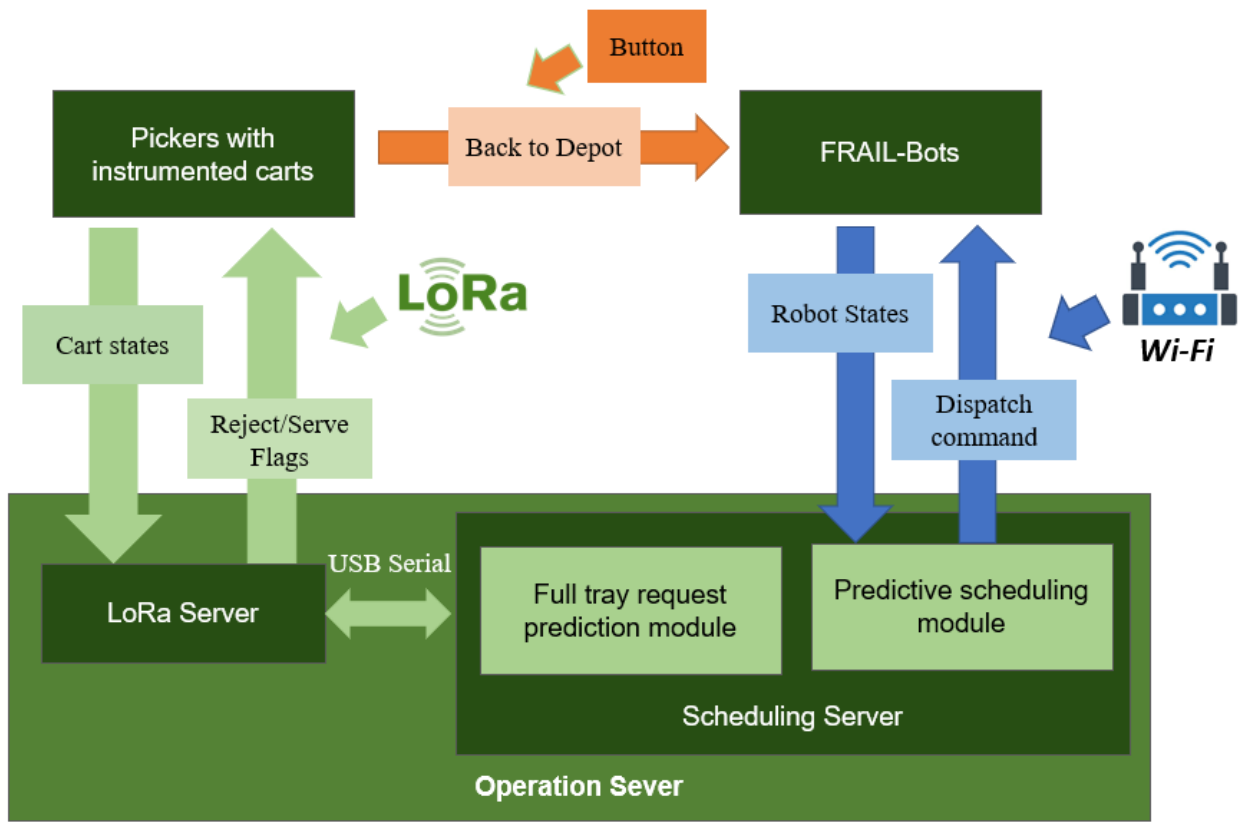

*Figure 25. Diagrams of harvest-aiding system components and their communications*

The data from instrumented carts are transmitted wirelessly to the server module that runs on a field computer at a collection station through LoRa ("LoRa", 2020). The LoRa server, scheduling server, and FRAIL-Bots communicate with each other using a ROS network (Figure 25) that utilizes different physical layers. The LoRa server module is connected through a USB cable to the scheduling server computer that works under the same Wi-Fi network with FRAIL-Bots. Each functional module in the ROS network is implemented and packaged into a ROS node which can subscribe and advertise the ROS messages from the other nodes. On the scheduling server, the tray request prediction module receives data from the cart and generates predictions of full tray requests. Given the subscribed ROS messages of robots' states and the predicted full tray requests, the predictive scheduling module calculates an optimized schedule. Then, the scheduling module publishes the dispatching commands, which are received by the available robots. After the robots arrive at the predicted full tray location, the picker will load their harvested full trays and take the empty tray from the arrived robot. Then they press a button on the robot to signal it back to the collection station. When the robots arrive back at the station, they will wait for the worker at the station to unload the full tray and replace it with an empty tray.

## 2. System components

### 2.1. Sub-system I: instrumented cart

The instrumented picking cart was modified and fabricated from a standard strawberry harvest cart (Figure 26). During harvesting, the cart system measures the mass of strawberries inside the tray located on the cart with two load cells underneath the supporting frame and receives the GPS coordinates from a GNSS receiver module (Piksi-Multi, Swift navigation, US) that receives WAAS (Wide Area Augmentation System) corrections. An IMU (BM160, Bosch,

German) is integrated on the Piksi-Multi to observe the instantaneous motion of the cart which is used to filter the tray mass measurements. The mass and GPS locations are transmitted to the scheduling server by a LoRa module (RFM96W LoRa Radio, Adafruit, US) sitting on the control board. A momentary contact button is installed for the picker to notify the system that he/she wants to be served by the robot.

The cart messages, composed of measured mass, GPS locations, and the button state, are transmitted at 1 Hz interval to the LoRa module on the server-side. The LoRa is a low-power wide-area network protocol, which uses license-free sub-gigahertz radio frequency bands ("LoRa", 2020). An SD card module installed on the control board is used to store all the sensor data during harvesting. Two LEDs (red and yellow) are used as indicators to communicate informative signals to the picker: a yellow LED represents that the tray transport request has been assigned to a robot; red LED means that the request is rejected, and the picker needs to transport the tray by themselves.

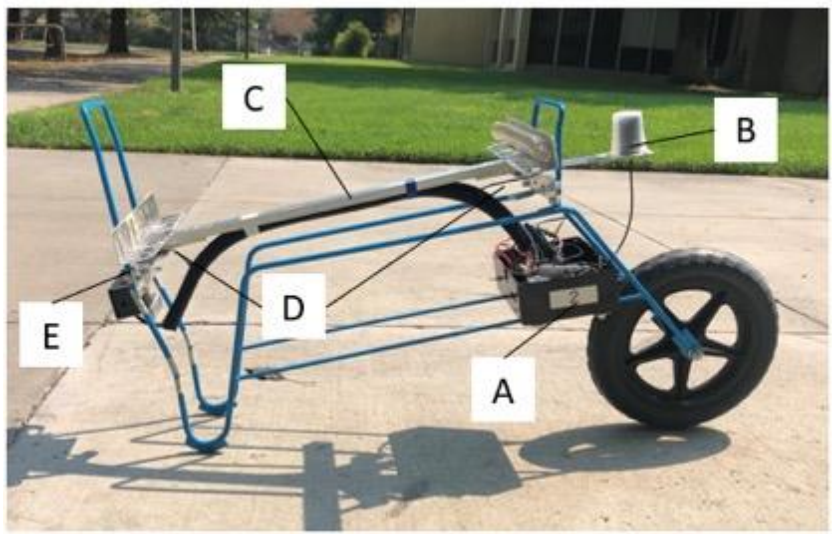

*Figure 26. The instrumented picking cart: A. Control box with Arduino Due, LoRa module, battery, and SD card logger inside, Piksi Multi GPS unit; B. GPS antenna; C. Supporting frame on the top of load cells; D. Load cells; E. Momentary push button, yellow and red LEDs.*

## 2.2. Sub-system II: FRAIL-Bot

Two identical crop-transport robots (aka FRAIL-Bots) were designed and built for this work (Figure 27). The bill of materials for each robot is approximately 10,000 USD; fabrication cost is not included. Constraints related to budget, available time and field deployment restricted the number of robots to two. Still, the developed approach is applicable – and can be tested – with two robots, and by using a crew size of six to eight people, reasonable robot-picker ratios can be used. The robots are designed to straddle the bed and occupy two furrows when driving inside the field. To avoid any interference of the robot with pickers in adjacent rows, pickers need to be spaced two furrows apart (with one empty furrow between them). Based on discussions with growers, and on the settings of the field experiments reported later, this arrangement was acceptable by the growers and the pickers.

Each robot works under supervised autonomy and its collaborative operation is governed by a finite state machine (Peng & Vougioukas, 2020). The hardware components for the FRAIL-Bot are labeled in Figure 27. The robot weighs approximately 50 kg, and it is driven by two DC motors with gearboxes and incremental encoders attached to the rear wheels (D and E). The steering system is integrated with two screw drives and angle sensors attached on their rotation axis (F and H). Two GPS module antennas (Swift navigation, US) are installed for getting the position and heading of the robot in open fields (C).  An emergency stop button (I) is installed on the side of the robot to stop the driving system.  A return button (E) on the front of the robot is used by the pickers to signal the robot that the full tray has been loaded and the robot must drive back to the collection station. The electronic devices including batteries, mini-computer (Intel

NUC, Intel Inc, US), driving motor controllers for rear-wheel motors, steering motor controllers, and two GPS modules are installed inside two wooden boxes (A and B).

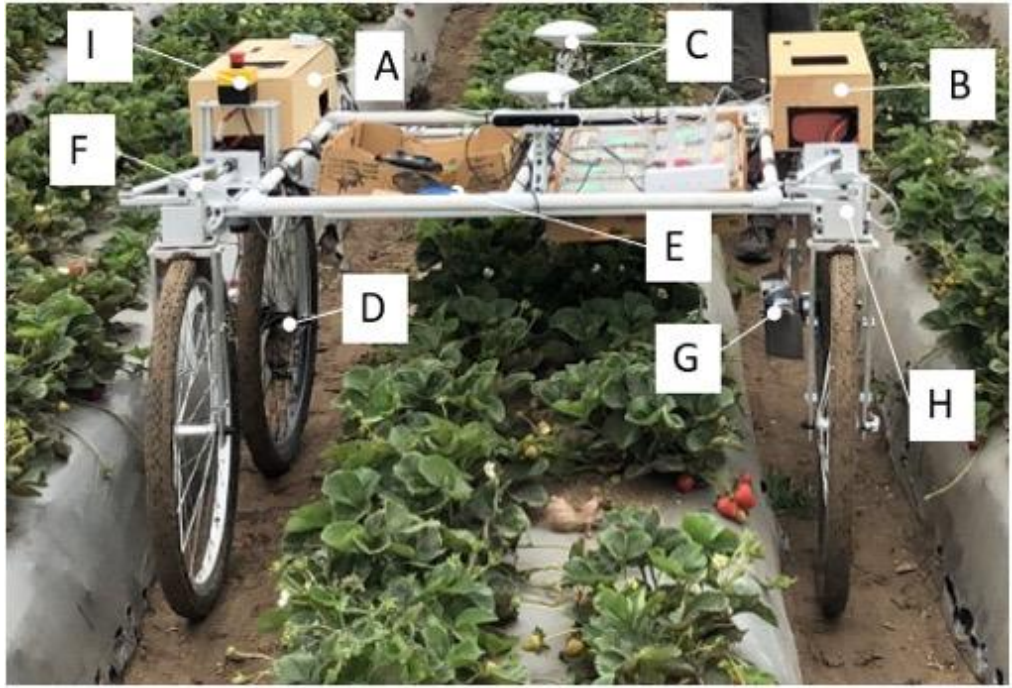

*Figure 27. Components of FRAIL-Bot: A. control box-I with a mini-computer, battery-I, motor controllers for the two rear driving motors and two steering motors B; Control box-II with two GPS modules; C. GPS antennas; D, G. DC motors with gearbox and incremental encoders; E. Return button; F, H. Steer-driving system; I Emergency button.*

The software architecture on each assigned FRAIL-Bot is shown in Figure 28. The FSM node first subscribes to the schedule message from the operation server including the dispatching time and dispatching location, from the operation server (explained in the section 3.2). When the dispatching time is reached, the navigation node generates the planning path from the current location of the robot to the assigned location inside the row. Given the planned path and the current robot location and heading, the path tracking module continuously outputs the control command to the motion control node, which converts the motion command of the robots to the commands of driving motors. A robot localization node fuses the subscribed sensor information, including GPS measurements, IMU data, and wheel odometry to estimate the pose and publishes the current pose of the robot into the ROS network.

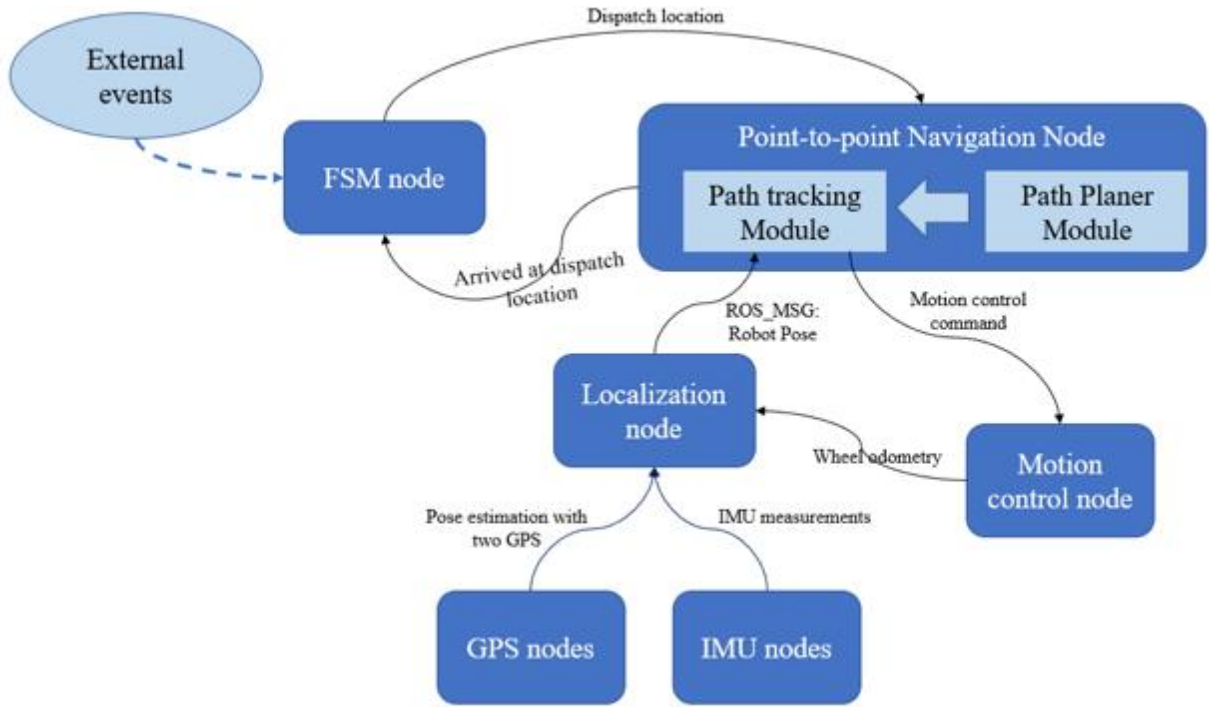

*Figure 28. The architecture of the FRAIL-Bot software under ROS*

The activity of FRAIL-Bot is guided by a finite state machine (FSM) introduced in our previous work (Peng & Vougioukas, 2020). The FSM implemented in the ROS node is shown in Figure 28. The state transitions are based on current states and designed external events (green oval in Figure 29) during the harvesting activity. Each FRAIL-Bot navigates autonomously from the collection station to the picker it will serve after receiving the dispatching command from the predictive scheduler. Upon arrival at the dispatch location, the robot waits until the picker fills their tray (if it is not full upon arrival), loads the full tray, takes an empty tray from the robot, and presses the momentary contact button to command the robot back. The robot navigates back to the collection station and enters the state "IDLE_IN_QUEUE", i.e., it waits for a worker at the collection station to remove the full tray and place an empty tray on the robot. Afterwards, this worker presses the momentary contact button to have the robot transit to the "AVAILABLE" state, where it waits to be dispatched again.

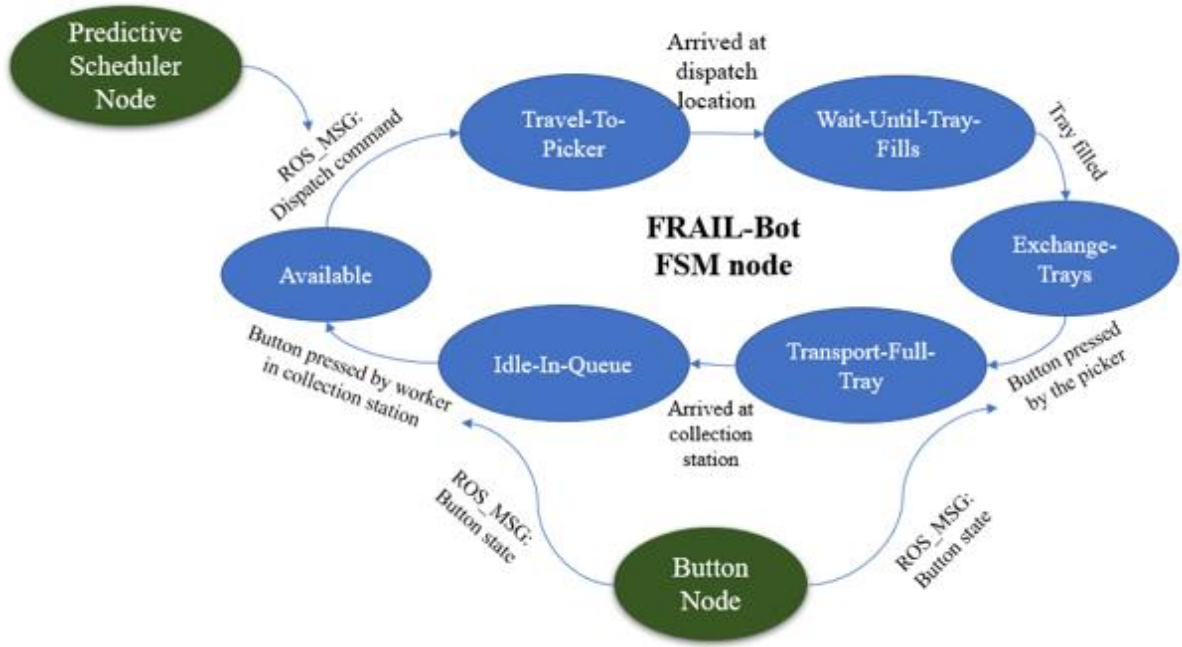

*Figure 29. FSM of FRAIL-Bot in the harvest-aiding system*

### 2.2.1. Motion control

The hardware diagram for the motion control node of the FRAIL-Bot is shown in Figure 30. A dual-channel motor controller is used to drive the two rear motors by UART serial communication. The left and right steering systems are driven by two DC motor control boards (1065B, Phidget Inc, Canada) with the PID controllers based on the feedback from angle encoders.

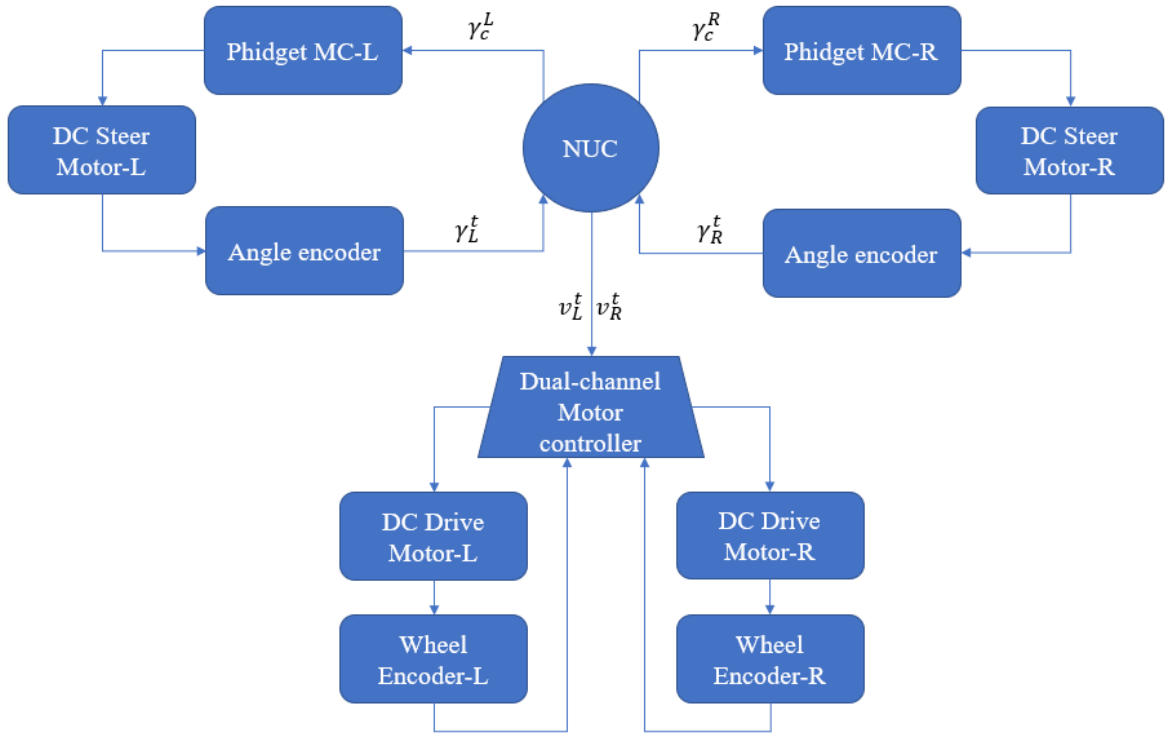

*Figure 30. Hardware diagram for FRAIL-Bot motion control*

The Ackerman model (Figure 31) was applied for the motion control of the FRAIL-Bot to have it maneuver smoothly (minimum skidding) in the field navigation. Given the motion command, linear velocity ($\boldsymbol{v_c}$) and steer angle ($\boldsymbol{\gamma_c}$), the steer angle command ($\gamma_c^L, \gamma_c^R$) for left and right steering systems can be calculated using the Ackerman model.

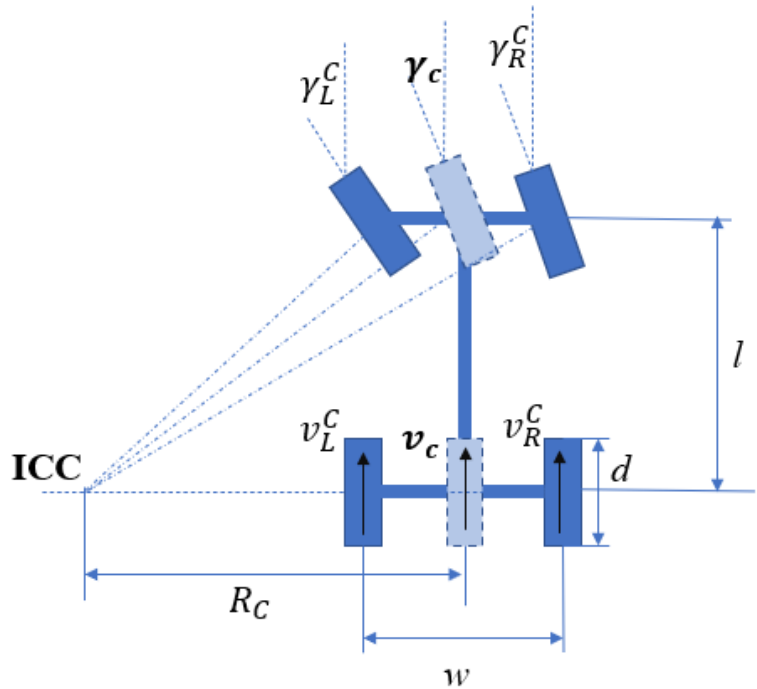

*Figure 31. Ackerman model for the calculated control command of FRAIL-Bot*

The response time of the steering angle control system is slower than the response of the speed control system of the wheel driving motors. Therefore, the wheel motor speeds are commanded to "follow" the robot rotational velocity that corresponds to the sensed steering angles of the front wheels (cascade control). The real-time control commands for each motor are calculated based on the motion control command ($\boldsymbol{v_c}, \boldsymbol{\gamma_c}$) shown in Figure 32. The rear motor commands ($v_c^L, v_c^R$) are calculated from the current steering angle ($\gamma_c^t$) and the linear velocity command ($\boldsymbol{v_c}$).

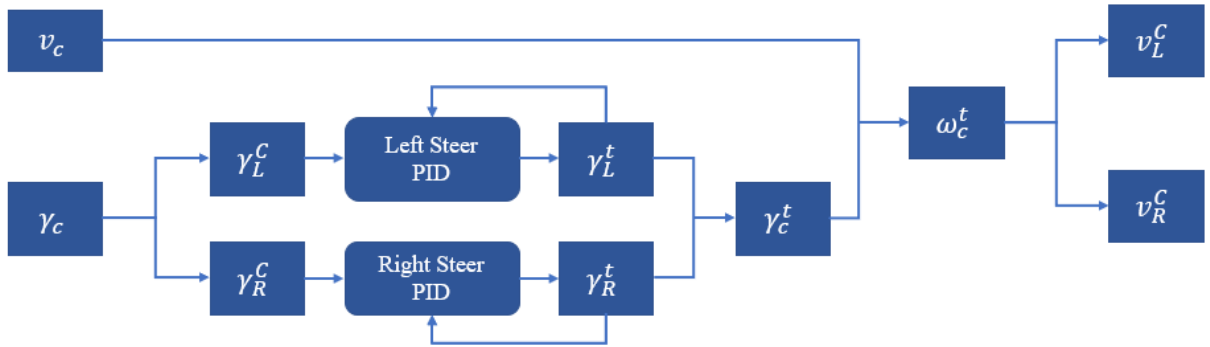

*Figure 32. Motors control in the motion control of the FRAIL-Bot*

## 2.2.2. Localization

The localization and heading of the FRAIL-Bot is mainly dependent on the RTK (Real Time Kinematic) solutions of two GNSS modules on the FRAIL-Bot. The GPS antennas on FRAIL-Bots are above the strawberry plants and the strawberry field is normally in an open area without nearby high buildings or tree canopies which might attenuate or occlude satellite radio signals. Thus, the RTK solutions are stable for most of the time. The Internet-based (NTrip) RTK system was used to get the geodetic solution of the rear GPS antenna under the transmission protocol of Radio Technical Commission for Maritime (RTCM) (Weber et al., 2005). The NTrip casters used in California are managed by UNAVCO community service (Bendick, 2012). As the GPS location of the NTrip caster is well surveyed on the geodetic frames before releasing, the RTK rover solutions corrected by the same base station in a mapped field are consistent over time. The front GPS also works in RTK mode with the real-time corrections from the rear GPS that works as the moving base station based on the same satellites' observation over their antenna attitudes (Swift Navigation, 2020). Using this feature, the attitude of the robot, heading angle, in the geodetic frame from the front GPS solution was obtained. Combining these two results, the complete localization solution of the robot in 2D space can be obtained in a mapped field.

The localization node of the FRAIL-Bot is implemented with the EFK node on the ROS package, named robot-localization (Moore & Stouch, 2016) by fusing the measurements of GPS modules, wheel odometry, and IMU as shown in Figure 33. The sensor information, including 2D pose estimation from two GPS modules, wheel odometry on the robot frame (base link frame), acceleration, and angular velocity on the IMU frame are packaged and published as ROS messages. The robot localization node works to combine the motion model and transformed sensor readings with EKF.

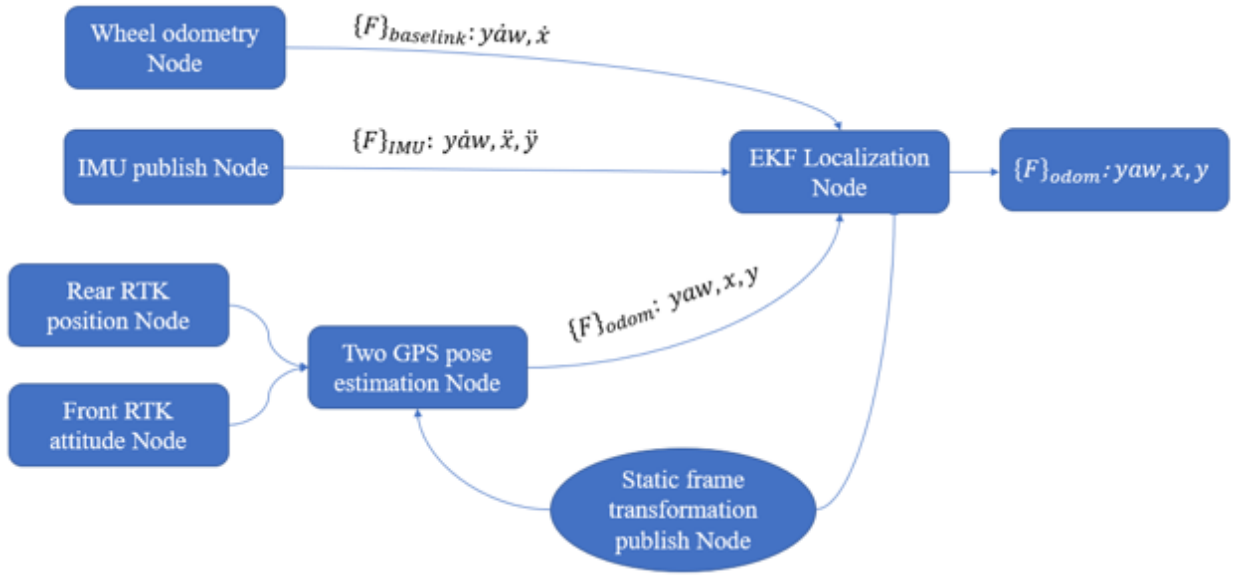

*Figure 33. FRAIL-Bot localization with robot-localization EKF node*

### 2.2.3. Field navigation

To transport trays during harvesting, the FRAIL-Bots travel back and forth from fixed "parking" locations next to the collection station on the headland to dynamic locations inside the field. A field map (Figure 3.b) with the geometric information of the plant beds and the collection station is essential for the FRAIL-Bots to generate feasible paths to their serving locations. The map is built with the RTK system and includes the end of bed points in the vicinity of the headland and pre-allocated collection stations.

To implement point-to-point navigation in the field, the path planner needs to generate a smooth path that allows the robot to enter the row with heading parallel to the row. A speed profile is also generated to vary the robot speed at different sections of the path. When driving on the headland the robot must execute maneuvers of large curvature and thus must travel at low speed. Also, when the robot is near a picker, the robot needs to run at a low speed, for safety purposes. Inside the rows, the robot can run at a higher speed. The path is planned as soon as the robot receives the dispatching command (Figure 34).

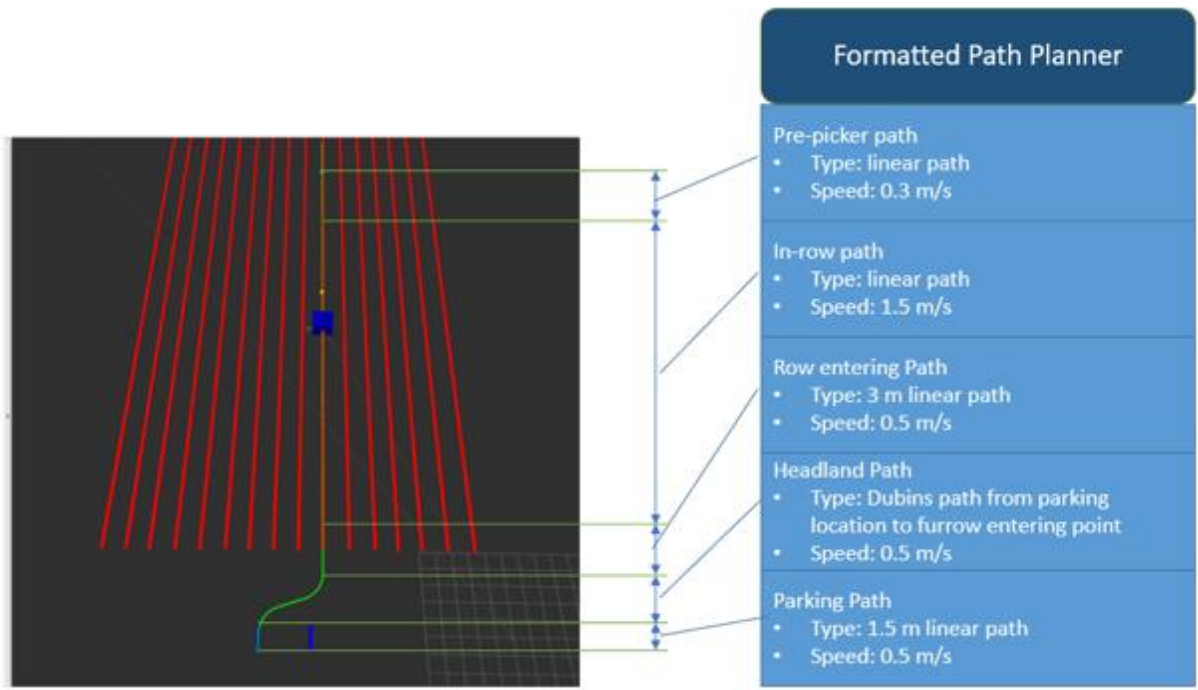

*Figure 34. An example of a formatted planning path from the robot parking location at the depot center to the dispatching location inside the row*

A pure-pursuit algorithm was used to have the robot track the path. The algorithm parameter, look ahead distance, is well-tuned and set differently given the designed speed profile. The look-ahead distance was tuned given different planned speed along the trajectory.

## 2.3. Sub-system III: operation server

The hardware of the operation server sub-system is composed of two parts: a LoRa server board and a scheduling server computer (Figure 35). The LoRa server board is connected to the server computer by a USB cable. It collects the cart states from the distributed instrumented carts in the field and publishes the received states as ROS messages. The FRAIL-Bots publish their states in the ROS network through a local Wi-Fi. The scheduling server module, running on the server computer, integrates the cart states and robot states to formulate and publish an online schedule message in the ROS network. FRAIL-Bots directly subscribe to the ROS schedule messages and execute the dispatching decisions. The LoRa server module also subscribes to the ROS schedule messages and transmits them to each instrumented cart through LoRa.

### 2.3.1. LoRa server module

The messages of each cart need to be reliably transmitted to the scheduling server in a high enough frequency ($\geq 1$ Hz) to build an online spatiotemporal full tray request (Khosro Anjom et al., 2019). The same LoRa module on the cart is used on the server. The module has adjustable parameters for bandwidth, spreading factor, code rating, and transmission power. The parameters need to be adjusted based on different application scenarios (Cattani et al., 2017). For our case, the module needed to transmit real-time data (25 bytes) from the cart to the collection station, across the whole harvesting block with a relatively high update rate (no less than 1 HZ). Based on our testing, this group of setting parameters for the RFW9X module meet our requirements well: bandwidth at 250 kHz, spreading factor at 6, code rate 4, and transmission power at 14 dBm. The data transmission covers the whole field block with low latencies. On the server side, the same module was used to receive the transmitted cart messages.

During transmission, each module can only work in a single channel, which means that the server can only get the message from one cart even if multiple carts send their messages at the same time. Our goal was to have the communication channel of the server evenly shared by all the carts in the time space. Based on our testing, the transmission of each cart state to the server takes around 100ms, so given $N$ carts, the server takes at least $N*100$ms to receive the cart states from all the carts once. If the size of the picking crew is over 10 (a typical number is 20 to 25), a LoRa gateway with multiple channels would be needed to meet our transmission requirements. In our experiment, we deployed two robots and a picker crew with 6 pickers, so a single channel LoRa server was used. A centralized time-split network communication protocol was developed in our work, as shown in Figure 35. Each period, the server broadcasts a reporting 'command' which is a short message containing the ID of a cart and a 'serve' or 'reject' flag updated from the predictive scheduler on the server computer. After that, the LoRa server waits for 100ms for the polled cart to report their cart states. The cart with that ID broadcasts its message after receiving the command message. The other carts will ignore the messages on the channel and keep waiting for their ID to be requested to report their states. If the server received the message from that cart or the wait time expires, it will move on to the next cart.

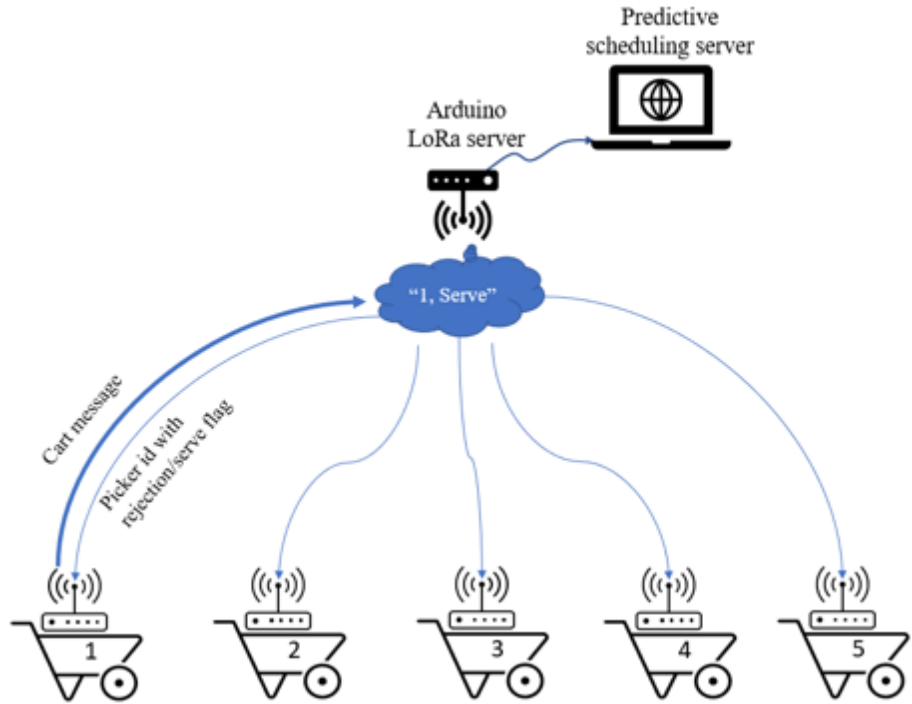

*Figure 35. The centralized topology of communication between LoRa modules on each cart and the server*

The above-described functionality of the LoRa server module is packaged into two ROS nodes (Figure 36):

a) Cart-states-pub node: It receives the data of cart states from the LoRa module on the server board, packages them into ROS messages, and advertises them to the server computer through a USB cable.

b) Server-reject-sub node: It subscribes to the 'serve' or 'reject' flags advertised by the predictive scheduling node and transmitted to each cart via LoRa.

## 2.3.2. Scheduling server module

Each functionality of the scheduling server module running on the server computer is packaged into a single ROS node. This module plays a centralized role in the whole system shown in Figure 36. The nodes running on the server computer are briefly explained as follows.

a) Tray-request-prediction node: It subscribes to the cart messages from the LoRa server board and updates the prediction of picking parameters, harvesting rate, and moving speed while picking. The predictive requests are generated and published on the ROS network when a certain fill ratio of the tray is reached, and the request button is pressed by the pickers.

b) FRAIL-Bot-scheduling node: It subscribes to robot states from the "FSM node" on each FRAIL-Bot and predictive tray transport requests from the tray-request-prediction node. Given them, this node runs a stochastic predictive scheduling algorithm and advertises a dispatching command to the FRAIL-Bots, as well as the rejection and serve flags to the LoRa server boards. The online solver of this node is explained in chapter 2.

c) FRAIL-Bots coordination node: It functions as the traffic management for the robots in the shared area of the headland, which is explained in section 2.3.2.2.

d) Operation visualization node: This node subscribes the messages from multiple nodes of different modules for visualization of the cart/robot states. It also provides some user interface to tune the parameters during the field operation, which is explained in section 2.3.2.1.

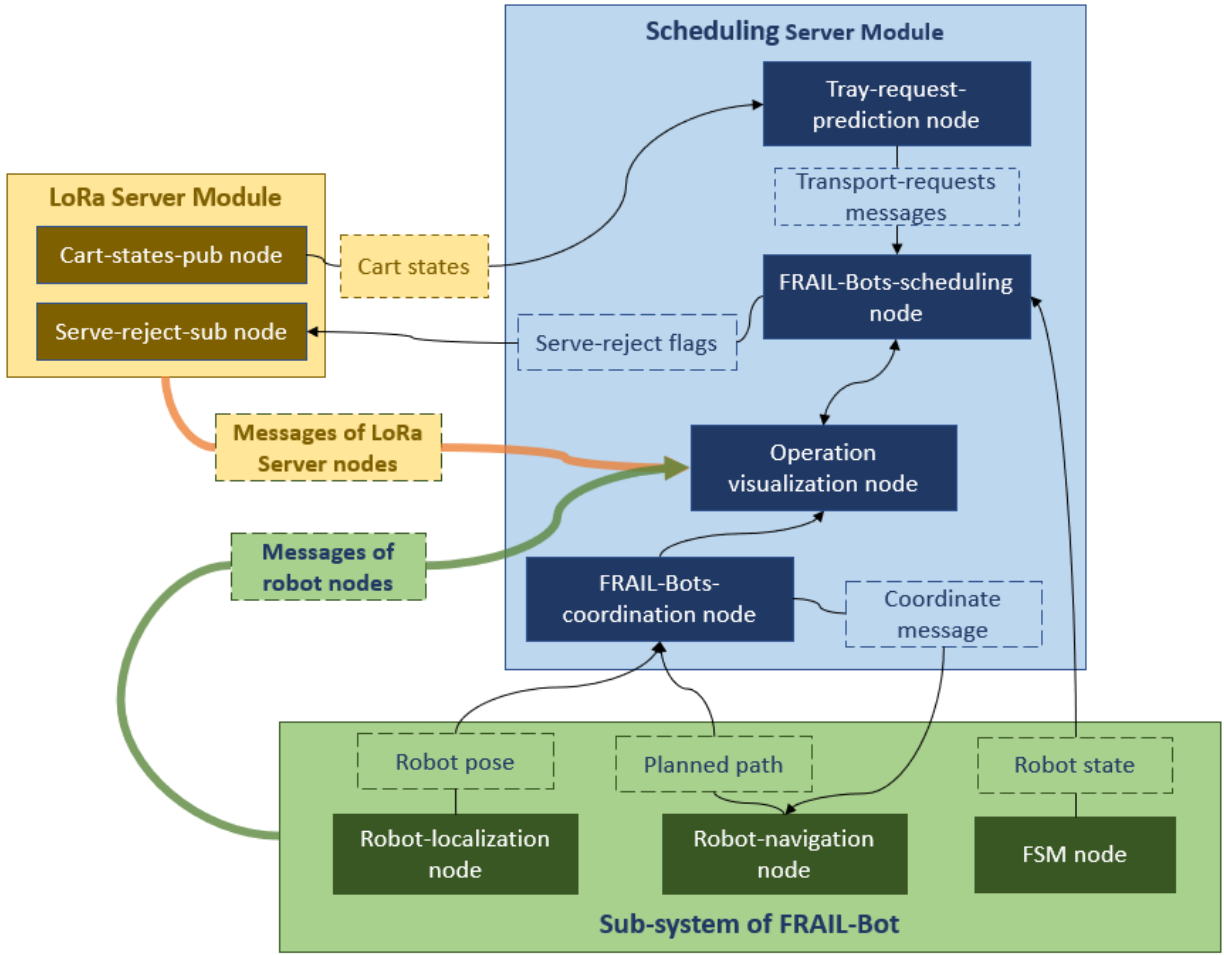

*Figure 36. The system architecture of the operation server modularized as ROS nodes.*

### 2.3.2.1. Operation visualization node

The user interface of the operation visualization module is shown in Figure 37. The sub-plot of "Field Map" shows the current locations of FRAIL-Bots and instrumented carts; the subplot of "Measured Weights" displays the measured weight of each cart over time. The other states of the carts are represented with four flags: "Full-tray", "Request", "Serving", "Reject". The "Full-tray" box turns green for 4 seconds after the picker lifts the full tray from the cart. The "Request" box turns green if the requesting button on that cart is pressed. The "Serving" box is changed to green when a FRAIL-Bot is dispatched to that cart and "Rejection" is changed to green when the scheduling server rejects the transport requests of that cart.

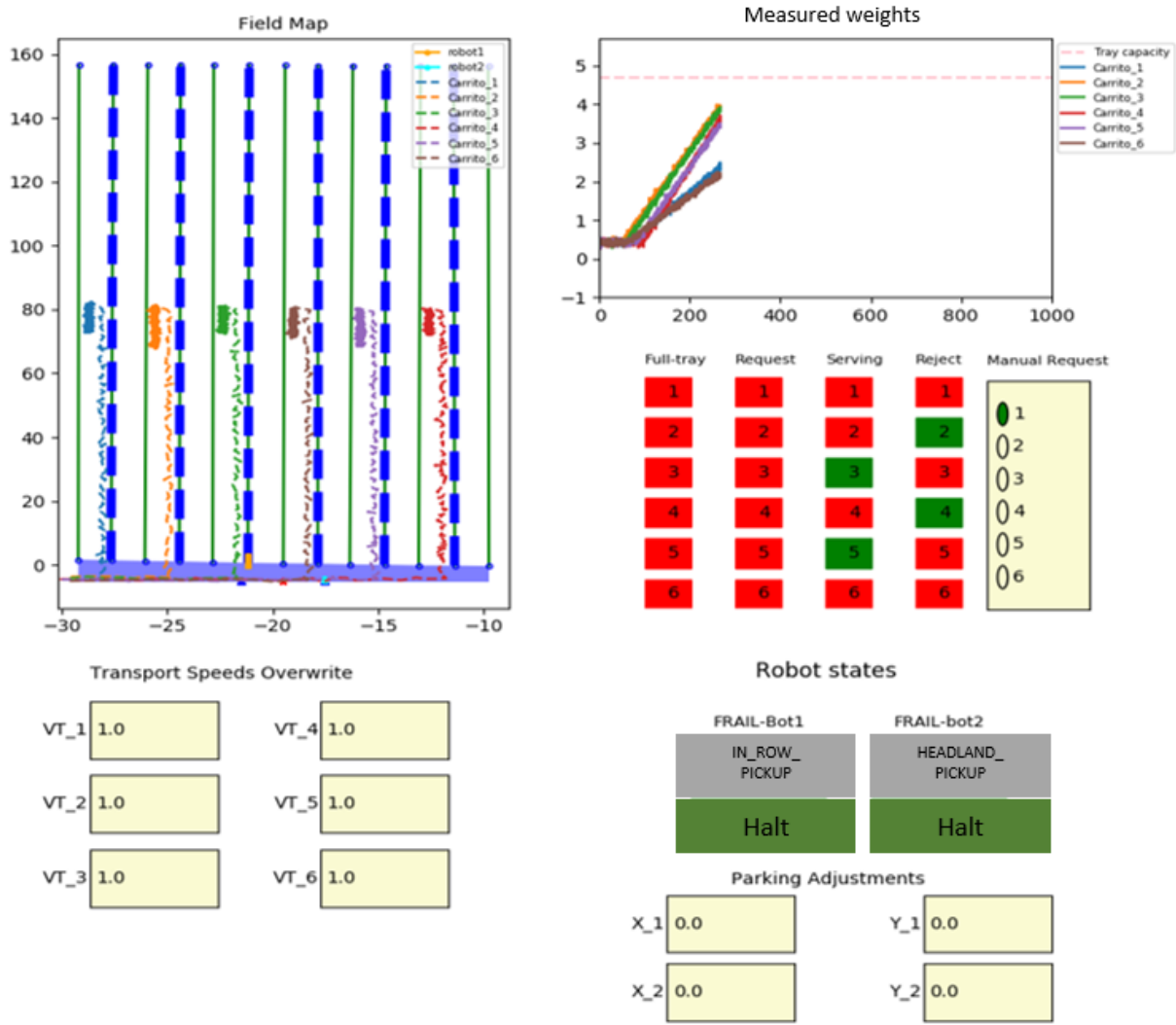

*Figure 37. System visualization for the states of instrumented carts and FRAIL-Bots and user interface for the parameters tuning.*

### 2.3.2.2. FRAIL-Bots coordination node

As the field headland space is shared by the robot team, a coordination node is required to prioritize the motion of one robot to solve the trajectory intersection. In this work, the harvest efficiency improvement was investigated after introducing the harvest-aiding system in the studied case. The function of the FRAIL-Bots coordination node is mainly to coordinate the robots so that they do not collide with each other when their paths are projected to intersect on the shared headland space.

The coordination node made the decision on which robot moves earlier based on the received planned trajectories and current locations of the two robots. When robots are driving on the headland, the robot coordinator node builds polygons encompassing the future paths from the robots' current pose to the end of their paths in the headland space, as shown in Figure 38. If the polygons intersect and one of the two robots has entered the intersected area, the coordinator will prioritize the robot inside the intersection area to go out of the intersecting area to avoid collision. As it is shown in the example in Figure 38, Robot1 has entered the intersection area while Robot2 has not yet. Given the planned paths and the robots' current poses, the coordinator predicts two instants for the robots: when they enter $(t_{r1}^1, t_{r2}^1)$ and exit $(t_{r1}^2, t_{r2}^2)$ the intersection area relative to current time instant. As Robot1 (blue block) has been inside the intersection area, $t_{r1}^1 = 0$ and $t_{r1}^2$ is predicted given the planned path. If the two robots' time intervals $[t_{r1}^1, t_{r1}^2]$ and $[max(0, t_{r2}^1 - margin), \ t_{r2}^2]$ intersect, the robot outside the intersection area (Robot2 in this case) will be commanded to stop before entering the intersection area, and wait until the robot time intervals do not intersect. A safety margin of 5 seconds is set when calculating the entering instant of the robot outside the intersection area.

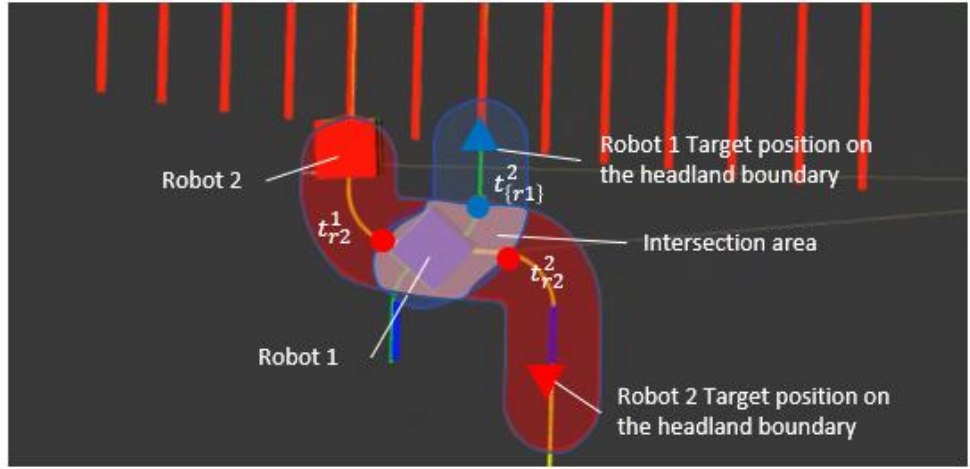

*Figure 38. Robots' coordination with online built encompassing polygons of their paths visualized on ROS-RVIZ*

# 3. Experimental design

The main goal of this chapter was to evaluate the harvest-aiding system with a crew of professional human pickers in commercial strawberry harvesting. Given the complexity of the system, subsets of the system were tested in multiple steps. In a first step, the implementation of the predictive scheduling module and the physical FRAIL-Bots was tested on a campus field, without actual pickers or carts; simulated tray-transport requests were generated and used to dispatch the actual robots. This is referred as "*Robot-in-the-loop Simulation Experiment*". In a second step, the instrumented carts were introduced, and lab members acted as pickers who collected fake strawberries from the ground to generate real weight for the carts and real tray-transport requests for the scheduler. This is referred to "*Robot-and-cart experiment*". The first two experiments were done on the same campus field, near the Western Center for Agricultural Equipment (WCAE) with ridged beds resembling the beds and furrows in commercial strawberry fields (Figure 39). Finally, the whole system was deployed and tested during commercial strawberry harvesting, with professional pickers, near Lompoc, CA; a crew of 6 pickers harvested a field block with the assistance of two robots. This is referred as '*Commercial field system experiment*'. For the campus experiment, all participating lab members were updated with departmental field safety training and had read a safety guidelines document prepared for this work ("Safety Guidelines for Data Collection for FRAIL-Bot"). The field experiments were conducted under the UC Davis Institutional Review Board (IRB) compliance protocol "IRB 575389-8".

As mentioned earlier, the FRAIL-Bot is designed to straddle the bed and occupy two furrows when driving inside the field. In all experiments, pickers were spaced two furrows apart (with one empty furrow between them). In a typical harvesting activity, they start from the midpoint of the row and move toward the collection station. Based on our previous work (Peng, Vougioukas, 2020), the FR threshold for requesting tray-transport service was selected at 70%, corresponding to pressing the call-robot button when 6 out of 8 clamshells became full. The predictive scheduler only served pickers who pushed the button.

## 3.1. Experiments on campus

### 3.1.1. Robot-in-the-loop simulation experiment

The functionality of the predictive scheduling module and FRAIL-Bots were first tested in a campus field with simulated pickers (Figure 39). The navigation module of the FRAIL-Bots was tested by inspecting the tracking errors while following designated paths from the collection station to locations inside rows. The activities of the simulated pickers ran on the server laptop with their cart messages published on ROS. These cart messages were processed by the predictive scheduling module in real-time (Peng et al., 2020).

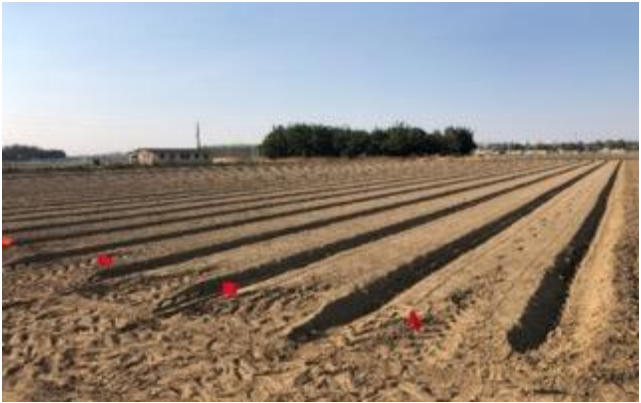 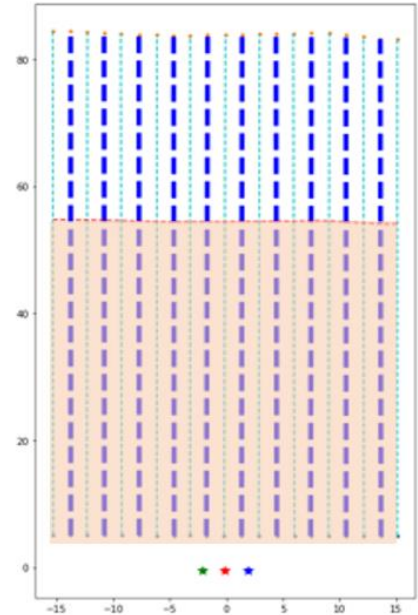

*Figure 39. a) Testing field block with raised beds near the campus of UC, Davis; b) schematic figure of the strawberry harvesting field block with two sections (upper and lower); furrows; plant beds; field split line, and collection stations.*

The field had twenty 80-meter-long rows and one collection station on the headland. Two FRAIL-Bots were deployed to serve simulated picking crews of sizes 4,5,6,7,8. The picking crew was simulated to harvest the shaded area (50-meter length) of the field block shown in Figure 39.b. Each crew size was evaluated once, and approximately 140 trays were collected from the harvested areas. Pickers' harvesting parameters were sampled from the histograms in Figure 10, Figure 11, and Figure 12 and their states were updated with the FSM defined in Chapter 3. The experimentally derived prediction uncertainty in Chapter 3 was used for the transport requests. The MSA algorithm ran on the predictive scheduler to dispatch FRAIL-Bots to serve these simulated pickers. The cart messages and robot states were saved into a ROS bag during the experiments. This experiment was to examine the implementation of the developed functional modules of sub-system II and III in the actual field environment. The evaluation metrics, $\Delta T^{fe}$ and $E_{ff}$, were used to evaluate the performance of the robot-aided harvesting.

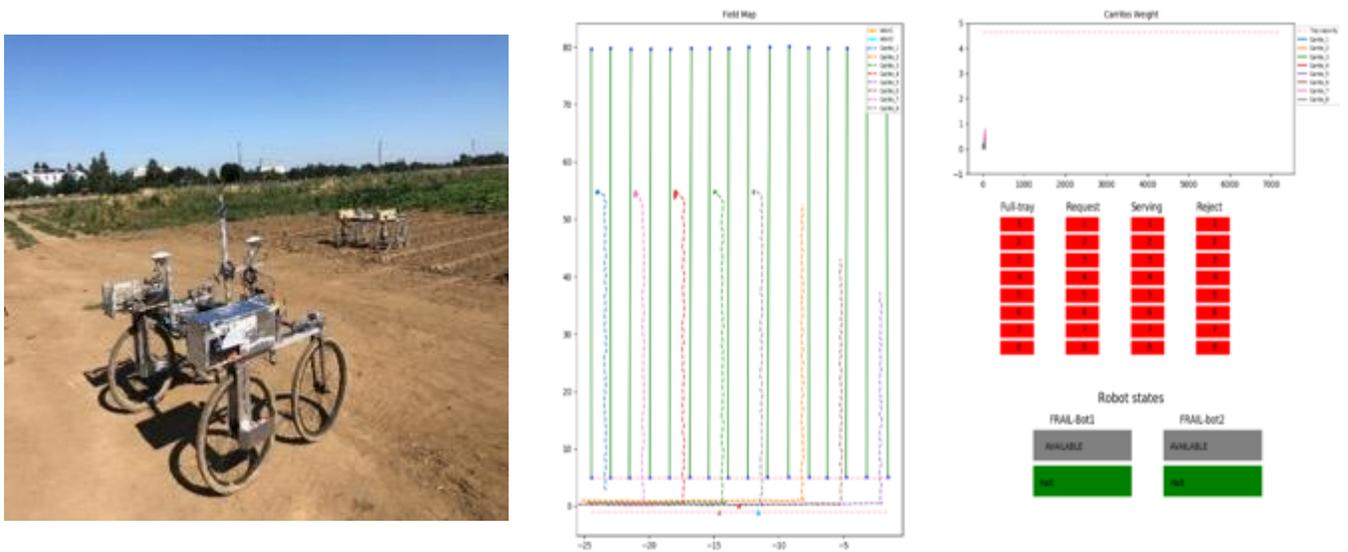

*Figure 40. (a) Two FRAIL-Bots are running in the testing field block; (b) The visualizer of the scheduling system to supervise the state of robots and simulated instrumented cart messages.*

### 3.1.2. Robot-and-cart experiment

In this experiment, we mainly investigated the implementation and integration of the whole system. Also, the intermediate functional modules of sub-system I and III were evaluated including the tray weighing system, transport request prediction module, as well as the communication system depicted in Figure 25.

After testing the scheduling module and FRAIL-Bots, the instrumented carts were integrated into our testing on the campus field, where 20 trays of red-painted rocks were distributed on the ground to emulate ripe strawberries (Figure 41.a). Four human pickers mimicked the picking activities of professional pickers by gathering the rocks and placing them in clam shells on their tray (Figure 41.b). One person stayed at the collection station to switch the empty tray with full trays transported by the robots. The pickers pressed the request button when 6 clam shells on the tray became full. The predictive module on the server side kept buffering the subscribed cart messages and used the collected data ahead of the button pressed instant to make the prediction of a full tray with the simple linear regression model (Khosro Anjom & Vougioukas, 2019).

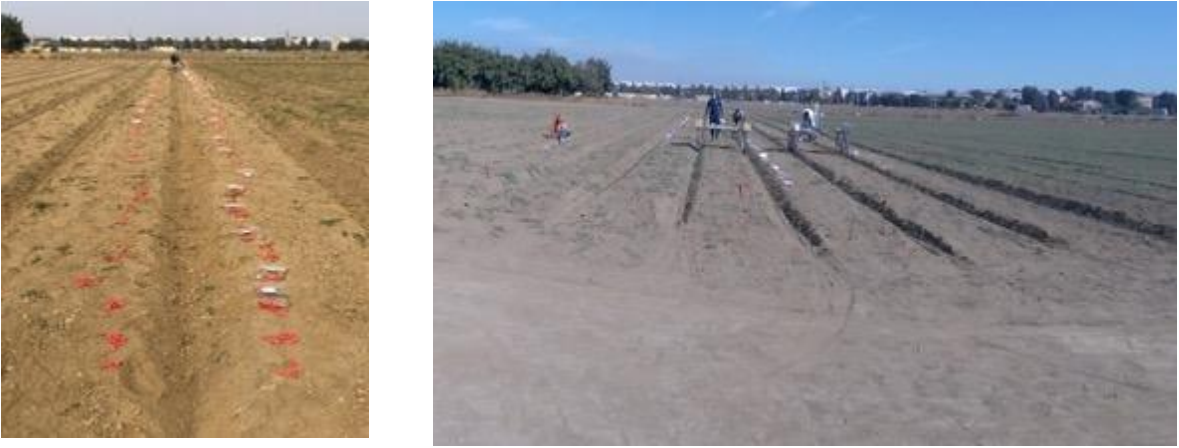

*Figure 41. a) Fake strawberries distributed on the testing field block; b) Four pickers and two FRAIL-Bots are working together to "harvest" on the testing field*

## 3.2. Commercial field system experiment

On Nov 10th and Nov 11th, 2020, the harvest-aiding system was evaluated with a crew of six professional pickers in a commercial field near Lompoc, CA shown as Figure 42. Each day, the pickers' working schedule was divided into 2 sessions. The first session was from 8:00 am to 11:00 am and the second session was from 11:30 am to 2:30 pm. In the first session, on November 10th, our system was set up and tested on the mapped field while the crew harvested the orange-colored field block (Figure 42b) in their usual manner, using our instrumented carts. The pickers collected the strawberries in 500-gram carton box trays. The gross mass of a full tray was around 4.5 Kg (10 lbs). Their harvesting data was collected and saved on the SD card modules of the carts. The robot states were saved on the server laptop. Data was processed based on the methods described in Section 3.3. From the data in that session, their walking speeds when transporting full trays were estimated. This data was used to estimate the performance of manual harvesting.

In the second session of November 10th and first session of Nov 11th, all 6 pickers started harvesting from the field's middle line and moved toward the unloading station in their typical harvesting manner. The crew harvested with the assistance of two FRAIL-Bots (blue area for Nov 10th and red area for Nov 11th on Figure 42b). The harvesting data using the co-robotic harvest-aiding system were recorded into ROS files on the server laptop, as well as in the SD cards of the carts. The evaluation results of our harvest-aiding co-robotic system (Section 2.2) were obtained from these two sessions. During robot-aided harvesting the pickers were asked to press the request button on the cart, when 6 of 8 clamshell boxes were full on their trays. It was explained to them that if their transport request was accepted by the robots, the yellow LED on their cart would turn on. In this case, they were instructed to wait for a robot, in case they filled their tray and a robot had not arrived. The scheduling system dispatched robots to serve the requests by solving online the stochastic predictive scheduling problem. The FRAIL-Bots were scheduled and dispatched to the predicted full-tray locations inside the rows. Upon arrival at the commanded location, the robots would stay still until the picker placed their tray on the robot and pressed a button that sent the robot back to the collection station.

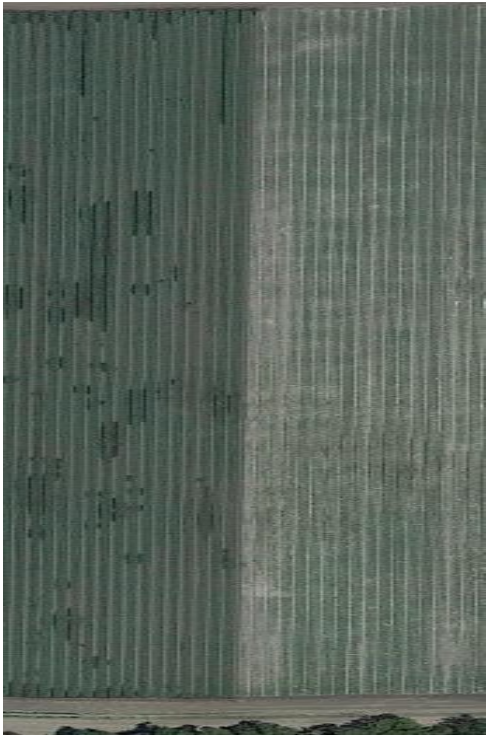 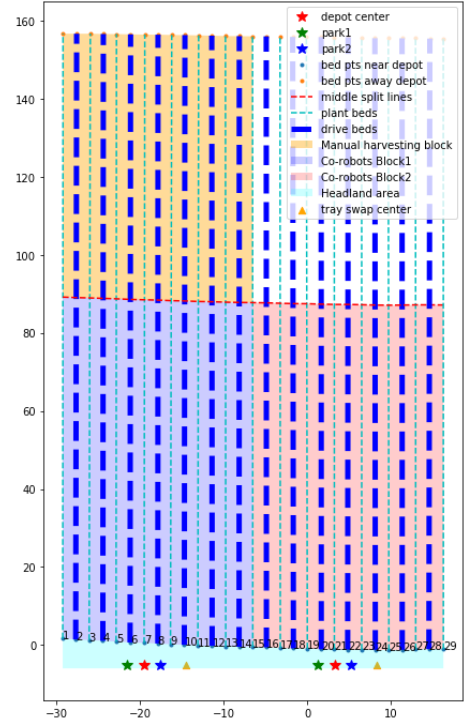

*Figure 42. a) Satellite picture of commercial field block near Lompoc, CA from Google Maps; b). Map of the field block built with RTK: blue and orange shaded areas were evaluated on Nov 10ᵗʰ; red and white shaded areas were evaluated on Nov 11ᵗʰ*

The system components are shown in Figure 43. The full-empty tray swap location was approximately 5 meters away from the depot center where the server laptop locates (red star in Figure 42.a), on its right-hand side, facing the field.

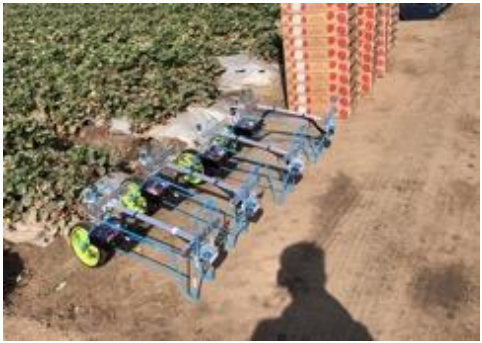 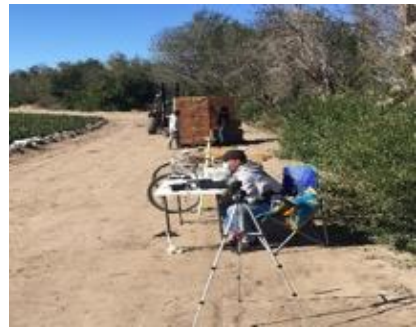

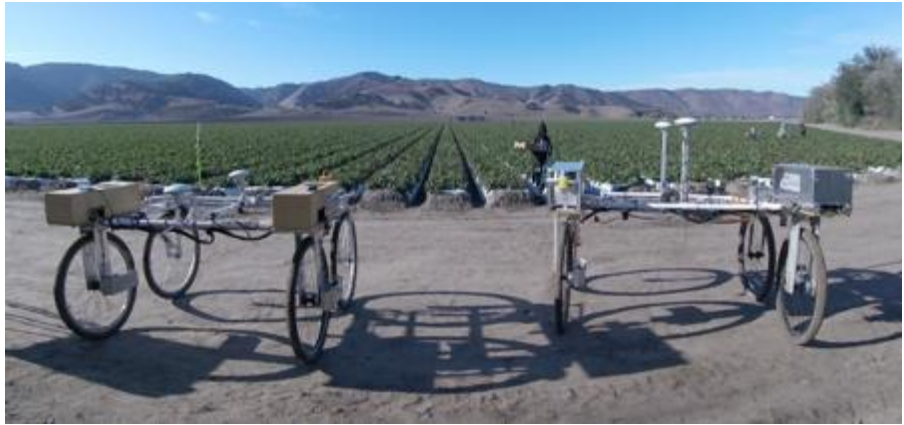

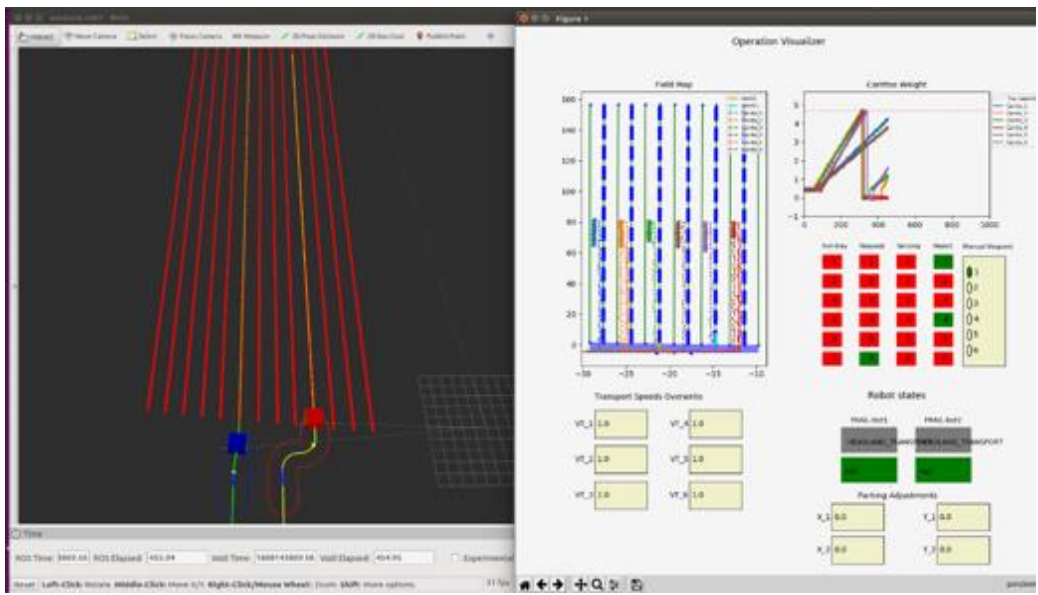

*Figure 43. Components of robotic harvest-aiding system: a) Instrumented carts; b) Collection station (aka, depot center) with scheduling server; c) Two FRAIL-Bots parked at the depot center; d) Scheduling server user interface built with Python Matplotlib and ROS RVIS for visualization and monitoring of the robots motions*

In the first session on Nov 10th, 33 trays of fruit were harvested manually in the field block. In the second session on Nov 10th, 41 trays were harvested by the 6 pickers working with the co-robotic harvest-aiding system. In the first session on Nov 11th, the same picking crew of 6 people harvested 24 trays of fruit using the harvest-aiding system. The data from these three sessions was processed and analyzed in Section 3.3. Non-parametric tests were used to compare the performance of manual and co-robotic harvesting.

## 3.3. Experimental data processing

The collected experimental data processing is introduced in this section. Table 14 shows the data fields recorded on the SD cards of the instrumented cart and transmitted wirelessly to the LoRa server. On the SD card, the data was recorded at the GPS epoch update frequency, i.e., at 10 HZ. The LoRa message transmission frequency of each cart was 1.6 Hz.

*Table 14 Data protocol recorded in the SD card on the instrumented cart*

| Item | Time stamp | Longitude | Latitude | Filtered mass | Button state |
|------|-----------|-----------|----------|---------------|--------------|
| Description | GPS epoch with accuracy of 100ms | Geodetic coordinates of the cart location | | Filtered mass readings of the tray by the instrumented cart | Status of button on the cart: 0 represents pressed and 1 represents not pressed |
| Data Type (Arduino Due) | Unsigned Int (32 bits) | Double (64 bits) | Double (64 bits) | Float (32 bits) | Unsigned short (8 bits) |

After receiving and processing a LoRa message, the "Cart-states-pub node" (Figure 36) converted it into a ROS message whose contents are shown in Table 15. The geodetic coordinates of the cart were converted to the coordinates in the field map.

*Table 15. Cart messages in ROS, converted by the LoRa server ROS node*

| Item | Header | Cart's ID | X | Y | Button Request |
|------|--------|-----------|---|---|----------------|
| Description | Time stamp of the data received. Frame ID of the data (local field frame); | 1~6 | Coordinate of the cart in the field map | | Request of the picker: 1 represents requesting and 0 represents not requesting |
| Data type (ROS) | std_msgs/Header.msg | std_msgs/Byte | std_msgs/Float32 | | std_msgs/Byte |

The methodology described in previous work (Khosro Anjom & Vougioukas, 2019) was used to compute these metrics from the cart and picker data; the time instants shown in Figure 44 were extracted from the logged data. The time instant $t_i^{\{end\}}$ when picking of a tray ended was identified by detecting a big drop in the measured mass of the tray, after the measured mass exceeded 4,000 grams. The instant $t_i^{\{start\}}$ that picking started was found by detecting the moment when the measured mass was around 500 grams. A tray-transport request time instant corresponded to a change in the state of the request button from "0" to "1". After extracting $t_i^{\{start\}}$ and $t_i^{\{end\}}$ for each tray, the productive interval $\Delta t_i^{ef}$ was calculated by Eq 38 and the non-productive interval $\Delta t_i^{fe}$ was calculated using Eq 39. The efficiency of the tray was calculated by Eq 40.

$$\Delta t_i^{ef} = t_i^{\{end\}} - t_i^{\{start\}} \qquad\qquad \text{Eq 38}$$

$$\Delta t_i^{fe} = t_{i+1}^{\{start\}} - t_i^{\{end\}} \qquad\qquad \text{Eq 39}$$

$$\text{E}_{ff_i} = \frac{\Delta t_i^{ef}}{\Delta t_i^{ef} + \Delta t_i^{fe}} \qquad\qquad \text{Eq 40}$$

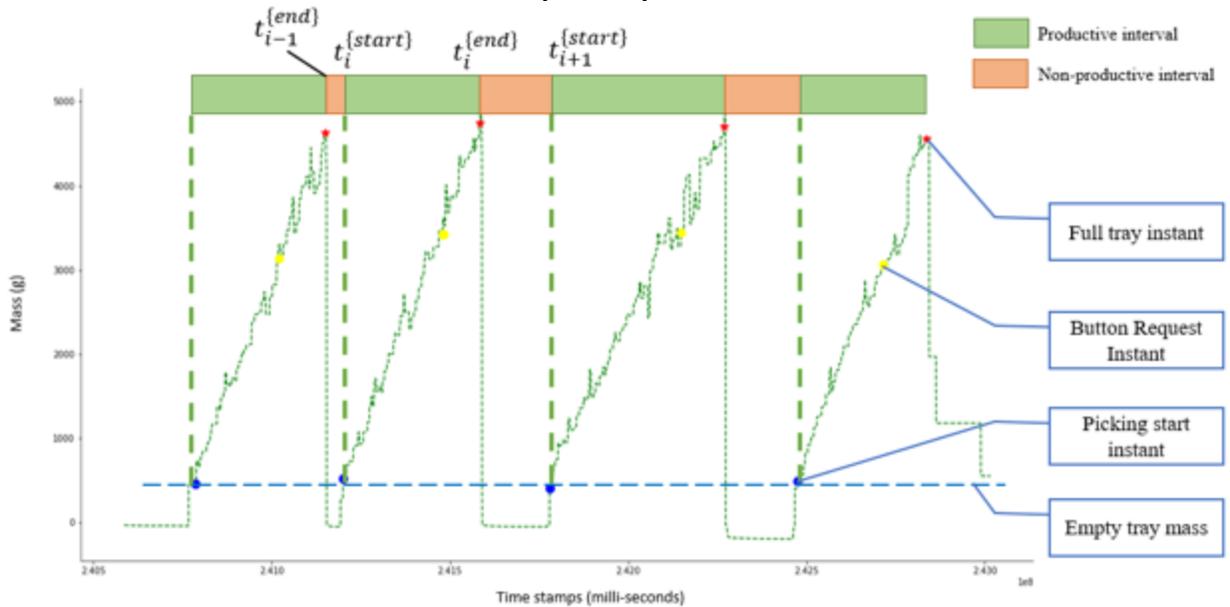

# 4. Experimental results and analysis

## 4.1. Results from *Robot-in-the-loop* experiment

The robot's navigation module was evaluated in the mapped field. The robots drove to their dispatched locations and then back to the collection station. The actual and planned paths are shown in Figure 45. All the paths were executed successfully. The tracking error was calculated by computing the Euclidean distance of each trajectory point to the closest planned path. The mean tracking error was 0.091m, the standard deviation was 0.078 m, and the maximum tracking error was 0.37m. Larger tracking errors occurred when the robot turned to enter or exit the rows.

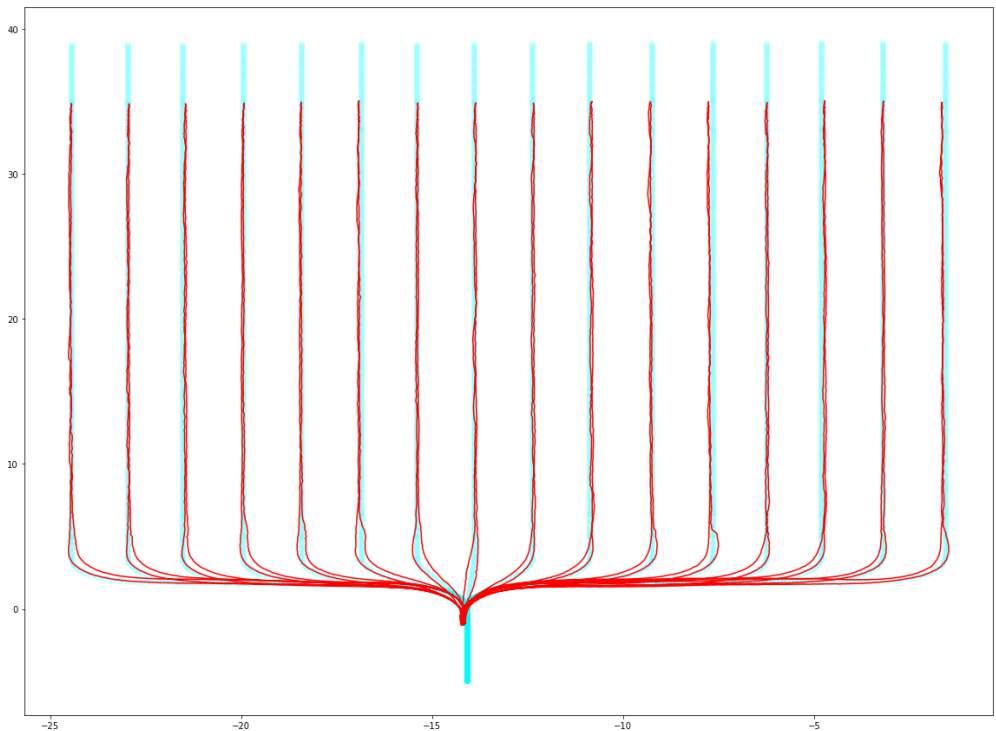

*Figure 45. FRAIL-Bot's real trajectory (red) and the planned path (cyan) from the navigation experiments in the mapped field*

In the '*Robot-in-the-loop*' experiments, different numbers of simulated pickers harvesting the whole field with the assistance of two FRAIL-Bots were evaluated. At the full tray instant (red star in Figure 44), the locations of those pickers could also be obtained from the simulation. Given these locations, the time required to transport these trays was estimated by approximating the transport speed from the mean value of the collected data (Seyyedhasani et al., 2020a). Hence, the evaluation metrics were estimated, if all these trays were transported by pickers instead of robots. Experimental results with different numbers of pickers are shown in Table 16. These mean values of the evaluated metrics were similar to the simulation results of Figure 23 in Chapter 3.

*Table 16. Mean harvesting efficiencies and non-productive times of co-robotic harvesting and manual harvesting for the experiment of simulated pickers in the campus field*

| Number of pickers | Co-robotic harvesting | | Manual harvesting | |
|---|---|---|---|---|
| | Average Harvesting efficiency per tray | Average Non-productive time per tray(s) | Harvesting efficiency | Non-productive time (s) |
| 4 | 0.889 | 33.8 | 0.791 | 76.3 |
| 5 | 0.876 | 45.2 | 0.785 | 77.5 |
| 6 | 0.873 | 46.2 | 0.783 | 73.1 |
| 7 | 0.861 | 49.1 | 0.794 | 76.6 |
| 8 | 0.848 | 52.0 | 0.797 | 75.9 |

## 4.2. Results from *Robot-and-cart* experiment

In the '*Robot-and-cart*' experiment on the campus field, the tray request prediction module shown in Figure 46 was evaluated. Simple linear regression was used to estimate the instant of next-tray transport request from the measured weights. The weight data up to one minute before the button was pressed were used for the regression. Purple points represent the data used to predict the next-tray transport request and black bars represent the prediction of the tray transport request time instants. The middle point of each black bar represents the mean of the predicted Gaussian distribution, while the width represents the standard deviation.

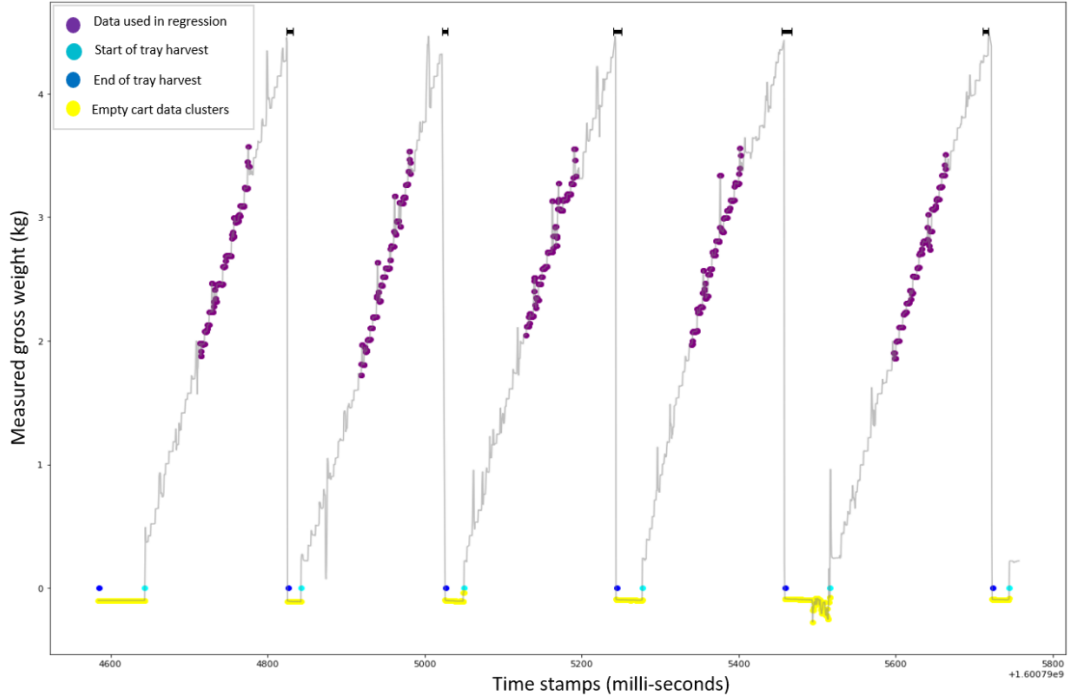

*Figure 46. The gray curves represent the tray's weight measurements over time. Purple points represent the data points used to predict the full tray time instant. The lengths of yellow segments represent the non-productive intervals during picking (each segment starts from a blue point and ends at a cyan point). Black bars represent the prediction of the times when trays become full.*

The temporal and spatial next-tray transport request predictions were evaluated in the experiments. The predicted full-tray locations (green polygons) and the true full-tray locations (gold cross) are shown in Figure 47. The mean distance error along the y direction was approximately 3.5 m. The temporal predictions of the collected 20 trays were evaluated using as metrics the mean average error (MAE) and mean standard error (MSE) (Anjom Khosro, 2019). The MAE of the prediction was 25.35s and MSE was 7.8s.

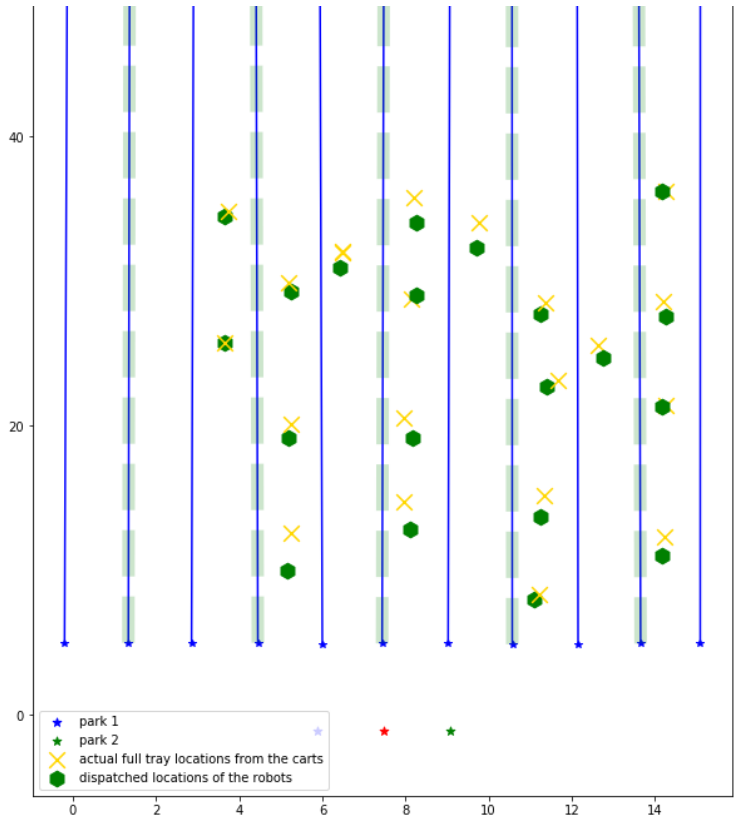

*Figure 47. Predicted full tray locations (green polygon) and the true full tray locations (gold cross)*

## 4.3. Results from the commercial field system experiment

The data from each instrumented cart was processed with the method described in Section 3.3 to get the productive time, $\Delta t_i^{ef}$, (aka 'one-tray-picking time') and non-productive time, $\Delta t_i^{fe}$, of each tray. The distance traversed to pick a single tray ('one-tray-picking distance') was obtained by calculating the Euclidean distance of the cart locations at the instances when picking started and ended for the tray. The results were combined and presented in this section.

### 4.3.1. Results for workers' harvesting parameters

The distributions of the one-tray-picking time, one-tray picking distance, and picker walking speed parameters were generated from the data. The one-tray picking time and one-tray picking distance parameters were assumed to be mainly dependent on the geospatial fruit distribution, for the same picking crew.

Figure 48 shows the distributions of the one-tray picking time of the crew for the co-robotic harvesting blocks of Nov 10th (session 2) and Nov 11th (session 1). The Mann-Whitney rank test was used to compare the two distributions. The null hypothesis was that for randomly selected values from the distribution of harvesting time per tray on Nov 10th and Nov 11th, the probability of the selected values on Nov 10th being greater than Nov 11th was equal to the probability of selected values on Nov 11th being greater than Nov 10th. The significance level of p-values (alpha) for rejecting the null hypothesis (Type I error) was chosen as 1%. The calculated p value was 2.13e-9, so the distributions of the two days differed significantly. A comparison of their mean values shows that, on average, the harvest crew took a longer time to harvest one tray on Nov 11th (894.62 s) than on Nov 10th (548.71 s).

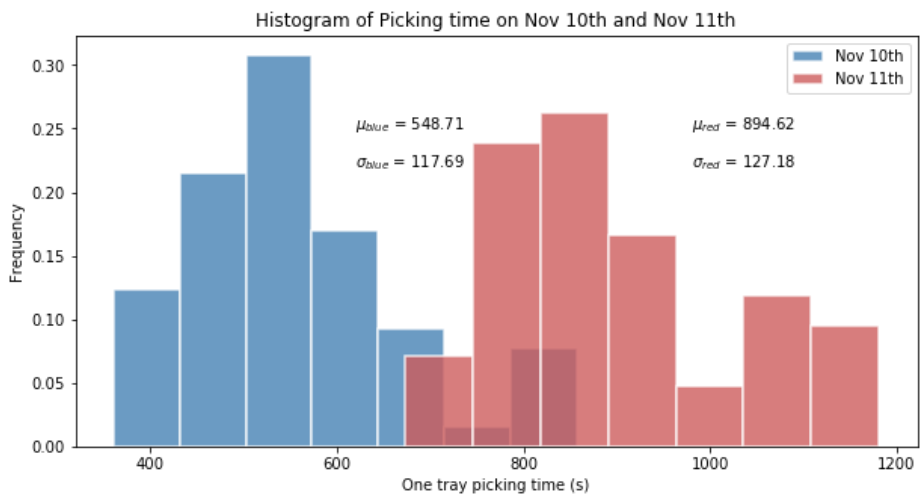

*Figure 48.  Histogram of the time it took pickers to fill one tray, on Nov 10th and Nov 11th*

Figure 49 shows the distributions of the one-tray picking distance, for the two days. The p value from the Mann-Whitney rank test of the two distributions was 2.21e-7, so they were significantly different. On average, the pickers moved a longer distance to collect a tray of strawberries on Nov 11[th] (33.31 m) than Nov 10[th] (17.92 m).

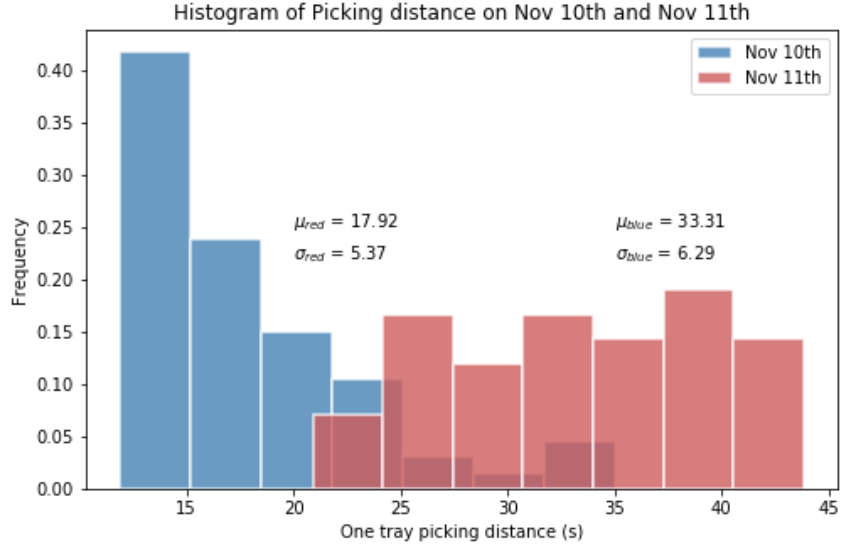

*Figure 49. Histogram of the distance traveled by pickers to fill one-tray, on Nov 10[th] and Nov 11[th]*

During manual harvesting, each picker would take their filled tray to the collection station, attach a sticker with their personal barcode on the tray, take an empty tray and walk back to the field to resume picking. Based on our observations, the pickers took around 8 seconds to stick their bar code on the tray and take an empty tray. Thus, the walking time to deliver a tray can be estimated by subtracting these 8 seconds from $\Delta t_i^{fe}$. The exact locations $\boldsymbol{L}_i^f$ when a picker start walking are detected by the weight change and the corresponding time instants are $t_i^{\{end\}}$. The transport distance can be computed from the coordinates of the collection station and $\boldsymbol{L}_i^f$. Hence, each picker's walking speed can be estimated from the computed transport distance and the measured time interval. The manual harvesting data from the first session of Nov 10[th] was used to estimate the mean walking speeds of the 6 pickers, as shown in Table 17.

| Picker ID# | Sample mean of walking speed (m/s) | Sample standard deviation of walking speed (m/s) | Number of manual transport measurements |
|---|---|---|---|
| 1 | 0.78 | 0.05 | 5 |
| 2 | 0.49 | 0.10 | 5 |
| 3 | 0.81 | 0.06 | 6 |
| 4 | 0.74 | 0.11 | 5 |
| 5 | 0.91 | 0.07 | 6 |
| 6 | 1.02 | 0.03 | 6 |

### 4.3.2. Harvesting performance of co-robotic harvesting

The non-productive time $\Delta t_i^{fe}$ and the efficiency $E_{ff_i}$ for each tray was obtained from the data collected during the co-robotic harvesting sessions. Obviously, it is impossible to have the picking crew re-harvest manually a field block that was harvested using the robots. Ideally, a large trial would use a large field and divide it into smaller blocks and then randomly assign manual and robotic treatments to the blocks. However, such a large trial was not acceptable by the grower; commercial harvesting is a costly operation that is planned based on the weather, crop condition, labor availability and customer demand. Hence, an estimation of the manual harvesting efficiency was made for the same field block that was harvested using robots, given the pickers' estimated walking speed and the measured locations where their trays had filled up.

As mentioned above, the location $\boldsymbol{L}_i^f$ of each tray can be indexed from the instant $t_i^{\{end\}}$ the tray becomes full. Thus, the non-productive time and the manual harvesting efficiency for each tray were estimated from $\boldsymbol{L}_i^f$ , the collection station position and the estimated walking speeds (from Section 2.1), In summary, the picking crew's harvesting performance with the co-robotic harvest-aiding system was measured directly, and the manual harvesting performance of the same picking crew for the same field block was estimated indirectly, using the above method.

Given the significant difference in the yields of the two fields, which manifested itself in very different one-tray picking time and distance statistics, the harvesting performance was evaluated on the two days separately. The frequency histograms of non-productive time for the manual harvesting and co-robotic harvesting on Nov 10th are shown in Figure 50.a, and the histograms of harvesting efficiency on that day are shown in Figure 50.b. The P-value of Mann-Whitney testing results of the performance data (non-productive time and harvesting efficiency) of manual harvesting and co-robotic harvesting on that day are shown in Table 18. Based on the calculated P-values, the manual and co-robotic distributions of the non-productive time and efficiency are significantly different, with a significance level at 1%.

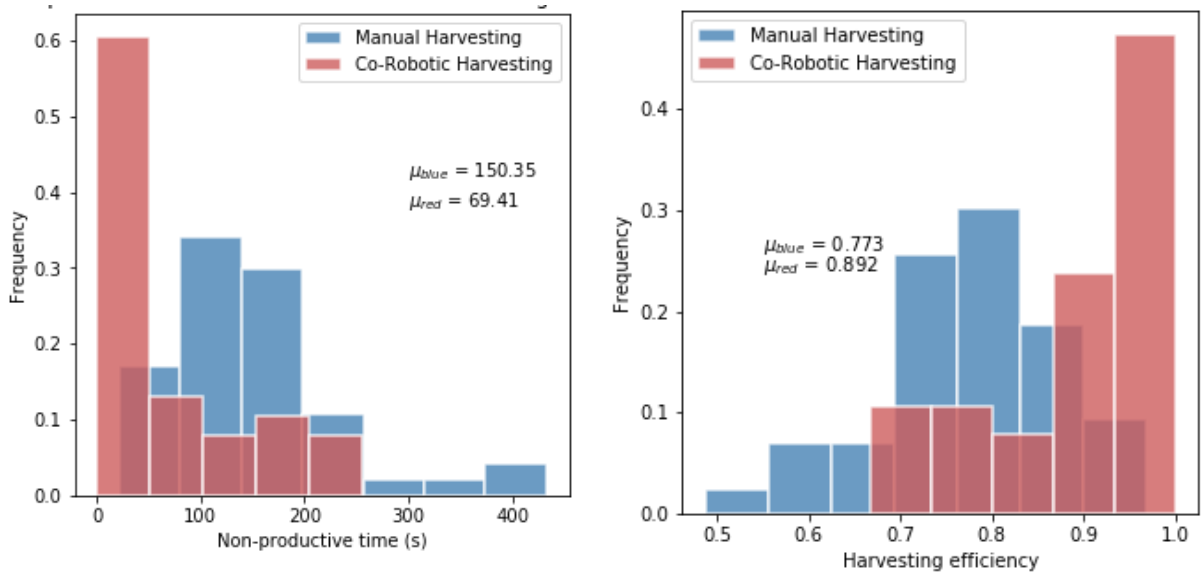

*Figure 50. Harvesting performance on Nov 10th: a) Frequency histogram of non-productive time of co-robotic and manual harvesting; b) Frequency histogram of harvesting efficiency of co-robotic and manual harvesting.*

*Table 18 Mann-Whitney rank test results for the means of the measured and estimated non-productive time and harvesting efficiency of the co-robotic and manual harvesting, respectively, on Nov 10th.*

| Item | Mean Non-productive time | Mean Harvesting efficiency |
|------|--------------------------|----------------------------|
| P value | 3.778e-6 | 1.3705e-6 |

Their mean values are shown in the figure. The mean non-productive time of co-robotic harvesting was reduced by more than half of the manual harvest non-productive time. The mean co-robotic harvesting efficiency increased by around 12% compared to manual harvesting. These results are shown in the corresponding pie charts, in Figure 51.

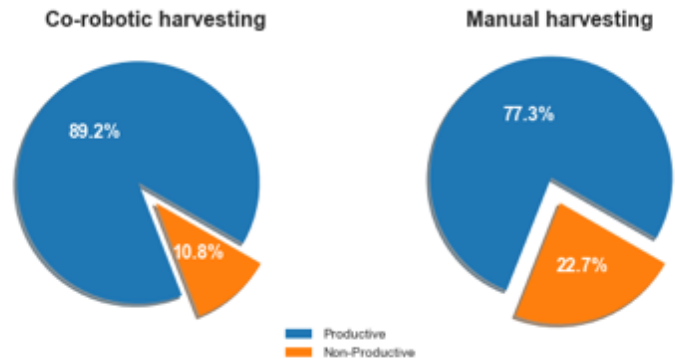

*Figure 51. Comparison between the mean harvesting efficiency of co-robotic and manual harvesting, based on experimental data, on Nov 10th.*

Similarly, the performance distributions of manual and co-robotic harvesting on Nov 11th are shown in Figure 52, and the Mann-Whitney rank test results are shown in Table 19. Based on the calculated P-values, the manual and co-robotic distributions of the non-productive time and efficiency are significantly different, at a significance level of 1%.  From Figure 52b, one can see that the mean non-productive time with the robots is 33% lower than that of manual harvesting. Figure 53 shows that the mean harvesting efficiency after introducing the robots improved by 8.8 %.

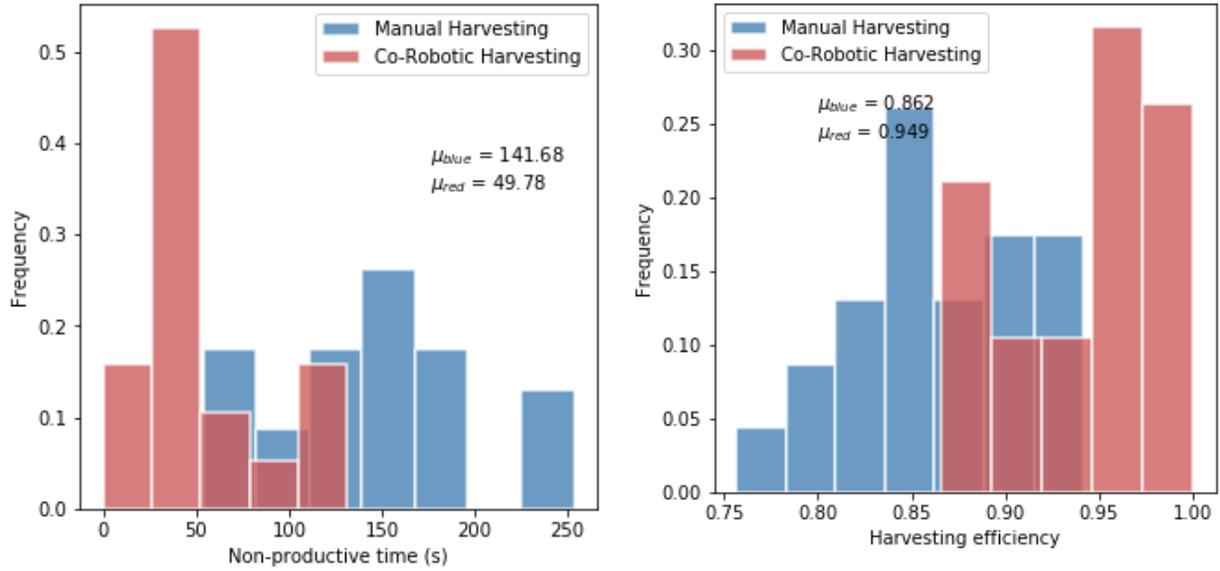

*Figure 52. Harvesting performance on Nov 11th: a) Histogram of non-productive time of co-robotic and manual harvesting; b) Histogram of harvesting efficiency of co-robotic and manual harvesting.*

*Table 19 Mann-Whitney rank test results for the measured and estimated non-productive time and harvesting efficiency of the co-robotic and manual harvesting, respectively, on Nov 11th*

| Item | Mean Non-productive time | Mean Harvesting efficiency |
|------|--------------------------|----------------------------|
| P value | 2.409e-7 | 3.723e-7 |

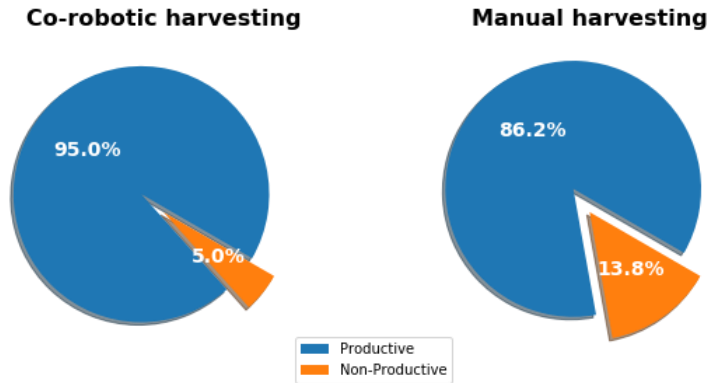

*Figure 53. Comparison between the mean harvesting efficiency of co-robotic and manual harvesting, based on experimental data, on Nov 11th.*

# 5. Discussion and conclusions

This chapter investigated the development and deployment of a harvest-aiding system comprising two tray-transport robots. The functionality of the modules inside the system was described and explained in detail. The core module, i.e., the predictive scheduling of the robot team, was mathematically modeled as an online dynamic scheduling problem with uncertain requests. The problem was solved with an adapted scenario-sampling based method (MSA) which output in real-time a fast and sub-optimal solution. The whole system was integrated and successfully deployed in commercial strawberry harvesting.

The field experiments demonstrated that the proof-of-concept system was functional in real-world commercial harvesting operations. The experimental results showed that the two harvest-aid robots significantly reduced the non-productive time of the six-person crew by around 60% and improved the harvesting efficiency by up to 10%, for the given crop load (which was low due to the season).

# Chapter 5 Summary, conclusions, and future work

In this dissertation, a co-robotic harvest-aiding system was developed to help alleviate the increasing labor shortage in strawberry harvesting. Dynamic robot scheduling under predictive transport requests was modeled and implemented. Two primary characteristics of the transport request predictions on scheduling performance were studied: earliness of prediction availability and uncertainty of predictions. These were studied on a harvesting simulator that modeled human pickers and transport robots and utilized manual harvesting model parameters estimated from data collected during harvesting of commercial strawberry fields. Finally, the whole system was integrated and deployed during commercial strawberry harvesting.

In Chapter 2, it was found that FR starts affecting the performance of the predictive scheduling after a FR threshold is reached. The FR threshold can be estimated in advance, given a specific harvesting configuration. The best possible case of predictive scheduling was examined. Given the robot-picker ratio of 1:3 and a robot travel speed of 1.5 m/s, the waiting time can be reduced by over 85% and the corresponding efficiency increase was over 15% with respect to all-manual harvesting.

In Chapter 3, more practical cases were integrated into the simulation platform: (1) the uncertainty of predictive transport requests was considered; (2) the robots traveled at a safer speed of 0.5 m/s on the headland and 1.2 m/s inside the furrow. From the simulation experiment results, when the robot picker ratio was 1:3, it was found that the harvesting efficiency improved over 8% relative to the manual harvesting and was 4% worse than the efficiency achieved by deterministic scheduling under perfect predictions. Also, when 7 or more robots aided the crew of 25 pickers, the co-robotic harvesting rate was higher than a 30-picker crew under manual harvesting.

In Chapter 4, the development of the harvest-aiding system was integrated and evaluated in real fields. These proof-of-concept experiments demonstrated that the whole system was implementable and applicable in a commercial harvesting scenario. Experimental results in a commercial field – during low-yield season - showed that when two robots aided a crew of six pickers, the harvesting efficiency increased by around 10% and the non-productive time was reduced by nearly 60%.

The simulations conducted in this thesis were based on picker data that were collected from specific crews, compensation schemes, fields, and picking seasons. The performance of the co-robotic harvest-aiding system – in simulation and in reality - depends strongly on these data and future work could apply the same methodology in different harvesting scenarios.

The following extensions to this thesis can support the deployment of harvest-aiding robots in real-world harvesting operations, and will be explored in future research:

(1) Since the capacity of the current FRAIL-Bot is more than one tray, the robot does not need to go back to the collection station immediately after getting one tray from a picker. Serving multiple pickers before returning to the collection station is expected to increase the efficiency. The problem is more complex and must be modelled explicitly and solved in real-time and render the system more applicable to other crops.

(2) Real-world interaction of human pickers and robots will involve unexpected events that will require increased autonomy, and safe, reactive behavior from the robots. Using existing sensor data and adding perception modalities (e.g., onboard cameras) to estimate the operating states of the pickers in the hybrid systems model could enhance human-robot collaboration and safety.

(3) At a technical level, the localization of the robots relies solely on RTK-GNSS. Errors in the field map or intermittent GNSS signal deterioration are not unlikely, and vision-aided navigation would enhance the robots' operational availability and robustness.

# Appendix

Here, it is shown that early requests, with an FR value below a certain threshold, cannot change the current schedule, and thus will not affect the performance of the system.

Let $\mathcal{S}^P$ be the set of all the next tray-transport requests. The size of $\mathcal{S}^P$, $|\mathcal{S}^P|$ is equal to Q, the number of pickers. Given an FR value, the set $\mathcal{S}^R$ of the predicted next tray-transport requests is a subset of $\mathcal{S}^P (\mathcal{S}^R \subseteq \mathcal{S}^P)$. When FR is 0, $S^R$ is equal to $S^P$, at any time instant. Let $S^{R\prime}$ be set of requests that have not been predicted yet, i.e., they are not available to the scheduler; $S^{R\prime}$ is the complement set $S^P \backslash S^R$. The size of $S^{R\prime}$ depends on FR. When FR is too small, some of the early requests - with too late release times – do not need to be considered at the instant when the scheduler dispatches the robot to the request whose release constraint has been reached. At a dispatching instant, the scheduler decides to schedule robot $k$ to request $R_i$. We assume this dispatch command will not be affected by ignoring any requests in $S^{R\prime}$, so, for $\forall R_j \in S^{R\prime}$, sum of the wait time of schedule $\{R_j, R_i\}$ must be larger than sum of the wait time of schedule $\{R_i, R_j\}$ for robot k. We can get wait times for these two possible schedules in Table A1, based on Eq 9.

| Schedule | Wait time of $\boldsymbol{R_i}$ | Wait time of $\boldsymbol{R_j}$ | Sum of wait times |
|---|---|---|---|
| $\{R_j, R_i\}$ | $\Delta t_j^p + \Delta t_j^r + \Delta t_i^u - \Delta t_i^f$ | $0$ | $\Delta t_j^p + \Delta t_j^r + \Delta t_i^u - \Delta t_i^f$ |
| $\{R_i, R_j\}$ | $\Delta t_i^u - \Delta t_i^f$ | $\Delta t_i^p$ | $\Delta t_i^u - \Delta t_i^f + \Delta t_i^p$ |

*Table A1. Wait times of two possible schedules: $\{R_j, R_i\}$ and $\{R_i, R_j\}$*

The sum of wait times of schedule $\{R_j, R_i\}$ must be larger than that of $\{R_i, R_j\}$; therefore:

$$\Delta t_j^p + \Delta t_j^r + \Delta t_i^u - \Delta t_i^f \geq \Delta t_i^u - \Delta t_i^f + \Delta t_i^p \qquad \text{(Eq A.1)}$$

The expression can be simplified further:

$$\Delta t_j^p + \Delta t_j^r \geq \Delta t_i^p \qquad \text{(Eq A.2)}$$

Substituting $\Delta t_j^r = \Delta t_{pick}^i (1 - FR_j) - \Delta t_j^u$ into Eq A.2, we get:

$$\Delta t_j^p + \Delta t_{pick}^i (1 - FR_j) - \Delta t_j^u \geq \Delta t_i^p \qquad \text{(Eq A.3)}$$

where $FR_j$ is the fill ratio of request $R_j$. Using Eq 15 for $R_j$ and $R_i$, Eq A.3 can be written as:

$$\Delta t_j^u + \Delta t_{pick}^i (1 - FR_j) \geq 2\Delta t_i^u \qquad \text{(Eq A.4)}$$

Solving for $FR_j$ results in Eq A.5.

$$FR_j \leq 1 - \frac{2\Delta t_i^u - \Delta t_j^u}{\Delta t_{pick}^i} \qquad \text{(Eq A.5)}$$

This inequality must always exist, even when the right-hand side is at minimum, i.e.:

$$FR_j \leq \left( 1 - \frac{2\Delta t_i^u - \Delta t_j^u}{\Delta t_{pick}^i} \right)_{min} \qquad \text{(Eq A.6)}$$

Approximating $\Delta t_i^u \approx \Delta t_j^u$ and $\Delta t_i^{pick} \approx \overline{\Delta t^{pick}}$, we get Eq A.7.

$$FR_j \leq \left( 1 - \frac{\Delta t_j^u}{\overline{\Delta t^{pick}}} \right)_{min} \qquad \text{(Eq A.7)}$$

When $\Delta t_j^u$ takes its maximum value, the right-hand side is minimized:

$$FR_j \leq 1 - \frac{(\Delta t_j^u)_{max}}{\overline{\Delta t^{pick}}} \qquad \text{(Eq A.8)}$$

Therefore, when a request in $S^{R\prime}$ meets Eq A.8, it can be ignored at the dispatching instant.